\documentclass[11pt,letterpaper]{article}

\usepackage[margin=1in]{geometry}
\usepackage[round,authoryear]{natbib}
\usepackage[T1]{fontenc}
\usepackage{times}
\usepackage{microtype}
\usepackage{amsmath,amssymb,amsthm,mathtools,bm,mathrsfs}
\usepackage{aliascnt}
\usepackage{booktabs,array,tabularx,multirow,longtable}
\usepackage{enumitem}
\usepackage{graphicx}
\usepackage{placeins}
\usepackage{xcolor}
\usepackage{url}
\usepackage{authblk}
\usepackage{algorithm}
\usepackage{algorithmic}
\usepackage{hyperref}
\usepackage{subcaption}
\usepackage[nameinlink,noabbrev,capitalise]{cleveref}
\hypersetup{colorlinks=true,linkcolor=blue!45!black,citecolor=blue!45!black,urlcolor=blue!45!black}
\newtheorem{theorem}{Theorem}[section]
\newaliascnt{proposition}{theorem}
\newtheorem{proposition}[proposition]{Proposition}
\aliascntresetthe{proposition}
\newaliascnt{lemma}{theorem}
\newtheorem{lemma}[lemma]{Lemma}
\aliascntresetthe{lemma}
\newaliascnt{corollary}{theorem}
\newtheorem{corollary}[corollary]{Corollary}
\aliascntresetthe{corollary}
\newaliascnt{assumption}{theorem}
\newtheorem{assumption}[assumption]{Assumption}
\aliascntresetthe{assumption}
\newaliascnt{condition}{theorem}

\aliascntresetthe{condition}
\theoremstyle{definition}
\newaliascnt{definition}{theorem}

\aliascntresetthe{definition}
\theoremstyle{remark}
\newaliascnt{remark}{theorem}
\newtheorem{remark}[remark]{Remark}
\aliascntresetthe{remark}
\crefname{theorem}{Theorem}{Theorems}
\Crefname{theorem}{Theorem}{Theorems}
\crefname{proposition}{Proposition}{Propositions}
\Crefname{proposition}{Proposition}{Propositions}
\crefname{lemma}{Lemma}{Lemmas}
\Crefname{lemma}{Lemma}{Lemmas}
\crefname{corollary}{Corollary}{Corollaries}
\Crefname{corollary}{Corollary}{Corollaries}
\crefname{assumption}{Assumption}{Assumptions}
\Crefname{assumption}{Assumption}{Assumptions}
\crefname{condition}{Condition}{Conditions}
\Crefname{condition}{Condition}{Conditions}
\crefname{definition}{Definition}{Definitions}
\Crefname{definition}{Definition}{Definitions}
\crefname{remark}{Remark}{Remarks}
\Crefname{remark}{Remark}{Remarks}

\setlist[itemize]{leftmargin=1.35em,itemsep=.15em,topsep=.25em}
\setlist[enumerate]{leftmargin=1.55em,itemsep=.15em,topsep=.25em}

\usepackage{pifont}

\crefname{algorithm}{Algorithm}{Algorithms}
\Crefname{algorithm}{Algorithm}{Algorithms}

\title{Finite-Time Online Sample Complexity for Conditional Diffusion Models}

\author[1]{Mudit Gaur}
\author[2]{Prashant Trivedi}
\author[3]{Sasidhar Kunapuli}
\author[4]{Amrit Singh Bedi}
\author[5]{Vaneet Aggarwal}

\affil[1]{Department of Electrical and Computer Engineering, Ohio State Univeristy}
\affil[2]{Department of Computer Science and Information Systems, BITS Pilani}
\affil[4]{ Department of Computer Science, University of Central Florida}
\affil[5]{ Edwardson School of Industrial Engineering, Purdue University}

\date{}

\begin{document}
\maketitle
\begin{abstract}
We study the finite-time training and sampling error of conditional diffusion
models trained by online stochastic-gradient denoising score matching. Rather
than assuming access to an exact empirical or population risk minimizer, we
analyze the score network produced after a finite number of stochastic-gradient
updates under an explicit localized Polyak--\L{}ojasiewicz (PL) condition,
stochastic-gradient regularity, and a trajectory-containment assumption. Our
analysis separates the network
approximation complexity from the online training-sample budget and propagates
both errors through the reverse diffusion process. Under anisotropic smoothness,
a joint intrinsic-dimensional structure for the data and conditioning
variables, sub-exponential tails, and explicit approximation-class
compatibility, we obtain a
conditional score approximation error of order
\[
\widetilde{\mathcal O}\!\left(M^{-\eta_{\mathrm{app}}}\right),
\qquad
\eta_{\mathrm{app}}
=
\frac{2\alpha_x}
{2\alpha_x+d_x'+(\alpha_x/\alpha_y)d_y'},
\]
where \(M\) denotes the approximation complexity, \(d_x'\) and \(d_y'\) are
the intrinsic response and conditioning dimensions, and \(\alpha_x,\alpha_y\)
are their corresponding smoothness exponents. For \(N=RB\) online denoising
samples used by SGD, the finite-time optimization error is controlled explicitly
in terms of the PL, smoothness, gradient-noise, and initialization constants.
Combining these ingredients with reverse-process stability yields a
total-variation bound that decomposes into terminal-initialization,
time-discretization, approximation/localization, and finite-time optimization
errors.

The novelty is the conditional synthesis: relative to conditional statistical
analyses based on exact risk minimizers, we retain the finite online
optimization error; relative to existing unconditional gradient-training
analyses, we couple optimization with joint response--conditioning intrinsic
geometry and conditional reverse-process stability.

Our primary guarantee is for the forward-smoothed conditional law at a positive
stopping time \(t_0>0\). Under an additional full-dimensional
integral density-regularity condition, we further control the smoothing bias and extend
the guarantee to the original conditional data law. The resulting analysis
provides a transparent training-to-sampling guarantee for conditional diffusion
under finite-time online stochastic optimization, while making explicit the
optimization and approximation assumptions required for the result.
\end{abstract}

\section{Introduction}
\label{sec:introduction}

Diffusion and score-based generative models have become a central framework for
high-dimensional generative modeling
\citep{sohl2015deep,ho2020denoising,song2021score}. Conditional diffusion
models extend this framework by learning a family of generative distributions
indexed by side information, enabling applications such as text-to-image
generation \citep{dhariwal2021diffusion,rombach2022high}, conditional molecular
design \citep{malusare2024improving}, and conditional time-series imputation
\citep{tashiro2021csdi}. In score-based formulations, the learned object is a
conditional score
\[
    s^\star(x,t,y)
    =
    \nabla_x \log p_t(x\mid y),
\]
which depends jointly on the noised data variable \(x\), the conditioning
variable \(y\), and diffusion time \(t\), and is used to construct the
reverse-time generative dynamics. Practical conditional diffusion systems employ
a variety of conditioning mechanisms, including classifier guidance
\citep{dhariwal2021diffusion,shenoy2024gradientfree}, classifier-free guidance
\citep{ho2022classifierfree}, and transformer-based score architectures
\citep{peebles2023scalable}.

A complete theoretical analysis of conditional diffusion training must connect
several distinct components. First, the conditional score must be approximated
by the chosen function class. Second, the score-matching objective must be
optimized using only a finite amount of computation and data. Third, the
resulting score-estimation error must be propagated through a discretized
reverse diffusion process. Finally, when the data or conditioning variables are
unbounded, the approximation argument must control contributions from regions
far from the bulk of the distribution. These issues are coupled in conditional
models because approximation takes place jointly over the response,
conditioning variable, and diffusion time.

In this work, we study this training-to-sampling problem for conditional
diffusion models trained by \emph{online stochastic-gradient denoising score
matching}. Our objective is not to assume that the training procedure returns
an exact optimizer and then analyze the resulting score estimator. Instead, we
ask how the error of a score network obtained after finitely many stochastic
gradient updates propagates to the distribution generated by the reverse
process. More precisely, we seek a decomposition in which approximation error,
finite-time optimization error, discretization error, and terminal
initialization error remain separately visible.

This perspective differs from a common abstraction in statistical analyses of
conditional diffusion, where the score estimator is defined through an exact
empirical or population risk minimizer
\citep{yuan2023reward,fu2024unveil,tang2025conditional,hu2025statistical}.
Such an abstraction is useful for isolating score-estimation error, but it does
not directly describe the output of a finite stochastic-gradient training
procedure. A fixed-dataset analysis may additionally require controlling the
generalization gap between empirical and population objectives. Finite-time
gradient training has already been analyzed in unconditional settings
\citep{gaur2025improvedsamplecomplexitydiffusion,han2024neural}. Recent work has
also analyzed SGD dynamics for unconditional score-based generative models
\citep{strasman2026nonasymptotic}. Thus finite-time gradient training is not, by
itself, the novelty claimed here. We instead consider an online
sampling model in which each stochastic-gradient update uses fresh draws of the
data pair, diffusion time, and Gaussian perturbation. This allows the
optimization error to be analyzed directly at the level of the population
randomized-time denoising objective.

Replacing exact minimizer access by finite-time optimization requires additional
optimization regularity. We make this tradeoff explicit. In particular, our
analysis assumes local smoothness, stochastic-gradient regularity, and a
localized Polyak--\L{}ojasiewicz (PL) condition along the relevant training
trajectory. The PL condition is \emph{not} derived in this paper for the full
neural score class; rather, it is an explicit optimization assumption under
which the effect of a finite online SGD budget can be quantified. Thus, the
finite-time result should be interpreted as a conditional training guarantee:
given the stated localized optimization geometry, we characterize how the
remaining optimization error enters the final sampling bound.

The approximation side of the analysis addresses a different difficulty.
Conditional score estimation can depend strongly on the dimensions of both the
response and the conditioning variable. Some existing guarantees exhibit
ambient-dimensional dependence \citep{fu2024unveil}, while intrinsic-dimensional
analyses exploit manifold or latent low-dimensional structure
\citep{tang2025conditional,hu2025statistical}. We work with a joint
intrinsic-dimensional structure for the clean pair \((X_0,Y)\), together with
anisotropic regularity that permits different smoothness along response and
conditioning coordinates. We also impose an explicit approximation-class
compatibility condition linking our smooth-gated score class to the localized
blockwise construction used in the cited conditional approximation theorem;
the geometric assumptions alone do not imply the claimed network rate. We
denote the corresponding intrinsic dimensions by
\(d_x'\) and \(d_y'\), and their smoothness exponents by \(\alpha_x\) and
\(\alpha_y\), respectively. The associated approximation exponent is
\[
    \eta_{\mathrm{app}}
    :=
    \frac{2\alpha_x}
    {2\alpha_x+d_x'+(\alpha_x/\alpha_y)d_y'}.
\]
We additionally allow the data and conditioning variables to have
sub-exponential tails. The proof localizes the denoising problem to
logarithmic-radius high-probability regions and controls the contribution of
their complements through tail decay.

\paragraph{Main result and scope.}
A central feature of our formulation is that we distinguish the complexity of
the score class from the amount of online training data. Let
\[
    M
    =
    \text{score-network approximation complexity},
    \qquad
    N=RB
    =
    \text{number of online denoising samples},
\]
where \(R\) is the number of SGD iterations and \(B\) is the mini-batch size.
For a discretization grid with \(K\) steps and total time weight \(H_K\), our
main theorem takes the schematic form
\[
\begin{aligned}
&\mathbb E_{\mathrm{SGD}}\mathbb E_Y
\left[
\operatorname{TV}\!\left(
p_{t_0}(\cdot\mid Y),
\widehat p_{t_0}(\cdot\mid Y)
\right)
\right]
\\
&\qquad\lesssim
C_{\mathrm{term}}e^{-T}
+
C_{\mathrm{disc}}
\sqrt{\frac{d_x+m_2}{K}}
+
C_{\mathrm{stab}}
\sqrt{
H_K\left[
\mathcal A_K(M)
+
\mathcal E_{\mathrm{opt}}(N,M)
\right]
},
\end{aligned}
\]
under the stated localized PL and trajectory-containment assumptions. Here
\(\mathcal A_K(M)\) is the score-class approximation/localization error and
\(\mathcal E_{\mathrm{opt}}(N,M)\) is the excess population denoising risk
remaining after finite-time online optimization. Under the anisotropic
intrinsic-dimensional approximation assumptions developed in this paper,
\[
    \mathcal A_K(M)
    =
    \widetilde{\mathcal O}
    \!\left(M^{-\eta_{\mathrm{app}}}\right).
\]
Under localized PL geometry and the decreasing step-size schedule analyzed
below, the optimization component satisfies a bound of the form
\[
    \mathcal E_{\mathrm{opt}}(N,M)
    \lesssim
    \frac{\Delta_{0,M} r_0^2}{(R+r_0)^2}
    +
    \frac{L_M\sigma_M^2}
    {\mu_M^2 B(R+r_0)},
    \qquad
    N=RB,
\]
where \(L_M\), \(\mu_M\), and \(\sigma_M^2\) denote the relevant smoothness,
PL, and stochastic-gradient variance constants, and
\(\Delta_{0,M}\) is the initial population excess risk. Keeping these quantities
explicit separates the approximation and optimization requirements of the
result. If these optimization constants remain uniformly controlled along the
chosen sequence of score classes, the leading requirements reduce, up to
logarithmic factors, to
\[
    M
    =
    \widetilde{\Theta}
    \!\left(\epsilon^{-2/\eta_{\mathrm{app}}}\right),
    \qquad
    N
    =
    \widetilde{\Theta}
    \!\left(\epsilon^{-2}\right),
\]
together with a sufficiently fine reverse-time discretization, for
\(\widetilde{\mathcal O}(\epsilon)\) expected total-variation error relative to the
positive-time target \(p_{t_0}(\cdot\mid Y)\).

The positive stopping time \(t_0>0\) is important for interpreting this
statement. Our primary result concerns the forward-smoothed conditional law
\(p_{t_0}(\cdot\mid y)\). This permits intrinsic low-dimensional clean-data
structure while the reverse-process analysis operates away from the singular
endpoint \(t=0\). A total-variation guarantee for the original law
\(p_0(\cdot\mid y)\) requires an additional smoothing-bias argument. We provide
such an extension under stronger full-dimensional integral density regularity.
The stopped-law theorem and the original-law extension therefore have different
assumptions and should be viewed as separate results.

Table~\ref{tab:conditional_diffusion_comparison} separates two adjacent lines
of work: conditional distribution-estimation theory and algorithm-dependent
score-training theory. The assumptions, estimator definitions, and target laws
differ, so their rates are not directly comparable. The table locates our
contribution at their intersection; it does not assert priority for finite-time
diffusion optimization in general.

\begin{table}[t]
\centering
\caption{Structural comparison with representative adjacent theory. ``Gradient
training'' records whether a concrete finite-run gradient method is analyzed.
The rows are not rate comparisons.}
\label{tab:conditional_diffusion_comparison}
\resizebox{\linewidth}{!}{
\begin{tabular}{lcccc}
\toprule
\textbf{Reference}
&
\textbf{Conditional}
&
\textbf{Intrinsic dimension}
&
\textbf{Gradient training}
&
\textbf{Sampling bound}
\\
\midrule
\citet{fu2024unveil} & Yes & No & No & Yes \\
\citet{tang2025conditional} & Yes & Yes & No & Yes \\
\citet{hu2025statistical} & Yes & Yes & No & Yes \\
\citet{li2025conditionalnonasymptotic} & Yes & No & No & Yes \\
\citet{gaur2025improvedsamplecomplexitydiffusion} & No & No & Finite-time SGD & Yes \\
\citet{han2024neural} & No & No & GD with early stopping & No (training only) \\
\citet{strasman2026nonasymptotic} & No & No & Finite-time SGD & No (training only) \\
\midrule
\textbf{This work} & Yes & Yes & Online SGD under local PL & Yes \\
\bottomrule
\end{tabular}
}
\end{table}

\paragraph{Contributions.}
The main contributions of the paper are as follows:
\begin{itemize}[leftmargin=0pt,labelindent=0pt,labelwidth=!,itemindent=6pt,
align=left,itemsep=2pt,topsep=2pt,parsep=0pt]

    \item \textbf{Finite-time training-to-sampling decomposition.}
    We give an end-to-end error decomposition for conditional diffusion trained
    by online stochastic-gradient denoising score matching. The bound keeps
    terminal initialization, reverse-time discretization, score-class
    approximation/localization, and finite-time optimization as separate
    components. In particular, the score network need not be an exact
    population or empirical minimizer; the remaining optimization error is
    quantified explicitly under localized PL geometry.

    \item \textbf{Explicit separation of approximation and optimization
    complexity.}
    We distinguish the score-class complexity \(M\) from the online denoising
    sample budget \(N\). Under anisotropic intrinsic-dimensional regularity and
    the stated approximation-class compatibility condition, the approximation
    contribution scales as
    \[
        \widetilde{\mathcal O}
        \!\left(M^{-\eta_{\mathrm{app}}}\right),
        \qquad
        \eta_{\mathrm{app}}
        =
        \frac{2\alpha_x}
        {2\alpha_x+d_x'+(\alpha_x/\alpha_y)d_y'},
    \]
    while the finite-time optimization term is controlled separately through
    the smoothness, PL, gradient-noise, initialization, and online-sample
    parameters. This separation avoids conflating network approximation scale
    with the number of samples consumed by SGD.

    \item \textbf{Conditional approximation with intrinsic structure and
    sub-exponential tails.}
    We combine a joint response--conditioning intrinsic geometry with
    anisotropic smoothness and a sub-exponential tail-localization argument.
    The resulting construction controls the randomized-time conditional score
    approximation on logarithmic-radius high-probability regions and then
    bounds the contribution of the tail complement.

    \item \textbf{Stopped-law guarantee and original-law extension.}
    Our primary theorem controls the learned discretized sampler relative to the
    forward-smoothed conditional law at a positive stopping time \(t_0\).
    Under additional full-dimensional integral density regularity, we prove a
    smoothing-bias bound and extend the analysis to total-variation convergence
    to the original conditional law. This makes explicit where stronger
    regularity is required and avoids conflating the two guarantees.
\end{itemize}

At a technical level, the analysis makes compatible three ingredients that are
usually treated separately: conditional score approximation, finite-time
stochastic optimization, and reverse-diffusion stability. We use existing
reverse-process stability and conditional approximation results as inputs where
appropriate, while proving the localization, randomized-time approximation
transfer, finite-time online-SGD excess-risk control, and their combination
into the final training-to-sampling guarantee. Our emphasis is therefore on a
transparent and auditable finite-time analysis, including the assumptions
under which each component is valid, rather than on claiming a uniformly
improved statistical exponent over existing conditional diffusion theory.
Relative to unconditional optimization analyses
\citep{gaur2025improvedsamplecomplexitydiffusion,han2024neural,
strasman2026nonasymptotic}, the additional contribution is the conditional
coupling: one time-conditioned score must approximate a family indexed by
\(Y\), the approximation rate depends on joint fibered geometry, and the
resulting randomized-time population error must match the forward-marginal
quantity in the conditional reverse-process stability bound. We do not claim a
new proof of PL geometry for neural networks or a new reverse-SDE
discretization theorem.

\section{Preliminaries and Problem Formulation}
\label{sec:prelims_problem}

Let \((X_0,Y)\sim P_0\) be a random pair, where
\(Y\in\mathbb R^{d_y}\) has marginal law \(P_Y\), and, conditional on
\(Y=y\), the data variable \(X_0\in\mathbb R^{d_x}\) has conditional law
\(p_0(\cdot\mid y)\). Thus
\[
    P_0(dx,dy)
    =
    p_0(dx\mid y)P_Y(dy).
\]
We write \(d=d_x+d_y\). The unconditional setting is recovered by taking
\(d_y=0\). We use \(p_t(\cdot\mid y)\) for a conditional probability law and,
when the corresponding density exists, for its density as well.

Throughout, \(\widetilde{\mathcal O}\) and \(\widetilde{\Theta}\) suppress
only multiplicative constants independent of the displayed asymptotic
variables and factors that are polylogarithmic in those variables. Dependence
that is polynomial in a vanishing time cutoff such as \(t_0\) is displayed
explicitly. Fixed distributional, geometric, and moment parameters may enter
the implicit constants unless a result states otherwise.

The generative target is the original conditional law
\(p_0(\cdot\mid y)\). Our primary theorem, however, controls the learned sampler
relative to the forward-smoothed law \(p_{t_0}(\cdot\mid y)\) at a positive
stopping time \(t_0>0\). Convergence to the original law is treated separately
through a smoothing-bias argument under stronger full-dimensional density
regularity.

\paragraph{Forward diffusion and conditional score.}
We consider the variance-preserving Ornstein--Uhlenbeck forward process
\begin{equation}
    dX_t
    =
    -X_t\,dt+\sqrt{2}\,dB_t,
    \qquad
    X_0\sim p_0(\cdot\mid y),
    \label{eq:forward_sde_cond}
\end{equation}
where \(B_t\) is a standard Brownian motion independent of \((X_0,Y)\).
Conditional on \(X_0=x_0\), the forward transition can be written as
\[
    X_t
    =
    e^{-t}X_0+\sigma_t\varepsilon,
    \qquad
    \sigma_t^2:=1-e^{-2t},
    \qquad
    \varepsilon\sim\mathcal N(0,I_{d_x}),
\]
with \(\varepsilon\) independent of \((X_0,Y)\). Consequently,
\[
    q_t(x\mid x_0)
    =
    \mathcal N\!\left(
        e^{-t}x_0,\,
        \sigma_t^2 I_{d_x}
    \right),
\]
and the corresponding kernel score is
\begin{equation}
    \nabla_x\log q_t(x\mid x_0)
    =
    -\frac{x-e^{-t}x_0}{\sigma_t^2}.
    \label{eq:kernel_score}
\end{equation}

Let \(p_t(\cdot\mid y)\) denote the conditional law of \(X_t\) given \(Y=y\).
The conditional marginal score is
\begin{equation}
    s^\star(x,t,y)
    :=
    \nabla_x\log p_t(x\mid y)
    \in\mathbb R^{d_x}.
    \label{eq:true_conditional_score}
\end{equation}
The distinction between dimensions is important throughout the paper:
\(x\in\mathbb R^{d_x}\), \(y\in\mathbb R^{d_y}\), whereas the score itself
always takes values in \(\mathbb R^{d_x}\).

\paragraph{Reverse process and stopped target.}
Fix diffusion times \(0<t_0<1\le T\), as required by the early-stopped
reverse-process stability result used below. Writing
\(\bar X_\tau=X_{T-\tau}\), stochastic time reversal
\citep{anderson1982reverse} gives
\begin{equation}
    d\bar X_\tau
    =
    \left(
        \bar X_\tau
        +
        2s^\star(\bar X_\tau,T-\tau,y)
    \right)d\tau
    +
    \sqrt{2}\,d\bar B_\tau,
    \qquad
    0\le \tau\le T-t_0,
    \label{eq:reverse_sde_cond}
\end{equation}
when initialized from the forward terminal law \(p_T(\cdot\mid y)\).
Running this process down to forward time \(t_0\) recovers
\(p_{t_0}(\cdot\mid y)\).

The implementable sampler differs from this ideal reverse process in three
ways: it is initialized from the reference Gaussian
\(\mathcal N(0,I_{d_x})\), the true score \(s^\star\) is replaced by a learned
score, and the reverse SDE is discretized. These three changes give rise,
respectively, to the terminal-initialization, score-estimation, and
time-discretization terms in the main theorem.

\paragraph{Approximation complexity and score class.}
We distinguish throughout the paper between the complexity of the score class
and the number of samples consumed by stochastic-gradient training. Let
\[
    M
    =
    \text{approximation-complexity index},
    \qquad
    N
    =
    \text{number of online denoising samples used for training}.
\]
These are independent quantities.

For each approximation scale \(M\), let
\[
    \mathcal F_{\Theta,M}
    =
    \{s_\theta:\theta\in\Theta_M\}
\]
denote the conditional score class. The class used in the approximation
analysis is a single time-conditioned, smooth-gated dyadic-block class. Let
\[
    t_0=u_0<u_1<\cdots<u_I=T
\]
be a dyadic cover satisfying
\[
    u_{i+1}\le 2u_i,
    \qquad
    i=0,\ldots,I-1.
\]
We take
\begin{equation}
    s_\theta(x,y,t)
    =
    \sum_{i=0}^{I-1}
    \chi_i(t)s_{\theta_i}^{(i)}(x,y,t),
    \label{eq:network_class_prelim}
\end{equation}
where \(\{\chi_i\}_{i=0}^{I-1}\) are fixed deterministic nonnegative smooth gates forming
a partition of unity over the time interval and each
\(s_{\theta_i}^{(i)}\) belongs to the corresponding blockwise conditional
score-approximation class.

The gates are part of the theoretical construction and are not optimized by
SGD. Their support, overlap, and derivative bounds are absorbed into the
approximation-class compatibility condition and the logarithmic factors in the
rate. The index \(M\) records the calibrated approximation scale of this class;
it is not identified with either the raw parameter count or the training sample
size \(N\).
For the localized approximants used in the proof, the parameter bounds imply a
polynomial output envelope: for fixed \(t_0>0\), there exist fixed exponents
\(q_1,q_2\) such that
\[
\|s_\theta(x,y,t)\|_2
\le
C M^{q_1}(1+\|x\|_2+\|y\|_2)^{q_2},
\qquad t\in[t_0,T].
\]
This condition is used only to control the tail complement after localization.

\paragraph{Time grid and randomized-time distribution.}
Let
\[
    t_0<t_1<\cdots<t_K=T
\]
be the time grid used in training and reverse sampling, and define
\[
    \Delta_k:=t_k-t_{k-1},
    \qquad
    H_K:=\sum_{k=1}^{K}\Delta_k=T-t_0,
    \qquad
    \Delta_{\max}:=\max_{1\le k\le K}\Delta_k.
\]
Let \(J\in\{1,\ldots,K\}\) be independent of
\((X_0,Y,\varepsilon)\), with
\begin{equation}
    \mathbb P(J=k)
    =
    \pi_{K,k}
    :=
    \frac{\Delta_k}{H_K}.
    \label{eq:grid_time_distribution}
\end{equation}
This choice makes expectation with respect to \(J\) exactly the normalized
time-weighted grid average that appears in the reverse-process stability
argument. The right-endpoint convention \(t_J=t_k\) matches the score
evaluation in the exponential-integrator reverse sampler used below.

A single randomized-time denoising observation is
\[
    \xi
    =
    (X_0,Y,J,\varepsilon),
\]
with
\[
    (X_0,Y)\sim P_0,
    \qquad
    J\sim\pi_K,
    \qquad
    \varepsilon\sim\mathcal N(0,I_{d_x}),
\]
all sampled independently except for the dependence between \(X_0\) and \(Y\)
specified by \(P_0\). The corresponding noised state and denoising target are
\begin{align}
    X_{t_J}
    &=
    e^{-t_J}X_0+\sigma_{t_J}\varepsilon,
    \label{eq:randomized_noised_state}
    \\
    Z_J
    &:=
    \nabla_x\log q_{t_J}(X_{t_J}\mid X_0)
    =
    -\frac{X_{t_J}-e^{-t_J}X_0}{\sigma_{t_J}^2}.
    \label{eq:randomized_denoising_target}
\end{align}

\paragraph{Randomized-time population denoising objective.}
For \(\theta\in\Theta_M\), define
\begin{equation}
    \mathcal L_K(\theta)
    :=
    \mathbb E_{\xi}
    \left[
        \left\|
            s_\theta(X_{t_J},Y,t_J)-Z_J
        \right\|_2^2
    \right].
    \label{eq:discrete_dsm_loss_prelim}
\end{equation}
This is a population objective: no fixed empirical dataset is introduced.
The network-class dependence enters through the feasible set
\(\Theta_M\). We define its best achievable population denoising risk by
\begin{equation}
    \mathcal L_{K,M}^\star
    :=
    \inf_{\theta\in\Theta_M}\mathcal L_K(\theta).
    \label{eq:population_optimal_loss}
\end{equation}
The infimum in \eqref{eq:population_optimal_loss} is an analytical benchmark
only. The training algorithm is neither assumed to have access to it nor
assumed to return a parameter attaining it.

\paragraph{Denoising identity and approximation error.}
Although the denoising objective uses the kernel score \(Z_J\), the reverse
process requires the marginal conditional score \(s^\star\). They are related
by the conditional denoising identity
\begin{equation}
    s^\star(X_{t_J},t_J,Y)
    =
    \mathbb E
    \left[
        Z_J
        \,\middle|\,
        X_{t_J},Y,t_J
    \right].
    \label{eq:denoising_identity}
\end{equation}
Consequently, for every measurable score predictor \(s\),
\begin{align}
&\mathbb E
\left[
    \|s(X_{t_J},Y,t_J)-Z_J\|_2^2
\right]
\nonumber\\
&\qquad=
C_K
+
\mathbb E
\left[
    \|s(X_{t_J},Y,t_J)
      -s^\star(X_{t_J},t_J,Y)\|_2^2
\right],
\label{eq:denoising_projection_identity}
\end{align}
where
\[
    C_K
    :=
    \mathbb E
    \left[
        \|Z_J-s^\star(X_{t_J},t_J,Y)\|_2^2
    \right]
\]
does not depend on \(s\).

This motivates the randomized-time best-in-class score approximation error
\begin{equation}
    \mathcal A_K(M)
    :=
    \inf_{\theta\in\Theta_M}
    \mathbb E
    \left[
        \left\|
            s_\theta(X_{t_J},Y,t_J)
            -
            s^\star(X_{t_J},t_J,Y)
        \right\|_2^2
    \right].
    \label{eq:approximation_error_M}
\end{equation}
By \eqref{eq:denoising_projection_identity},
\begin{equation}
    \mathcal L_{K,M}^\star
    =
    C_K+\mathcal A_K(M).
    \label{eq:loss_approximation_identity}
\end{equation}
Importantly, \(\mathcal A_K(M)\) is an approximation property of the score
class; it does not involve the online training budget \(N\). Under the
intrinsic-dimensional and anisotropic regularity assumptions introduced below,
we will prove a bound of the form
\[
    \mathcal A_K(M)
    \le
    \widetilde{\mathcal O}
    \left(
        M^{-\eta_{\mathrm{app}}}
    \right),
\]
subject to the stated grid-resolution and localization conditions.

\paragraph{Online stochastic-gradient training.}
The training algorithm uses fresh randomized-time denoising observations.
At SGD iteration \(r=0,\ldots,R-1\), a mini-batch
\[
    \{\xi_{r,b}\}_{b=1}^{B}
\]
of \(B\) independent copies of \(\xi\) is drawn, and the mini-batch stochastic
gradient is
\[
    \bar g_r
    :=
    \frac{1}{B}
    \sum_{b=1}^{B}
    g(\theta_r;\xi_{r,b}),
\]
where \(g(\theta;\xi)\) denotes the stochastic gradient of the squared
denoising loss associated with one online sample. The update is
\begin{equation}
    \theta_{r+1}
    =
    \theta_r-\eta_r\bar g_r.
    \label{eq:online_sgd_update}
\end{equation}
The total online denoising-sample budget is therefore
\begin{equation}
    N=RB.
    \label{eq:online_sample_budget}
\end{equation}
We denote the final parameter and score by
\[
    \widehat\theta:=\theta_R,
    \qquad
    \widehat s:=s_{\widehat\theta}.
\]

For later use, define the finite-time population optimization error
\begin{equation}
    \mathcal E_{\mathrm{opt}}(N,M)
    :=
    \mathbb E_{\mathrm{SGD}}
    \left[
        \mathcal L_K(\widehat\theta)
        -
        \mathcal L_{K,M}^\star
    \right],
    \label{eq:optimization_error_NM}
\end{equation}
with the expectation taken over the randomness of the online stochastic-gradient
procedure. The localized optimization analysis assumes explicitly that the
unmodified SGD trajectory remains in the PL region throughout the analyzed run.
The notation \(\mathcal E_{\mathrm{opt}}(N,M)\) suppresses the chosen
factorization \(N=RB\), initialization, and step-size schedule. The general
bound retains \(R\) and \(B\) explicitly; only under the fixed-\(B\)
specialization does the leading rate depend on \(N\) alone.

Thus \(M\) controls the representation error of the score class, whereas
\(N\) controls how accurately online SGD optimizes the population denoising
objective within that class. These two quantities are kept separate throughout
the analysis.

\paragraph{Generated law.}
For a conditioning value \(y\), let
\(\widehat p_{t_0}(\cdot\mid y)\) denote the law produced by initializing the
discretized reverse process at
\(\mathcal N(0,I_{d_x})\), using the grid
\(\{t_k\}_{k=0}^K\), and replacing \(s^\star\) by the learned score
\(\widehat s\). Specifically, with \(h_k=\Delta_k\), the sampler uses the
exponential-integrator update
\begin{equation}
X_{t_{k-1}}
=
e^{h_k}X_{t_k}
+2(e^{h_k}-1)\widehat s(X_{t_k},Y,t_k)
+\sqrt{e^{2h_k}-1}\,\xi_k,
\qquad
\xi_k\sim\mathcal N(0,I_{d_x}),
\label{eq:reverse_exponential_integrator}
\end{equation}
for \(k=K,\ldots,1\). The primary goal of the analysis is to control
\[
    \mathbb E_{\mathrm{SGD}}\mathbb E_Y
    \left[
        \operatorname{TV}
        \left(
            p_{t_0}(\cdot\mid Y),
            \widehat p_{t_0}(\cdot\mid Y)
        \right)
    \right].
\]
A separate corollary adds the smoothing bias
\(\operatorname{TV}(p_0,p_{t_0})\) in order to obtain a guarantee relative to
the original conditional data law.

\paragraph{Notation summary.}
For convenience, the principal quantities used in the analysis are summarized
below.
\begin{center}
\small
\begin{minipage}{0.98\linewidth}
\begin{tabular}{cl}
\toprule
\textbf{Symbol} & \textbf{Meaning} \\
\midrule
\(d_x\) & ambient dimension of the response/data variable \(X_0\) \\
\(d_y\) & ambient dimension of the conditioning variable \(Y\) \\
\(d_x'\) & intrinsic response dimension under the structural assumption \\
\(d_y'\) & intrinsic conditioning dimension under the structural assumption \\
\(s^\star(x,t,y)\) & true conditional score, taking values in \(\mathbb R^{d_x}\) \\
\(M\) & score-class / approximation-complexity index \\
\(N\) & total number of online denoising samples used by SGD \\
\(R\) & number of stochastic-gradient iterations \\
\(B\) & mini-batch size, so that \(N=RB\) \\
\(K\) & number of reverse-time discretization intervals \\
\(\Delta_k\) & grid spacing \(t_k-t_{k-1}\) \\
\(H_K\) & total grid weight \(\sum_k\Delta_k=T-t_0\) \\
\(\mathcal A_K(M)\) & randomized-time score approximation error \\
\(\mathcal E_{\mathrm{opt}}(N,M)\) & finite-time population optimization error \\
\bottomrule
\end{tabular}
\end{minipage}
\end{center}

Algorithm~\ref{alg:conditional_ou_online} gives the corresponding
training-and-sampling procedure. The theoretical analysis below considers its
online fresh-sample form described above.

\section{Main Results}
\label{sec:main_results}

We now state the assumptions and main finite-time guarantees. A central
distinction throughout this section is between the approximation-complexity
index \(M\), which determines the richness of the score class
\(\mathcal F_{\Theta,M}\), and the online denoising-sample budget
\(N=RB\), which determines the amount of stochastic-gradient training.
The two quantities are not identified.

\subsection{Distributional and Geometric Assumptions}

\begin{assumption}[Distributional regularity and approximation-class compatibility]
\label{ass:distributional}

Let \((X_0,Y)\sim P_0\) take values in
\(\mathbb R^{d_x}\times\mathbb R^{d_y}\). The following conditions hold.

\begin{enumerate}[leftmargin=0pt,labelindent=0pt,labelwidth=!,
itemindent=6pt,align=left,itemsep=3pt,topsep=2pt,parsep=0pt]

    \item \textbf{Joint fibered intrinsic geometry.}
    There exists an embedded manifold
    \[
        \mathcal M
        \subset
        \mathbb R^{d_x}\times\mathbb R^{d_y}
    \]
    of dimension
    \[
        \dim(\mathcal M)=d_x'+d_y',
    \]
    containing the support of \(P_0\). Let
    \[
        \pi_Y:\mathcal M\longrightarrow \mathcal M_Y
    \]
    denote the restriction to \(\mathcal M\) of the projection onto the
    conditioning coordinate, where
    \(\mathcal M_Y\subset\mathbb R^{d_y}\) is an embedded
    \(d_y'\)-dimensional manifold. We assume that \(\pi_Y\) is a smooth
    submersion. Hence, for \(P_Y\)-a.e.\ \(y\), the fiber
    \[
        \mathcal M_y
        :=
        \pi_Y^{-1}(y)
    \]
    is a \(d_x'\)-dimensional submanifold describing the intrinsic support of
    \(X_0\mid Y=y\).

    On each localization region used below, the joint manifold admits a finite
    family of fiber-compatible charts of the form
    \[
        (u,v)
        \in
        \mathbb R^{d_x'}\times\mathbb R^{d_y'}
        \longmapsto
        \Phi(u,v)
        =
        \bigl(
            \Phi_X(u,v),\Phi_Y(v)
        \bigr),
    \]
    with uniformly controlled chart and inverse-chart Lipschitz constants,
    Jacobians, reach, condition number, and the encoder--decoder quantities
    required by the manifold-adapted approximation argument. The chart
    smoothness satisfies the hypotheses of the approximation theorem invoked
    below, and the geometric approximation remainder is no larger than the
    leading density-approximation term.

    \item \textbf{Anisotropic smooth conditional laws.}
    The conditioning law \(P_Y\) admits a density with respect to the intrinsic
    volume measure on \(\mathcal M_Y\). In the local coordinate \(v\), this
    density is locally H\"older smooth of order \(\alpha_y>0\).

    For \(P_Y\)-a.e.\ \(y\), the conditional law
    \(P_{X_0\mid Y=y}\) admits a density with respect to the induced volume
    measure on the fiber \(\mathcal M_y\), bounded by a constant
    \(B_{\mathrm{dens}}<\infty\). In a fiber-compatible chart, the resulting
    conditional density, viewed as a function of \((u,v)\), has
    response-coordinate H\"older smoothness \(\alpha_x>0\) in \(u\) and
    conditioning-coordinate H\"older smoothness \(\alpha_y>0\) in \(v\).
    The corresponding localized H\"older constants are uniformly controlled
    on the logarithmic-radius regions used in the approximation proof. The
    localized density nondegeneracy and inverse-density quantities required by
    the conditional score approximation construction are also controlled on
    those regions.

    \item \textbf{Sub-exponential tails.}
    There exist constants \(\lambda_Y,\lambda_X>0\) such that
    \[
        \mathbb E
        \exp\!\left(\lambda_Y\|Y\|_2\right)
        <\infty,
    \]
    and
    \[
        \operatorname*{ess\,sup}_{y\sim P_Y}
        \mathbb E
        \left[
            \exp\!\left(\lambda_X\|X_0\|_2\right)
            \,\middle|\,
            Y=y
        \right]
        <\infty.
    \]

    \item \textbf{Approximation-class compatibility.}
    On every logarithmic localization region used in the proof, the chart,
    decoder, density, and geometric-remainder conditions required by the
    blockwise conditional score-approximation theorem of
    \citet{tang2025conditional} hold with the growth recorded in the
    \(\widetilde{\mathcal O}\)-notation. The class
    \(\mathcal F_{\Theta,M}\) contains smooth-activation, smooth-gated
    realizations of those localized blockwise approximants at squared-error
    rate \(M^{-\eta_{\mathrm{app}}}\), with activation and gate smoothing
    affecting only logarithmic factors.

\end{enumerate}

Define the anisotropic approximation exponent
\begin{equation}
    \eta_{\mathrm{app}}
    :=
    \frac{2\alpha_x}
    {2\alpha_x+d_x'+(\alpha_x/\alpha_y)d_y'}.
    \label{eq:eta_app_main}
\end{equation}
\end{assumption}

The fibered formulation in Assumption~\ref{ass:distributional} makes precise
the joint low-dimensional structure used by the approximation argument.
In particular, the separate statements that \(Y\) is low dimensional and
\(X_0\mid Y=y\) is low dimensional are not by themselves sufficient to imply
a regular joint intrinsic geometry. The submersion and local
fiber-compatible-chart conditions make explicit the effective anisotropic
dimension
\[
    d_x'
    +
    \frac{\alpha_x}{\alpha_y}d_y'
\]
that appears in \eqref{eq:eta_app_main}.

The sub-exponential tail condition is used to reduce the approximation problem
to logarithmic-radius high-probability regions. It also implies the finite
conditional second-moment quantity
\begin{equation}
    m_2
    :=
    \operatorname*{ess\,sup}_{y\sim P_Y}
    \mathbb E
    \left[
        \|X_0\|_2^2
        \,\middle|\,
        Y=y
    \right]
    <\infty.
    \label{eq:conditional_second_moment}
\end{equation}

\subsection{Finite-Time Optimization Assumption}

\begin{assumption}[Localized PL and stochastic-gradient regularity]
\label{ass:local_population_pl}

Fix a score-class complexity \(M\) and time grid \(K\). Let
\(\mathcal L_K\) be the randomized-time population denoising objective in
\eqref{eq:discrete_dsm_loss_prelim}. Define its optimal value over the class by
\[
    \mathcal L_{K,M}^\star
    :=
    \inf_{\theta\in\Theta_M}
    \mathcal L_K(\theta).
\]
There exist a parameter region
\(\mathcal U_{K,M}\subseteq\Theta_M\) and constants
\[
    L_M<\infty,
    \qquad
    \mu_M>0,
    \qquad
    \sigma_M^2<\infty
\]
such that the following conditions hold.

The notation \(L_M,\mu_M,\sigma_M^2\) suppresses their possible dependence
on the current grid, cutoff, and horizon \((K,t_0,T)\). Every statement below
uses the constants associated with the particular randomized-time objective
\(\mathcal L_K\) under consideration.

\begin{enumerate}[leftmargin=0pt,labelindent=0pt,labelwidth=!,
itemindent=6pt,align=left,itemsep=3pt,topsep=2pt,parsep=0pt]

    \item \textbf{Local smoothness.}
    For all \(\theta,\theta'\in\mathcal U_{K,M}\),
    \[
        \mathcal L_K(\theta')
        \le
        \mathcal L_K(\theta)
        +
        \left\langle
            \nabla\mathcal L_K(\theta),
            \theta'-\theta
        \right\rangle
        +
        \frac{L_M}{2}
        \|\theta'-\theta\|_2^2.
    \]

    \item \textbf{Localized PL inequality.}
    For all \(\theta\in\mathcal U_{K,M}\),
    \begin{equation}
        \|\nabla\mathcal L_K(\theta)\|_2^2
        \ge
        2\mu_M
        \left(
            \mathcal L_K(\theta)
            -
            \mathcal L_{K,M}^\star
        \right).
        \label{eq:localized_PL_main}
    \end{equation}

    \item \textbf{Stochastic-gradient oracle.}
    For a single online denoising sample \(\xi\), the stochastic gradient
    \(g(\theta;\xi)\) satisfies, for every
    \(\theta\in\mathcal U_{K,M}\),
    \[
        \mathbb E
        \left[
            g(\theta;\xi)
            \,\middle|\,
            \theta
        \right]
        =
        \nabla\mathcal L_K(\theta),
    \]
    and
    \[
        \mathbb E
        \left[
            \|g(\theta;\xi)
            -
            \nabla\mathcal L_K(\theta)\|_2^2
            \,\middle|\,
            \theta
        \right]
        \le
        \sigma_M^2.
    \]
    For a mini-batch of size \(B\), the corresponding variance is bounded by
    \(\sigma_M^2/B\).

    \item \textbf{Trajectory containment.}
    The initialization satisfies
    \(\theta_0\in\mathcal U_{K,M}\), and the analyzed SGD run satisfies
    \[
        \theta_r\in\mathcal U_{K,M},
        \qquad
        r=0,\ldots,R.
    \]
    This is an explicit almost-sure condition on the unmodified SGD trajectory
    analyzed below. A projected or stopped variant would require its own
    descent analysis.

\end{enumerate}
\end{assumption}

Assumption~\ref{ass:local_population_pl} is an explicit optimization
regularity assumption. We do not derive the PL inequality
\eqref{eq:localized_PL_main} for the entire nonlinear score-network class.
Instead, the theorem quantifies the finite-time optimization error under this
localized geometry and trajectory-containment condition. In
particular, the constants \(L_M\), \(\mu_M\), and \(\sigma_M^2\) are retained
explicitly rather than being hidden inside the headline rate.

The optimization diagnostics reported in Appendix~\ref{sec:experiments} record
finite-run loss and parameter trajectories. They do not test the population PL
inequality, identify a region on which it holds, or establish
Assumption~\ref{ass:local_population_pl}.

\subsection{General Training-to-Sampling Bound}

\begin{theorem}[Finite-time total-variation decomposition]
\label{thm:tv_bound}

Let \(M\) be the score-class complexity, let \(N=RB\) be the total number of
online denoising samples used by SGD, and let
\(\widehat p_{t_0}(\cdot\mid y)\) be the law generated by the admissibly
discretized reverse sampler initialized from
\(\mathcal N(0,I_{d_x})\) and using
\(\widehat s=s_{\widehat\theta}\).

Assume \(0<t_0<1\le T\).

Suppose Assumptions~\ref{ass:distributional} and
\ref{ass:local_population_pl} hold. Let
\[
    \Delta_k:=t_k-t_{k-1},
    \qquad
    H_K:=\sum_{k=1}^{K}\Delta_k,
    \qquad
    \Delta_{\max}:=
    \max_{1\le k\le K}\Delta_k.
\]
Assume that the reverse-time grid is admissible in the sense of
Lemma~\ref{lem:forward_law_tv_stability}. Then
\begin{align}
&
\mathbb E_{\mathrm{SGD}}\mathbb E_Y
    \left[
        \operatorname{TV}
        \left(
            p_{t_0}(\cdot\mid Y),
            \widehat p_{t_0}(\cdot\mid Y)
        \right)
    \right]
\nonumber\\
&\qquad\le
C_{\mathrm{term}}e^{-T}
+
C_{\mathrm{disc}}
\sqrt{
    \frac{d_x+m_2}{K}
}
+
C_{\mathrm{stab}}
\sqrt{
    H_K
    \left[
        \mathcal A_K(M)
        +
        \mathcal E_{\mathrm{opt}}(N,M)
    \right]
}.
\label{eq:tv_general_decomposition}
\end{align}
Here \(C_{\mathrm{term}},C_{\mathrm{disc}}\), and
\(C_{\mathrm{stab}}\) depend only on fixed diffusion, moment, and grid
parameters. The notation for \(C_{\mathrm{disc}}\) absorbs the logarithmic
factors associated with the admissible grid; none of these factors has
polynomial dependence on \(M\), \(N\), or \(K\).
\end{theorem}

Theorem~\ref{thm:tv_bound} is the structural form of our main result.
It separates four sources of error:
\[
\underbrace{e^{-T}}_{\text{terminal initialization}},
\qquad
\underbrace{K^{-1/2}}_{\text{time discretization}},
\qquad
\underbrace{\sqrt{H_K\mathcal A_K(M)}}_{\text{approximation/localization}},
\qquad
\underbrace{
\sqrt{H_K\mathcal E_{\mathrm{opt}}(N,M)}
}_{\text{finite-time optimization}}.
\]
In particular, no relationship between \(M\) and \(N\) is imposed by the
decomposition itself.

\subsection{Instantiation of the Approximation and Optimization Terms}

The first ingredient needed to specialize
Theorem~\ref{thm:tv_bound} is the approximation bound. Under
Assumption~\ref{ass:distributional}, the smooth-gated dyadic-block construction
described in Section~\ref{sec:prelims_problem} and proved in
Appendix~\ref{subsec:proof_randomized_time_approx} yields
\begin{equation}
    \mathcal A_K(M)
    \le
    \widetilde C_{\mathrm{app}}
    \left(
        M^{-\eta_{\mathrm{app}}}
        +
        \Delta_{\max}
    \right),
    \label{eq:approximation_M_main}
\end{equation}
where
\[
    \eta_{\mathrm{app}}
    =
    \frac{2\alpha_x}
    {2\alpha_x+d_x'+(\alpha_x/\alpha_y)d_y'}.
\]
The logarithmic factors contained in
\(\widetilde C_{\mathrm{app}}\) arise from tail localization, the dyadic time
cover, and the localized chart and smoothness constants. Consequently, if
\begin{equation}
    \Delta_{\max}
    \le
    \widetilde{\mathcal O}
    \left(
        M^{-\eta_{\mathrm{app}}}
    \right),
    \label{eq:grid_fineness_M}
\end{equation}
then
\begin{equation}
    \mathcal A_K(M)
    \le
    \widetilde{\mathcal O}
    \left(
        M^{-\eta_{\mathrm{app}}}
    \right).
    \label{eq:approximation_M_simplified}
\end{equation}

The second ingredient is finite-time online optimization.

\begin{proposition}[Finite-time online SGD under localized PL geometry]
\label{prop:population_pl_opt_main}

Suppose Assumption~\ref{ass:local_population_pl} holds. Define the initial
population excess risk by
\[
    \Delta_{0,M}
    :=
    \mathcal L_K(\theta_0)
    -
    \mathcal L_{K,M}^\star,
\]
and run \(R\) SGD iterations with mini-batch size \(B\), so that
\(N=RB\). Choose
\begin{equation}
    \eta_r
    =
    \frac{2}{\mu_M(r+r_0)},
    \qquad
    r=0,\ldots,R-1,
    \label{eq:decreasing_stepsize_main}
\end{equation}
where
\[
    r_0
    \ge
    \max
    \left\{
        4,\,
        \frac{2L_M}{\mu_M}
    \right\}.
\]
Then there exist universal constants \(C_1,C_2>0\) such that
\begin{align}
    \mathcal E_{\mathrm{opt}}(N,M)
    &\le
        C_1
        \frac{
            \Delta_{0,M}r_0^2
        }{
            (R+r_0)^2
        }
        +
        C_2
        \frac{
            L_M\sigma_M^2
        }{
            \mu_M^2 B(R+r_0)
        }.
    \label{eq:explicit_PL_optimization}
\end{align}
\end{proposition}

Proposition~\ref{prop:population_pl_opt_main} makes explicit the quantities
suppressed in the earlier sample-complexity presentation. The first term in
\eqref{eq:explicit_PL_optimization} is the initialization transient; the second
is the stochastic-gradient contribution. The dependence on
\(L_M,\mu_M,\sigma_M^2\), and \(\Delta_{0,M}\) is retained because these
quantities can, in principle, change as the score class becomes more expressive.

Combining Theorem~\ref{thm:tv_bound},
\eqref{eq:approximation_M_main}, and
Proposition~\ref{prop:population_pl_opt_main} yields the explicit version of
the stopped-law guarantee.

\begin{corollary}[Explicit stopped-law bound]
\label{cor:explicit_stopped_tv}

Under the assumptions of Theorem~\ref{thm:tv_bound}, run online SGD with
\eqref{eq:decreasing_stepsize_main}. If the approximation construction satisfies
\eqref{eq:approximation_M_main}, then
\begin{align}
&
\mathbb E_{\mathrm{SGD}}\mathbb E_Y
    \left[
        \operatorname{TV}
        \left(
            p_{t_0}(\cdot\mid Y),
            \widehat p_{t_0}(\cdot\mid Y)
        \right)
    \right]
\nonumber\\
&\quad\le
C_{\mathrm{term}}e^{-T}
+
C_{\mathrm{disc}}
\sqrt{\frac{d_x+m_2}{K}}
\nonumber\\
&\qquad+
C_{\mathrm{stab}}
\sqrt{
H_K
\left[
\widetilde C_{\mathrm{app}}
\left(
M^{-\eta_{\mathrm{app}}}
+
\Delta_{\max}
\right)
+
C_1
\frac{\Delta_{0,M}r_0^2}{(R+r_0)^2}
+
C_2
\frac{L_M\sigma_M^2}
{\mu_M^2B(R+r_0)}
\right]
}.
\label{eq:tv_explicit_MN}
\end{align}
\end{corollary}

Equation~\eqref{eq:tv_explicit_MN} is the main quantitative
training-to-sampling statement. It does not require that the PL, smoothness,
or gradient-noise constants be uniform across network sizes, and therefore
makes explicit the tradeoff between increasing approximation capacity and
the resulting optimization geometry.

\subsection{Simplified Rate Under Uniform Optimization Geometry}

A simpler rate follows if the optimization constants remain controlled along
the chosen sequence of score classes.

\begin{corollary}[Simplified complexity for the stopped conditional law]
\label{cor:stopped_rate}

Suppose the assumptions of Corollary~\ref{cor:explicit_stopped_tv} hold and,
along a family of score classes, horizons, and admissible grids indexed by
\(\epsilon\downarrow0\), the constants of the corresponding objectives satisfy
\begin{equation}
    \sup_{\epsilon} L_{M(\epsilon)}
    \le
    \overline L
    <\infty,
    \qquad
    \inf_{\epsilon}\mu_{M(\epsilon)}
    \ge
    \underline\mu
    >0,
    \qquad
    \sup_{\epsilon}\sigma_{M(\epsilon)}^2
    \le
    \overline\sigma^2
    <\infty,
    \label{eq:uniform_opt_geometry}
\end{equation}
and
\[
    \sup_{\epsilon}\Delta_{0,M(\epsilon)}
    \le
    \overline\Delta_0
    <\infty.
\]
Here each displayed constant is evaluated for the grid and horizon paired with
\(M(\epsilon)\). In the stopped-law result, \(t_0\) is fixed; when
\eqref{eq:uniform_opt_geometry} is invoked for the original-law extension, the
same uniformity is additionally required along its varying cutoffs
\(t_0(\epsilon)\).
Fix \(t_0>0\) independently of \(\epsilon\). Take \(B\) to be a positive
constant independent of \(\epsilon\), and choose a fixed \(r_0\) satisfying
\[
    r_0
    \ge
    \max
    \left\{
        4,\,
        \frac{2\overline L}{\underline\mu}
    \right\}.
\]
Then
\begin{equation}
    \mathcal E_{\mathrm{opt}}(N,M)
    =
    \mathcal O(N^{-1}),
    \label{eq:opt_N_simple}
\end{equation}
where the initialization transient is of smaller order
\(\mathcal O(N^{-2})\).

Consequently, there exists \(\epsilon_0\in(0,1)\) such that, for every
\(0<\epsilon<\epsilon_0\), one may choose
\begin{equation}
    M
    =
    \widetilde{\Theta}
    \left(
        \epsilon^{-2/\eta_{\mathrm{app}}}
    \right)
    =
    \widetilde{\Theta}
    \left(
        \epsilon^{
        -\left(
            2+d_x'/\alpha_x+d_y'/\alpha_y
        \right)}
    \right),
    \label{eq:M_epsilon_stopped}
\end{equation}
\begin{equation}
    N
    =
    \widetilde{\Theta}
    \left(
        \epsilon^{-2}
    \right),
    \label{eq:N_epsilon_stopped}
\end{equation}
and
\[
    T
    =
    \Theta(\log(1/\epsilon)).
\]
Take an admissible grid satisfying
\[
    \Delta_{\max}
    \le
    \widetilde{\mathcal O}
    \left(
        M^{-\eta_{\mathrm{app}}}
    \right)
\]
and, up to logarithmic and fixed moment factors,
\begin{equation}
    K
    =
    \widetilde{\Theta}
    \left(
        (d_x+m_2)\epsilon^{-2}
    \right).
    \label{eq:K_epsilon_stopped}
\end{equation}
Then
\begin{equation}
\mathbb E_{\mathrm{SGD}}\mathbb E_Y
    \left[
        \operatorname{TV}
        \left(
            p_{t_0}(\cdot\mid Y),
            \widehat p_{t_0}(\cdot\mid Y)
        \right)
    \right]
\le
\widetilde{\mathcal O}(\epsilon).
\label{eq:stopped_final_eps}
\end{equation}
\end{corollary}

Corollary~\ref{cor:stopped_rate} should be interpreted with an important
qualification. The requirement
\[
    N
    =
    \widetilde{\Theta}(\epsilon^{-2})
\]
is the number of fresh online denoising samples consumed by SGD
\emph{under the uniform optimization-geometry condition}
\eqref{eq:uniform_opt_geometry}. Separately, the score class must have
approximation complexity
\[
    M
    =
    \widetilde{\Theta}
    \left(
        \epsilon^{-2/\eta_{\mathrm{app}}}
    \right).
\]
Thus the approximation requirement should not be interpreted as an additional
online sample-complexity term. If \(L_M\), \(\mu_M\), or \(\sigma_M^2\)
deteriorate with increasing \(M\), the appropriate guarantee is the explicit
bound \eqref{eq:tv_explicit_MN}, rather than the simplified
\(\widetilde{\mathcal O}(\epsilon^{-2})\) online-sample statement.

\subsection{Extension to the Original Conditional Data Law}

The preceding results deliberately concern the positive-time law
\(p_{t_0}(\cdot\mid y)\). This distinction is essential when the clean data are
supported on a strict lower-dimensional manifold. For every \(t_0>0\), the OU
forward process Gaussian-smooths the data law, so a singular
\(p_0(\cdot\mid y)\) need not approach \(p_{t_0}(\cdot\mid y)\) in ambient
total variation.

To obtain convergence to the original conditional law, we therefore impose the
stronger full-dimensional density condition in
Assumption~\ref{ass:distributional_full} and use the smoothing-bias result in
Appendix~\ref{app:tv_to_p0_extension}.

Define
\[
    a_\beta
    :=
    \max
    \left\{
        1,\frac{2}{\beta}
    \right\},
\]
and
\begin{equation}
    \rho_0
    :=
    \frac{2}
    {2\alpha_x+d_x+(\alpha_x/\alpha_y)d_y'},
    \qquad
    \zeta_0
    :=
    \frac{2\alpha_y}
    {2\alpha_y+d_y'},
    \qquad
    \gamma_0
    :=
    \frac{\alpha_y d_x}
    {2\alpha_y+d_y'}.
    \label{eq:p0_exponents}
\end{equation}

\begin{corollary}[TV bound for the original conditional law]
\label{cor:main_tv_to_p0}

Suppose Assumptions~\ref{ass:distributional_full} and
\ref{ass:local_population_pl} hold, and assume \(0<t_0<1\le T\). Define
\[
m_2
:=
\operatorname*{ess\,sup}_{y\sim P_Y}
\mathbb E\!\left[\|X_0\|_2^2\,\middle|\,Y=y\right]<\infty,
\]
which is finite by Assumption~\ref{ass:distributional_full}. Choose a class of
approximation complexity \(M\) satisfying the later-time-regime condition
\begin{equation}
    M^{-\rho_0}
    \le
    t_0,
    \label{eq:second_regime_M}
\end{equation}
and take an admissible grid satisfying
\begin{equation}
    \Delta_{\max}
    \le
    \widetilde{\mathcal O}
    \left(
        t_0^{-\gamma_0}M^{-\zeta_0}
    \right).
    \label{eq:p0_grid_fineness_M}
\end{equation}
Run online SGD with \(N=RB\) samples and the step-size schedule in
Proposition~\ref{prop:population_pl_opt_main}. Then
\begin{align}
&
\mathbb E_{\mathrm{SGD}}\mathbb E_Y
    \left[
        \operatorname{TV}
        \left(
            p_0(\cdot\mid Y),
            \widehat p_{t_0}(\cdot\mid Y)
        \right)
    \right]
\nonumber\\
&\quad\le
C_{\mathrm{term}}e^{-T}
+
C_{\mathrm{disc}}
\sqrt{\frac{d_x+m_2}{K}}
\nonumber\\
&\qquad+
C_{\mathrm{smooth}}
t_0^{(\beta\wedge2)/2}
+
C_{\mathrm{OU}}d_x t_0
\nonumber\\
&\qquad+
C_{\mathrm{stab}}
\sqrt{
    H_K
    \left[
        \widetilde C_{\mathrm{app},0}
        t_0^{-\gamma_0}M^{-\zeta_0}
        +
        \mathcal E_{\mathrm{opt}}(N,M)
    \right]
}.
\label{eq:p0_general_MN_bound}
\end{align}
\end{corollary}

Corollary~\ref{cor:main_tv_to_p0} exhibits the additional smoothing-bias terms
that do not occur in the stopped-law theorem. It also replaces the
lower-dimensional response approximation rate by the appropriate
full-dimensional later-time rate
\[
    t_0^{-\gamma_0}M^{-\zeta_0}.
\]
Accordingly, the original-law and stopped-law guarantees have different
regularity assumptions and different complexity requirements.

Under the uniform optimization conditions
\eqref{eq:uniform_opt_geometry}, there exists \(\epsilon_0\in(0,1)\) such that
the preceding expression admits the following simplified specialization for
every \(0<\epsilon<\epsilon_0\). Define
\begin{equation}
\Gamma_{\mathrm{app},0}
:=
\max
\left\{
    2+\frac{d_y'}{\alpha_y}
      +\frac{a_\beta d_x}{2},
    \;
    \frac{a_\beta}{2}
    \left(
        2\alpha_x+d_x
        +
        \frac{\alpha_x}{\alpha_y}d_y'
    \right)
\right\},
\label{eq:Gamma_app_0}
\end{equation}
and
\begin{equation}
    \Lambda_0
    :=
    \max
    \left\{
        2,\,
        \Gamma_{\mathrm{app},0}\zeta_0
        -
        a_\beta\gamma_0
    \right\}.
    \label{eq:Lambda_0_revised}
\end{equation}
For every \(0<\epsilon<\epsilon_0\), choose
\begin{equation}
    t_0
    =
    \widetilde{\Theta}
    \left(
        \epsilon^{a_\beta}
    \right),
    \qquad
    M
    =
    \widetilde{\Theta}
    \left(
        \epsilon^{-\Gamma_{\mathrm{app},0}}
    \right),
    \qquad
    N
    =
    \widetilde{\Theta}
    \left(
        \epsilon^{-2}
    \right),
    \label{eq:p0_MN_choices}
\end{equation}
together with
\[
    T
    =
    \Theta(\log(1/\epsilon))
\]
and an admissible grid satisfying, up to fixed dimension, moment, and
logarithmic factors,
\begin{equation}
    K
    =
    \widetilde{\Theta}
    \left(
        \epsilon^{-\Lambda_0}
    \right).
    \label{eq:p0_K_choice}
\end{equation}
Then
\begin{equation}
\mathbb E_{\mathrm{SGD}}\mathbb E_Y
    \left[
        \operatorname{TV}
        \left(
            p_0(\cdot\mid Y),
            \widehat p_{t_0}(\cdot\mid Y)
        \right)
    \right]
\le
\widetilde{\mathcal O}(\epsilon).
\label{eq:p0_final_eps}
\end{equation}

The complexity exponent
\(\Gamma_{\mathrm{app},0}\) in
\eqref{eq:Gamma_app_0} concerns the approximation class \(M\), whereas the
online stochastic-gradient sample requirement remains
\(\widetilde{\Theta}(\epsilon^{-2})\) under uniform optimization geometry.
This separation is the analogue, for the original-law extension, of the
distinction between \(M\) and \(N\) in
Corollary~\ref{cor:stopped_rate}.

\section{Conclusion}
\label{sec:conclusion}

We analyzed conditional diffusion trained by fresh-sample online denoising SGD.
The main result is a training-to-sampling decomposition for the stopped law
\(p_{t_0}(\cdot\mid y)\) that keeps terminal initialization, reverse-time
discretization, score-class approximation, and finite-time optimization
separate. In particular, \(M\) denotes approximation capacity, whereas
\(N=RB\) denotes the number of online denoising samples; these are distinct
resources. Under the stated anisotropic fibered-geometry and
approximation-class compatibility assumptions,
\(\mathcal A_K(M)=\widetilde{\mathcal O}(M^{-\eta_{\mathrm{app}}})\). Under
uniform localized optimization geometry, the separate sufficient scales are
\[
M=\widetilde{\Theta}\!\left(
\epsilon^{-(2+d_x'/\alpha_x+d_y'/\alpha_y)}
\right),
\qquad
N=\widetilde{\Theta}(\epsilon^{-2}),
\]
up to the stated grid and logarithmic factors.

The optimization guarantee is conditional on an explicit PL geometry and
trajectory-containment assumption; the paper does not derive that geometry for
general neural score networks. Likewise, the intrinsic-dimensional
approximation step adapts existing conditional approximation theory under the
class-containment and regularity conditions stated in the paper. Under stronger
full-dimensional Besov and weighted-gradient regularity, the smoothing bias can
also be controlled, yielding an extension to \(p_0(\cdot\mid y)\). Establishing
trajectory containment for concrete architectures and optimizers, and deriving
conditional optimization guarantees without a PL assumption, remain important
open problems.

\newpage
\bibliography{references}

\newpage
\appendix
\onecolumn
\begin{algorithm}[h]
\caption{Conditional Score Training and Sampling for the OU Conditional Diffusion Model}
\label{alg:conditional_ou_online}
\begin{algorithmic}[1]
\STATE \textbf{Input:} Online sampling access to \(P_0(dx,dy)\), time interval
\([t_0,T]\), admissible grid \(\{t_k\}_{k=0}^K\) with \(t_K=T\), complexity
\(M\), initial parameter \(\theta_0\in\Theta_M\), score network
\(s_\theta\in\mathcal F_{\Theta,M}\), mini-batch size
\(B\), number of training iterations \(R\), and decreasing step sizes
\(\eta_r=2/(\mu_M(r+r_0))\), where
\(r_0\ge \max\{4,2L_M/\mu_M\}\).
\STATE The online denoising-sample budget is \(N=RB\).
\STATE Set \(\Delta_k=t_k-t_{k-1}\), \(H_K=\sum_{k=1}^K\Delta_k\), and
\(\mathbb P(J=k)=\Delta_k/H_K\), for \(k=1,\ldots,K\).
\STATE Initialize \(\theta\leftarrow\theta_0\).

\STATE \textbf{Training: Conditional Denoising Score Matching}
\FOR{$r=0$ to $R-1$}
    \STATE Draw a fresh mini-batch
    \(\{(x_{0,i},y_i)\}_{i=1}^{B}\sim P_0\).
    \STATE Sample \(J_i\sim\pi_K\) and
    \(\varepsilon_i\sim\mathcal N(0,I_{d_x})\), independently.
    \STATE Set \(t_i=t_{J_i}\), \(\sigma_{t_i}^2=1-e^{-2t_i}\), and
    \(x_{t_i,i}=e^{-t_i}x_{0,i}+\sigma_{t_i}\varepsilon_i\).
    \STATE Set
    \(z_i=-(x_{t_i,i}-e^{-t_i}x_{0,i})/\sigma_{t_i}^2\).
    \STATE Compute
    \(\widehat{\mathcal L}_r(\theta)
    =B^{-1}\sum_{i=1}^{B}
    \|s_\theta(x_{t_i,i},y_i,t_i)-z_i\|_2^2\).
    \STATE Update
    \(\theta\leftarrow\theta-\eta_r\nabla_\theta\widehat{\mathcal L}_r(\theta)\).
\ENDFOR
\STATE Set \(\widehat\theta\leftarrow\theta\).

\STATE \textbf{Sampling: Conditional Generation}
\STATE Given \(y\), sample \(x_{t_K}\sim\mathcal N(0,I_{d_x})\).
\FOR{$k=K$ down to $1$}
    \STATE Set \(h_k=t_k-t_{k-1}\) and sample
    \(\xi_k\sim\mathcal N(0,I_{d_x})\).
    \STATE Evaluate \(\widehat s_k=s_{\widehat\theta}(x_{t_k},y,t_k)\).
    \STATE Set
    \(x_{t_{k-1}}=e^{h_k}x_{t_k}+2(e^{h_k}-1)\widehat s_k
    +\sqrt{e^{2h_k}-1}\,\xi_k\).
\ENDFOR
\STATE \textbf{Return} \(x_{t_0}\).
\end{algorithmic}
\end{algorithm}
\section{Related Work}
Diffusion and score-based generative models build on reverse-time stochastic
dynamics \citep{anderson1982reverse} and have been developed into modern
denoising diffusion and score-SDE frameworks \citep{sohl2015deep,ho2020denoising,
song2021score}. Recent theoretical work has studied how score-estimation
error propagates to sampling error, as well as discretization and convergence of
reverse diffusion samplers \citep{chen2023sampling,benton2024nearly,pmlr-v235-li24ad,li2024sharp,dou2024theory}. A parallel line
of work develops statistical guarantees for diffusion models, including score
approximation, minimax distribution estimation, low-dimensional adaptation, and
sample-complexity bounds for denoising score matching
\citep{chen2023score,oko2023diffusion,zhang2024minimax,wibisono2024optimal,
li2024adapting,huang2024denoisingdiffusionprobabilisticmodels,liang2025low}.
Conditional diffusion models have also been studied for guided generation and
conditional distribution estimation, including classifier-free guidance,
reward-directed conditional diffusion, and minimax or manifold-adaptive
conditional diffusion guarantees
\citep{batzolis2021conditional,yuan2023reward,
fu2024unveil,tang2025conditional,hu2025statistical,
li2025conditionalnonasymptotic}.

Algorithm-dependent diffusion training has been analyzed in several
unconditional settings. \citet{gaur2025improvedsamplecomplexitydiffusion}
derive a finite-time SGD sample-complexity bound without empirical-risk
minimizer access; \citet{han2024neural} analyze optimization and generalization
for a particular two-layer score parameterization trained by gradient descent;
and \citet{strasman2026nonasymptotic} study non-asymptotic SGD convergence for
weighted denoising score matching, including an overparameterized-network
regime. Our contribution is not the first finite-time optimization analysis for
diffusion models. It combines such an online population-optimization component
with conditional, joint-intrinsic score approximation and conditional
reverse-process stability, while keeping approximation capacity and online
training budget separate. This work is inspired by existing theoretical analyses such as \citep{pmlr-v202-gaur23a},  \citep{gaur2023global},\citep{gaur2024closinggapachievingglobal} as well as flow-matching \citep{gaur2025generativemodelingcontinuousflows} and diffusion models \citep{gaur2025improvedsamplecomplexitydiffusion,srikanth2026discrete}.
\section{Proofs of Intermediate Lemmas}
\label{app:proofs_of_lemmas}

This section proves the approximation, finite-time optimization, and
smoothing-bias ingredients used in the main results. Throughout, we keep the
score-class complexity \(M\) separate from the online denoising-sample budget
\(N=RB\).

\subsection{Randomized-Time Approximation Error}
\label{subsec:proof_randomized_time_approx}

\begin{lemma}[Randomized-time conditional score approximation]
\label{lem:approx_cond_concise_main}
Suppose Assumption~\ref{ass:distributional} holds, let \(M\ge2\), and let
\(\mathcal F_{\Theta,M}\) be the smooth-gated time-conditioned score class
described in Section~\ref{sec:prelims_problem}. Then
\begin{equation}
    \mathcal A_K(M)
    \le
    \widetilde{\mathcal O}
    \left(
        M^{-\eta_{\mathrm{app}}}
        +
        \Delta_{\max}
    \right),
    \label{eq:approx_lemma_M_general}
\end{equation}
where
\[
    \eta_{\mathrm{app}}
    =
    \frac{2\alpha_x}
    {2\alpha_x+d_x'+(\alpha_x/\alpha_y)d_y'}.
\]
Consequently, if
\begin{equation}
    \Delta_{\max}
    \le
    \widetilde{\mathcal O}
    \left(
        M^{-\eta_{\mathrm{app}}}
    \right),
    \label{eq:approx_lemma_grid_condition}
\end{equation}
then
\begin{equation}
    \mathcal A_K(M)
    \le
    \widetilde{\mathcal O}
    \left(
        M^{-\eta_{\mathrm{app}}}
    \right).
    \label{eq:approx_lemma_M_simplified}
\end{equation}
\end{lemma}

\begin{proof}
Recall from \eqref{eq:approximation_error_M} that
\[
    \mathcal A_K(M)
    :=
    \inf_{\theta\in\Theta_M}
    \mathbb E
    \left[
        \left\|
            s_\theta(X_{t_J},Y,t_J)
            -
            s^\star(X_{t_J},t_J,Y)
        \right\|_2^2
    \right].
\]
Unlike the previous formulation, the proof does not introduce or assume the
existence of a parameter attaining this infimum. It is sufficient to construct
one comparison predictor in \(\mathcal F_{\Theta,M}\).

\paragraph{Step 1: Localization.}
Let
\[
    R_M
    :=
    C_{\mathrm{loc}}\log M,
\]
where \(C_{\mathrm{loc}}>0\) will be chosen sufficiently large, and define
\begin{equation}
    \mathcal G_M
    :=
    \left\{
        \|Y\|_2\le R_M,\;
        \|X_0\|_2\le R_M,\;
        \|\varepsilon\|_2\le R_M
    \right\}.
    \label{eq:GM_definition}
\end{equation}
Here
\[
    X_{t_J}
    =
    e^{-t_J}X_0
    +
    \sqrt{1-e^{-2t_J}}\,\varepsilon,
    \qquad
    \varepsilon\sim\mathcal N(0,I_{d_x}).
\]

By Lemma~\ref{lem:randomized_time_tail_localization}, applied with localization
radius \(R_M\), the sub-exponential assumptions on \(X_0\) and \(Y\), together
with Gaussian concentration for \(\varepsilon\), imply that there exist
constants \(C_1,C_2,c_1,c_2>0\) such that
\begin{equation}
    \mathbb P(\mathcal G_M^c)
    \le
    C_1e^{-c_1R_M}
    +
    C_2e^{-c_2R_M^2}.
    \label{eq:GM_tail_probability}
\end{equation}
Moreover, on \(\mathcal G_M\),
\[
    \|X_{t_J}\|_2
    \le
    2R_M,
\]
and, because \(t_J\ge t_0>0\), the denoising target
\[
    Z_J
    :=
    \nabla_x\log q_{t_J}(X_{t_J}\mid X_0)
    =
    -\frac{
        X_{t_J}-e^{-t_J}X_0
    }{
        1-e^{-2t_J}
    }
\]
satisfies
\begin{equation}
    \|Z_J\|_2
    \le
    \frac{R_M}
    {\sqrt{1-e^{-2t_0}}}.
    \label{eq:localized_target_bound}
\end{equation}

Thus, after localization, the approximation problem is restricted to a compact
response--conditioning region whose radius grows only logarithmically with
\(M\).

\paragraph{Step 2: Approximation on the localized region.}
Apply Lemma~\ref{lem:localized_manifold_score_approx} with approximation scale
\(M\). Under the joint fibered intrinsic geometry, anisotropic smoothness, and
approximation-class compatibility in Assumption~\ref{ass:distributional},
there exists a comparison
network
\[
    \widetilde s_M
    \in
    \mathcal F_{\Theta,M}
\]
such that
\begin{align}
&
\mathbb E
\left[
    \left\|
        \widetilde s_M(X_{t_J},Y,t_J)
        -
        s^\star(X_{t_J},t_J,Y)
    \right\|_2^2
    \mathbf 1_{\mathcal G_M}
\right]
\nonumber\\
&\qquad\le
\widetilde C_{\mathrm{app}}
\left(
    M^{-\eta_{\mathrm{app}}}
    +
    \Delta_{\max}
\right).
\label{eq:localized_candidate_M}
\end{align}
The hidden logarithmic factors account for the localization radius
\(R_M=O(\log M)\), the dyadic time cover, the lower cutoff \(t_0\), and the
uniform chart and H\"older constants on the localized region.

The exponent
\[
    \eta_{\mathrm{app}}
    =
    \frac{2\alpha_x}
    {2\alpha_x+d_x'+(\alpha_x/\alpha_y)d_y'}
\]
arises from balancing approximation resolution in the response and
conditioning intrinsic coordinates. In particular, if \(h_x\) and \(h_y\)
denote the corresponding resolutions, the anisotropic balance
\[
    h_y
    \asymp
    h_x^{\alpha_x/\alpha_y}
\]
gives the effective intrinsic covering dimension
\[
    d_x'
    +
    \frac{\alpha_x}{\alpha_y}d_y',
\]
and hence the squared approximation rate appearing in
\eqref{eq:localized_candidate_M}.

\paragraph{Step 3: Control of the tail complement.}
It remains to bound the approximation error on \(\mathcal G_M^c\). The
bounded-parameter smooth score class used in the approximation construction has
at most polynomial output growth. Together with the sub-exponential moments of
\(X_0\) and \(Y\), Gaussian moments of \(\varepsilon\), and the positive lower
cutoff \(t_0\), Lemma~\ref{lem:randomized_time_tail_localization} gives
constants \(q_{\mathrm{mom}},q_{\mathrm{log}}\ge0\) such that
\begin{align}
&
\mathbb E
\left[
    \left\|
        \widetilde s_M(X_{t_J},Y,t_J)
        -
        s^\star(X_{t_J},t_J,Y)
    \right\|_2^4
\right]
\nonumber\\
&\qquad\le
C_{\mathrm{mom}}
M^{q_{\mathrm{mom}}}
(\log M)^{q_{\mathrm{log}}}.
\label{eq:candidate_fourth_moment_M}
\end{align}

By Cauchy--Schwarz,
\begin{align}
&
\mathbb E
\left[
    \left\|
        \widetilde s_M(X_{t_J},Y,t_J)
        -
        s^\star(X_{t_J},t_J,Y)
    \right\|_2^2
    \mathbf 1_{\mathcal G_M^c}
\right]
\nonumber\\
&\qquad\le
\left(
\mathbb E
\left[
    \left\|
        \widetilde s_M(X_{t_J},Y,t_J)
        -
        s^\star(X_{t_J},t_J,Y)
    \right\|_2^4
\right]
\right)^{1/2}
\mathbb P(\mathcal G_M^c)^{1/2}.
\label{eq:tail_CS_M}
\end{align}
Substituting \eqref{eq:GM_tail_probability} and
\eqref{eq:candidate_fourth_moment_M}, and using
\(R_M=C_{\mathrm{loc}}\log M\), gives
\[
\mathbb P(\mathcal G_M^c)
\le
C_1M^{-c_1C_{\mathrm{loc}}}
+
C_2
\exp\!\left[
    -c_2C_{\mathrm{loc}}^2(\log M)^2
\right].
\]
The second term is super-polynomially small. Since
\(q_{\mathrm{mom}}\) is fixed for the chosen approximation family, we may
choose \(C_{\mathrm{loc}}\) sufficiently large that the first term dominates
the polynomial moment envelope by an arbitrary prescribed polynomial power.
In particular, we may ensure
\begin{equation}
\mathbb E
\left[
    \left\|
        \widetilde s_M(X_{t_J},Y,t_J)
        -
        s^\star(X_{t_J},t_J,Y)
    \right\|_2^2
    \mathbf 1_{\mathcal G_M^c}
\right]
\le
\widetilde{\mathcal O}(M^{-2}).
\label{eq:tail_candidate_M}
\end{equation}
Since \(0<\eta_{\mathrm{app}}\le1\), this term is of smaller order than
\(M^{-\eta_{\mathrm{app}}}\).

\paragraph{Step 4: Pass from the candidate to the class infimum.}
Combining \eqref{eq:localized_candidate_M} and
\eqref{eq:tail_candidate_M} yields
\begin{align}
&
\mathbb E
\left[
    \left\|
        \widetilde s_M(X_{t_J},Y,t_J)
        -
        s^\star(X_{t_J},t_J,Y)
    \right\|_2^2
\right]
\nonumber\\
&\qquad\le
\widetilde{\mathcal O}
\left(
    M^{-\eta_{\mathrm{app}}}
    +
    \Delta_{\max}
\right).
\label{eq:full_candidate_M}
\end{align}
Since \(\widetilde s_M\in\mathcal F_{\Theta,M}\), the definition of
\(\mathcal A_K(M)\) immediately gives
\[
    \mathcal A_K(M)
    \le
    \mathbb E
    \left[
        \left\|
            \widetilde s_M(X_{t_J},Y,t_J)
            -
            s^\star(X_{t_J},t_J,Y)
        \right\|_2^2
    \right].
\]
Therefore
\[
    \mathcal A_K(M)
    \le
    \widetilde{\mathcal O}
    \left(
        M^{-\eta_{\mathrm{app}}}
        +
        \Delta_{\max}
    \right),
\]
which proves \eqref{eq:approx_lemma_M_general}. Under
\eqref{eq:approx_lemma_grid_condition}, the two terms have the same or smaller
order, giving
\[
    \mathcal A_K(M)
    \le
    \widetilde{\mathcal O}
    \left(
        M^{-\eta_{\mathrm{app}}}
    \right).
\]
\end{proof}

\begin{remark}[No risk-minimizer access is used]
\label{rem:no_minimizer_approx_proof}
Lemma~\ref{lem:approx_cond_concise_main} is formulated entirely through the
infimum \(\mathcal A_K(M)\). The proof constructs a single admissible comparison
network and upper-bounds the infimum by its error. Neither existence of an exact
best-in-class minimizer nor algorithmic access to such a minimizer is required.
\end{remark}

\subsection{Finite-Time Online Optimization}
\label{subsec:proof_finite_time_optimization}

We next prove Proposition~\ref{prop:population_pl_opt_main}. For mathematical
clarity, the localized optimization argument is stated for a training trajectory
that remains inside the PL region throughout the analyzed run. This avoids
conditioning the stochastic-gradient oracle on a future trajectory event. If
trajectory containment is instead available only with high probability, an
additional stopped-process or exit-time argument is required; such a conclusion
does not follow merely by conditioning the final bound on the future
localization event.

\begin{proof}[Proof of Proposition~\ref{prop:population_pl_opt_main}]
Fix \(M\) and \(K\), and write
\[
    F(\theta)
    :=
    \mathcal L_K(\theta),
    \qquad
    F^\star
    :=
    \mathcal L_{K,M}^\star,
\]
and
\begin{equation}
    \Delta_{0,M}
    :=
    F(\theta_0)-F^\star.
    \label{eq:Delta0M_proof}
\end{equation}
Note that \(F^\star\) is the infimum of the population objective over
\(\Theta_M\); attainment is not required.

At iteration \(r\), SGD uses a fresh mini-batch
\(\{\xi_{r,b}\}_{b=1}^B\) and forms
\[
    \bar g_r
    :=
    \frac{1}{B}
    \sum_{b=1}^B
    g(\theta_r;\xi_{r,b}),
\]
followed by
\begin{equation}
    \theta_{r+1}
    =
    \theta_r-\eta_r\bar g_r.
    \label{eq:sgd_update_proof_M}
\end{equation}
After \(R\) iterations, the total number of fresh online denoising samples is
\[
    N=RB.
\]

Let \(\mathscr F_r\) denote the sigma-field generated by the initialization and
all stochastic-gradient samples through iteration \(r-1\). Since the batch at
iteration \(r\) is fresh and independent of \(\mathscr F_r\), the
stochastic-gradient assumption gives
\begin{equation}
    \mathbb E
    \left[
        \bar g_r
        \,\middle|\,
        \mathscr F_r
    \right]
    =
    \nabla F(\theta_r)
    \label{eq:batch_unbiased_M}
\end{equation}
and
\begin{equation}
    \mathbb E
    \left[
        \|\bar g_r-\nabla F(\theta_r)\|_2^2
        \,\middle|\,
        \mathscr F_r
    \right]
    \le
    \frac{\sigma_M^2}{B}.
    \label{eq:batch_variance_M}
\end{equation}
Consequently,
\begin{align}
\mathbb E
\left[
    \|\bar g_r\|_2^2
    \,\middle|\,
    \mathscr F_r
\right]
&=
\|\nabla F(\theta_r)\|_2^2
+
\mathbb E
\left[
    \|\bar g_r-\nabla F(\theta_r)\|_2^2
    \,\middle|\,
    \mathscr F_r
\right]
\nonumber\\
&\le
\|\nabla F(\theta_r)\|_2^2
+
\frac{\sigma_M^2}{B}.
\label{eq:batch_second_moment_M}
\end{align}

We use the decreasing step-size sequence
\begin{equation}
    \eta_r
    =
    \frac{2}{\mu_M(r+r_0)},
    \qquad
    r=0,\ldots,R-1,
    \label{eq:stepsize_M_proof}
\end{equation}
where
\[
    r_0
    \ge
    \max
    \left\{
        4,\,
        \frac{2L_M}{\mu_M}
    \right\}.
\]
Hence
\[
    \eta_r
    \le
    \frac{2}{\mu_M r_0}
    \le
    \frac{1}{L_M}.
\]

Because the analyzed trajectory remains in
\(\mathcal U_{K,M}\), local \(L_M\)-smoothness gives
\begin{align}
F(\theta_{r+1})
&\le
F(\theta_r)
-
\eta_r
\left\langle
    \nabla F(\theta_r),\bar g_r
\right\rangle
+
\frac{L_M\eta_r^2}{2}
\|\bar g_r\|_2^2.
\label{eq:smooth_descent_M}
\end{align}
Taking conditional expectation given \(\mathscr F_r\) and applying
\eqref{eq:batch_unbiased_M}--\eqref{eq:batch_second_moment_M},
\begin{align}
\mathbb E
\left[
    F(\theta_{r+1})
    \,\middle|\,
    \mathscr F_r
\right]
&\le
F(\theta_r)
-
\eta_r
\|\nabla F(\theta_r)\|_2^2
\nonumber\\
&\quad+
\frac{L_M\eta_r^2}{2}
\left(
    \|\nabla F(\theta_r)\|_2^2
    +
    \frac{\sigma_M^2}{B}
\right)
\nonumber\\
&=
F(\theta_r)
-
\eta_r
\left(
    1-\frac{L_M\eta_r}{2}
\right)
\|\nabla F(\theta_r)\|_2^2
+
\frac{L_M\eta_r^2\sigma_M^2}{2B}.
\label{eq:one_step_M}
\end{align}
Since \(\eta_r\le1/L_M\),
\[
    1-\frac{L_M\eta_r}{2}
    \ge
    \frac12.
\]
Therefore
\begin{equation}
\mathbb E
\left[
    F(\theta_{r+1})
    \,\middle|\,
    \mathscr F_r
\right]
\le
F(\theta_r)
-
\frac{\eta_r}{2}
\|\nabla F(\theta_r)\|_2^2
+
\frac{L_M\eta_r^2\sigma_M^2}{2B}.
\label{eq:one_step_M_half}
\end{equation}

The localized PL inequality gives
\[
    \|\nabla F(\theta_r)\|_2^2
    \ge
    2\mu_M
    \left(
        F(\theta_r)-F^\star
    \right).
\]
Substituting into \eqref{eq:one_step_M_half} yields
\begin{align}
&
\mathbb E
\left[
    F(\theta_{r+1})-F^\star
    \,\middle|\,
    \mathscr F_r
\right]
\nonumber\\
&\qquad\le
\left(
    1-\mu_M\eta_r
\right)
\left(
    F(\theta_r)-F^\star
\right)
+
\frac{L_M\eta_r^2\sigma_M^2}{2B}.
\label{eq:PL_recursion_M}
\end{align}

Define
\[
    e_r
    :=
    \mathbb E_{\mathrm{SGD}}
    \left[
        F(\theta_r)-F^\star
    \right].
\]
Taking expectations in \eqref{eq:PL_recursion_M} and substituting
\eqref{eq:stepsize_M_proof},
\begin{equation}
    e_{r+1}
    \le
    \left(
        1-\frac{2}{r+r_0}
    \right)e_r
    +
    \frac{
        2L_M\sigma_M^2
    }{
        \mu_M^2B(r+r_0)^2
    }.
    \label{eq:scalar_PL_M}
\end{equation}
Set
\[
    A_M
    :=
    \frac{2L_M\sigma_M^2}{\mu_M^2B}.
\]
Then
\[
    e_{r+1}
    \le
    \left(
        1-\frac{2}{r+r_0}
    \right)e_r
    +
    \frac{A_M}{(r+r_0)^2}.
\]

Since \(r_0\ge4\), all contraction factors are nonnegative. Unrolling the
recursion gives
\begin{align}
e_R
&\le
\Delta_{0,M}
\prod_{j=0}^{R-1}
\left(
    1-\frac{2}{j+r_0}
\right)
\nonumber\\
&\quad+
\sum_{j=0}^{R-1}
\frac{A_M}{(j+r_0)^2}
\prod_{\ell=j+1}^{R-1}
\left(
    1-\frac{2}{\ell+r_0}
\right).
\label{eq:unroll_start_M}
\end{align}

For the initialization term,
\begin{align}
\prod_{j=0}^{R-1}
\left(
    1-\frac{2}{j+r_0}
\right)
&=
\prod_{j=0}^{R-1}
\frac{j+r_0-2}{j+r_0}
\nonumber\\
&=
\frac{
    (r_0-2)(r_0-1)
}{
    (R+r_0-2)(R+r_0-1)
}
\nonumber\\
&\le
\frac{r_0^2}{(R+r_0-2)^2}.
\label{eq:initial_product_M}
\end{align}
For \(j\le R-1\),
\begin{align}
\prod_{\ell=j+1}^{R-1}
\left(
    1-\frac{2}{\ell+r_0}
\right)
&=
\frac{
    (j+r_0-1)(j+r_0)
}{
    (R+r_0-2)(R+r_0-1)
}.
\label{eq:noise_product_M}
\end{align}
Hence
\begin{align}
e_R
&\le
\frac{
    r_0^2\Delta_{0,M}
}{
    (R+r_0-2)^2
}
\nonumber\\
&\quad+
\frac{A_M}{
    (R+r_0-2)(R+r_0-1)
}
\sum_{j=0}^{R-1}
\frac{j+r_0-1}{j+r_0}
\nonumber\\
&\le
\frac{
    r_0^2\Delta_{0,M}
}{
    (R+r_0-2)^2
}
+
\frac{
    A_M R
}{
    (R+r_0-2)(R+r_0-1)
}
\nonumber\\
&\le
\frac{
    r_0^2\Delta_{0,M}
}{
    (R+r_0-2)^2
}
+
\frac{
    A_M
}{
    R+r_0-2
}.
\label{eq:unrolled_M_final}
\end{align}
Absorbing the harmless shifts by \(2\) into universal constants
\(C_1,C_2>0\) gives
\begin{equation}
\mathbb E_{\mathrm{SGD}}
\left[
    \mathcal L_K(\theta_R)
    -
    \mathcal L_{K,M}^\star
\right]
\le
C_1
\frac{
    \Delta_{0,M}r_0^2
}{
    (R+r_0)^2
}
+
C_2
\frac{
    L_M\sigma_M^2
}{
    \mu_M^2B(R+r_0)
}.
\label{eq:optimization_explicit_M_final}
\end{equation}
Since \(\widehat\theta=\theta_R\), this is the desired finite-time
optimization bound.

Finally, \(N=RB\). If \(B\) and \(r_0\) are fixed and the quantities
\[
    L_M,\qquad
    \mu_M^{-1},\qquad
    \sigma_M^2,\qquad
    \Delta_{0,M}
\]
remain uniformly bounded over the chosen sequence of score classes, then
\(R\asymp N\), and
\[
    \mathbb E_{\mathrm{SGD}}
    \left[
        \mathcal L_K(\widehat\theta)
        -
        \mathcal L_{K,M}^\star
    \right]
    =
    \mathcal O(N^{-1}),
\]
with an initialization transient of order \(N^{-2}\). This simplified rate is
therefore a consequence of uniform optimization geometry; it is not assumed in
the general finite-time bound.
\end{proof}

\begin{remark}[Trajectory localization]
\label{rem:trajectory_localization_rigor}
The preceding proof uses stochastic-gradient unbiasedness with respect to the
natural filtration \(\mathscr F_r\). If one conditions instead on a future event
of the form
\[
    \mathcal E
    =
    \{
        \theta_0,\ldots,\theta_R\in\mathcal U_{K,M}
    \},
\]
the gradient noise need not remain conditionally unbiased after that
conditioning. Thus a high-probability trajectory-localization formulation
requires an additional exit-time or stopped-process argument, or an
algorithmically enforced localization mechanism. The factor
\((1-\delta)^{-1}\) alone does not establish this property. For
the main theorem we therefore use the trajectory-contained form of
Assumption~\ref{ass:local_population_pl}.
\end{remark}

\subsection{Smoothing Bias for the Original Data Law}
\label{subsec:smoothing_bias_proof}

The preceding approximation and optimization results control the learned
sampler relative to the positive-time law \(p_{t_0}\). To extend the guarantee
to \(p_0\), we need to quantify the error introduced by forward Gaussian
smoothing. Such a statement requires full-dimensional density regularity.

\begin{lemma}[Smoothing bias under integral density regularity]
\label{lem:tv_smoothing_integral}
Suppose Assumption~\ref{ass:distributional_full} holds. In particular, for
\(P_Y\)-a.e.\ \(y\), the conditional law \(p_0(\cdot\mid y)\) admits a
Lebesgue density on \(\mathbb R^{d_x}\) satisfying the uniform
\(B^{\beta\wedge2}_{1,\infty}\) and weighted-gradient bounds stated there.

Let \(p_t(\cdot\mid y)\) denote the conditional law of
\[
    X_t
    =
    \alpha_t X_0+\sigma_t W,
    \qquad
    \alpha_t=e^{-t},
    \qquad
    \sigma_t^2=1-e^{-2t},
    \qquad
    W\sim\mathcal N(0,I_{d_x}).
\]
Then, for sufficiently small \(t>0\),
\begin{align}
\mathbb E_Y
\left[
    \operatorname{TV}
    \left(
        p_0(\cdot\mid Y),
        p_t(\cdot\mid Y)
    \right)
\right]
&\le
C_{\mathrm{sm}}t^{(\beta\wedge2)/2}
+C_{\mathrm{OU}}d_x t.
\label{eq:smoothing_bias_integral}
\end{align}
The exponent saturates at order \(t\) when \(\beta\ge2\). The constants are
uniform in \(y\) and depend only on the integral regularity and dimension
parameters in Assumption~\ref{ass:distributional_full}.
\end{lemma}

\begin{proof}
Fix \(y\) in the full-\(P_Y\)-measure set from
Assumption~\ref{ass:distributional_full}, write
\(p=p_0(\cdot\mid y)\), and set
\[
s:=\beta\wedge2,
\qquad
\alpha_t=e^{-t},
\qquad
\tau_t:=\frac{\sigma_t}{\alpha_t}=\sqrt{e^{2t}-1}.
\]
If \(\varphi_\tau\) denotes the centered Gaussian density with covariance
\(\tau^2I_{d_x}\), then the density of \(X_t\) is
\[
p_t(x)
=
\alpha_t^{-d_x}
\bigl(p*\varphi_{\tau_t}\bigr)(x/\alpha_t).
\]
Let \(D_a f(x):=a^{-d_x}f(x/a)\). Since \(D_a\) is an \(L^1\)-isometry,
\begin{align}
\|p_t-p\|_1
&\le
\|D_{\alpha_t}(p*\varphi_{\tau_t})-D_{\alpha_t}p\|_1
+\|D_{\alpha_t}p-p\|_1
\nonumber\\
&=
\|p*\varphi_{\tau_t}-p\|_1
+\|D_{\alpha_t}p-p\|_1.
\label{eq:smoothing_semigroup_split}
\end{align}

The standard Gaussian mollification characterization of
\(B^s_{1,\infty}\), valid for \(0<s\le2\), gives
\[
\|p*\varphi_{\tau_t}-p\|_1
\le
C_s\tau_t^s\|p\|_{B^s_{1,\infty}}
\le
C_s' t^{s/2},
\]
for sufficiently small \(t\), because
\(\tau_t^2=e^{2t}-1\asymp t\).

It remains to control the OU rescaling. For \(a\in[\alpha_t,1]\), weak
differentiation and a change of variables give
\begin{align}
\left\|\frac{d}{da}D_a p\right\|_1
&\le
\frac{1}{a}
\left(
d_x\|p\|_1
+\int_{\mathbb R^{d_x}}
\|u\|_2\|\nabla p(u)\|_2\,du
\right).
\end{align}
Integrating from \(a=\alpha_t\) to \(a=1\), using
\(-\log\alpha_t=t\), and applying the uniform weighted-gradient assumption,
we obtain
\[
\|D_{\alpha_t}p-p\|_1
\le
C_{\mathrm{OU}}d_x t.
\label{eq:ou_scaling_L1}
\]
Combining the last two estimates with
\eqref{eq:smoothing_semigroup_split} and using that total variation is half
the \(L^1\) distance proves
\[
\operatorname{TV}\bigl(p_0(\cdot\mid y),p_t(\cdot\mid y)\bigr)
\le
C_{\mathrm{sm}}t^{(\beta\wedge2)/2}
+C_{\mathrm{OU}}d_x t.
\]
The constants are uniform in \(y\); averaging over \(Y\) proves
\eqref{eq:smoothing_bias_integral}.
\end{proof}

\begin{remark}[Why the smoothing lemma is separate from the stopped-law theorem]
\label{rem:smoothing_separate}
Lemma~\ref{lem:tv_smoothing_integral} is used only for the extension from
\(p_{t_0}(\cdot\mid y)\) to \(p_0(\cdot\mid y)\). It is not required for the
primary stopped-law result. In particular, the exact low-dimensional support
allowed by Assumption~\ref{ass:distributional} is compatible with the
stopped-law theorem but generally incompatible with vanishing ambient total
variation between \(p_0\) and \(p_{t_0}\). This is why the original-law
extension uses the separate full-dimensional
Assumption~\ref{ass:distributional_full}.
\end{remark}

\section{Obtaining the Final Result}
\label{app:obtaining_final_result}

We now combine the reverse-process stability, score approximation, and
finite-time optimization bounds to prove Theorem~\ref{thm:tv_bound} and its
simplified complexity specialization. The key point is that the approximation
complexity \(M\) and the online denoising-sample budget \(N=RB\) remain
separate throughout the argument.

Recall
\[
    \Delta_k:=t_k-t_{k-1},
    \qquad
    H_K:=\sum_{k=1}^{K}\Delta_k=T-t_0,
    \qquad
    \Delta_{\max}
    :=
    \max_{1\le k\le K}\Delta_k,
\]
and
\[
    \mathbb P(J=k)
    =
    \frac{\Delta_k}{H_K}.
\]
The randomized-time population denoising objective is
\begin{equation}
    \mathcal L_K(\theta)
    :=
    \mathbb E
    \left[
        \left\|
            s_\theta(X_{t_J},Y,t_J)
            -
            Z_J
        \right\|_2^2
    \right],
    \qquad
    Z_J
    :=
    \nabla_x\log q_{t_J}(X_{t_J}\mid X_0),
    \label{eq:final_LK_recall}
\end{equation}
and, for a score class of complexity \(M\),
\[
    \mathcal L_{K,M}^\star
    :=
    \inf_{\theta\in\Theta_M}
    \mathcal L_K(\theta).
\]
The learned score
\[
    \widehat s
    =
    s_{\widehat\theta}
\]
is the output of \(R\) online SGD iterations with mini-batch size \(B\), so
that
\[
    N=RB.
\]

We also recall the randomized-time best-in-class approximation error
\begin{equation}
    \mathcal A_K(M)
    :=
    \inf_{\theta\in\Theta_M}
    \mathbb E
    \left[
        \left\|
            s_\theta(X_{t_J},Y,t_J)
            -
            s^\star(X_{t_J},t_J,Y)
        \right\|_2^2
    \right],
    \label{eq:final_AK_recall}
\end{equation}
and the finite-time population optimization error
\begin{equation}
    \mathcal E_{\mathrm{opt}}(N,M)
    :=
    \mathbb E_{\mathrm{SGD}}
    \left[
        \mathcal L_K(\widehat\theta)
        -
        \mathcal L_{K,M}^\star
    \right].
    \label{eq:final_Eopt_recall}
\end{equation}
Under the trajectory-contained localized PL formulation used in
Proposition~\ref{prop:population_pl_opt_main}, the expectation in
\eqref{eq:final_Eopt_recall} is taken over the ordinary online SGD randomness
while the analyzed trajectory remains in the stated PL region.

\subsection{Proof of the General Training-to-Sampling Bound}

\begin{proof}[Proof of Theorem~\ref{thm:tv_bound}]

\noindent\textbf{Step 1: Introduce the intermediate reverse law.}

For each fixed conditioning value \(y\), let
\(p_{t_0}(\cdot\mid y)\) denote the forward-smoothed target law at time
\(t_0\). Equivalently, this is the law obtained by running the exact reverse SDE
from \(p_T(\cdot\mid y)\) down to time \(t_0\) using
\[
    s^\star(x,t,y)
    =
    \nabla_x\log p_t(x\mid y).
\]

Let
\[
    \widetilde p_{t_0}(\cdot\mid y)
\]
denote the law of the admissibly discretized reverse process initialized from
the exact terminal law \(p_T(\cdot\mid y)\) and using the learned score
\(\widehat s\). Finally, let
\[
    \widehat p_{t_0}(\cdot\mid y)
\]
denote the same learned-score discretized reverse process initialized instead
from the Gaussian reference law
\(\mathcal N(0,I_{d_x})\).

The triangle inequality gives
\begin{align}
&
\operatorname{TV}
\left(
    p_{t_0}(\cdot\mid y),
    \widehat p_{t_0}(\cdot\mid y)
\right)
\nonumber\\
&\qquad\le
\operatorname{TV}
\left(
    p_{t_0}(\cdot\mid y),
    \widetilde p_{t_0}(\cdot\mid y)
\right)
+
\operatorname{TV}
\left(
    \widetilde p_{t_0}(\cdot\mid y),
    \widehat p_{t_0}(\cdot\mid y)
\right).
\label{eq:final_tv_triangle}
\end{align}

The two terms in \eqref{eq:final_tv_triangle} represent, respectively, the
learned-score/discretization error and the terminal-initialization error.

\medskip
\noindent\textbf{Step 2: Terminal-initialization error.}

For a fixed realization of \(\widehat s\), the laws
\(\widetilde p_{t_0}(\cdot\mid y)\) and
\(\widehat p_{t_0}(\cdot\mid y)\) are obtained by applying the same
discretized reverse Markov kernel to two different initial laws,
\(p_T(\cdot\mid y)\) and \(\mathcal N(0,I_{d_x})\). Total variation contracts
under a Markov kernel. Therefore
\begin{align}
&
\operatorname{TV}
\left(
    \widetilde p_{t_0}(\cdot\mid y),
    \widehat p_{t_0}(\cdot\mid y)
\right)
\nonumber\\
&\qquad\le
\operatorname{TV}
\left(
    p_T(\cdot\mid y),
    \mathcal N(0,I_{d_x})
\right).
\label{eq:terminal_markov_contraction}
\end{align}
Let \(\gamma=\mathcal N(0,I_{d_x})\). Conditional on \(X_0=x\), the terminal
law is
\[
\mathcal N\!\left(e^{-T}x,(1-e^{-2T})I_{d_x}\right).
\]
Convexity of relative entropy in its first argument and the Gaussian
relative-entropy formula therefore give
\begin{align}
\mathrm{KL}\!\left(p_T(\cdot\mid y)\,\middle\|\,\gamma\right)
&\le
\frac12e^{-2T}
\mathbb E\!\left[\|X_0\|_2^2\,\middle|\,Y=y\right]
\nonumber\\
&\quad+
\frac{d_x}{2}
\left[-e^{-2T}-\log(1-e^{-2T})\right]
\nonumber\\
&\le C(d_x+m_2)e^{-2T},
\label{eq:terminal_ou_kl}
\end{align}
where the last inequality uses \(T\ge1\). Pinsker's inequality and averaging
over \(Y\) give
\begin{equation}
    \mathbb E_Y
    \left[
        \operatorname{TV}
        \left(
            \widetilde p_{t_0}(\cdot\mid Y),
            \widehat p_{t_0}(\cdot\mid Y)
        \right)
    \right]
    \le
    C_{\mathrm{term}}e^{-T}.
    \label{eq:terminal_error_final}
\end{equation}

\medskip
\noindent\textbf{Step 3: Reverse-process stability for the learned score.}

Apply Lemma~\ref{lem:forward_law_tv_stability} with
\(\bar s=\widehat s\). Since the corresponding discretized law initialized
from \(p_T(\cdot\mid y)\) is precisely
\(\widetilde p_{t_0}(\cdot\mid y)\), the lemma gives, for every fixed
realization of \(\widehat s\),
\begin{align}
&
\mathbb E_Y
\left[
    \operatorname{TV}
    \left(
        p_{t_0}(\cdot\mid Y),
        \widetilde p_{t_0}(\cdot\mid Y)
    \right)
\right]
\nonumber\\
&\qquad\le
C_{\mathrm{stab}}
\left(
    \sum_{k=1}^{K}
    \Delta_k
    \mathbb E
    \left[
        \left\|
            \widehat s(X_{t_k},Y,t_k)
            -
            s^\star(X_{t_k},t_k,Y)
        \right\|_2^2
    \right]
\right)^{1/2}
\nonumber\\
&\qquad\quad+
C_{\mathrm{disc}}
\sqrt{
    \frac{d_x+m_2}{K}
}.
\label{eq:final_reverse_stability}
\end{align}
Here
\[
    m_2
    =
    \operatorname*{ess\,sup}_{y\sim P_Y}
    \mathbb E
    \left[
        \|X_0\|_2^2
        \,\middle|\,
        Y=y
    \right]
    <\infty
\]
by Assumption~\ref{ass:distributional}.

The score error in \eqref{eq:final_reverse_stability} is evaluated under the
forward noised marginals
\[
    Y\sim P_Y,
    \qquad
    X_{t_k}\sim p_{t_k}(\cdot\mid Y),
\]
which is exactly the distribution entering the randomized-time population
denoising objective.

Define the randomized-time score error of the learned network by
\begin{equation}
    \widehat{\mathcal A}_K
    :=
    \mathbb E
    \left[
        \left\|
            \widehat s(X_{t_J},Y,t_J)
            -
            s^\star(X_{t_J},t_J,Y)
        \right\|_2^2
    \right].
    \label{eq:learned_score_error_final}
\end{equation}
Since
\[
    \mathbb P(J=k)
    =
    \frac{\Delta_k}{H_K},
\]
we have the exact identity
\begin{align}
&H_K\widehat{\mathcal A}_K
\nonumber\\
&\qquad=
\sum_{k=1}^{K}
\Delta_k
\mathbb E
\left[
    \left\|
        \widehat s(X_{t_k},Y,t_k)
        -
        s^\star(X_{t_k},t_k,Y)
    \right\|_2^2
\right].
\label{eq:grid_weight_score_identity}
\end{align}
Hence \eqref{eq:final_reverse_stability} becomes
\begin{align}
&
\mathbb E_Y
\left[
    \operatorname{TV}
    \left(
        p_{t_0}(\cdot\mid Y),
        \widetilde p_{t_0}(\cdot\mid Y)
    \right)
\right]
\nonumber\\
&\qquad\le
C_{\mathrm{stab}}
\sqrt{
    H_K\widehat{\mathcal A}_K
}
+
C_{\mathrm{disc}}
\sqrt{
    \frac{d_x+m_2}{K}
}.
\label{eq:stability_Ahat}
\end{align}

\medskip
\noindent\textbf{Step 4: Decompose learned-score error into approximation and optimization.}

The denoising identity gives
\[
    s^\star(X_{t_J},t_J,Y)
    =
    \mathbb E
    \left[
        Z_J
        \,\middle|\,
        X_{t_J},Y,t_J
    \right].
\]
Therefore, for every \(\theta\in\Theta_M\), the orthogonal projection identity
for conditional expectations yields
\begin{align}
\mathcal L_K(\theta)
&=
C_K
+
\mathbb E
\left[
    \left\|
        s_\theta(X_{t_J},Y,t_J)
        -
        s^\star(X_{t_J},t_J,Y)
    \right\|_2^2
\right],
\label{eq:final_denoising_projection}
\end{align}
where
\[
    C_K
    :=
    \mathbb E
    \left[
        \left\|
            Z_J
            -
            s^\star(X_{t_J},t_J,Y)
        \right\|_2^2
    \right]
\]
does not depend on \(\theta\).

Taking the infimum over \(\theta\in\Theta_M\) in
\eqref{eq:final_denoising_projection} gives
\begin{equation}
    \mathcal L_{K,M}^\star
    =
    C_K+\mathcal A_K(M).
    \label{eq:final_Lstar_identity}
\end{equation}
No minimizer attaining either infimum is required.

Applying \eqref{eq:final_denoising_projection} at
\(\widehat\theta\) and then using
\eqref{eq:final_Lstar_identity},
\begin{align}
\widehat{\mathcal A}_K
&=
\mathcal L_K(\widehat\theta)-C_K
\nonumber\\
&=
\left[
    \mathcal L_K(\widehat\theta)
    -
    \mathcal L_{K,M}^\star
\right]
+
\mathcal A_K(M).
\label{eq:Ahat_exact_decomp}
\end{align}
Taking expectation over the online SGD randomness yields
\begin{equation}
    \mathbb E_{\mathrm{SGD}}
    \left[
        \widehat{\mathcal A}_K
    \right]
    =
    \mathcal A_K(M)
    +
    \mathcal E_{\mathrm{opt}}(N,M).
    \label{eq:Ahat_expectation_decomp}
\end{equation}

This identity is the point at which approximation and finite-time optimization
enter the sampling analysis, and the two quantities remain separate.

\medskip
\noindent\textbf{Step 5: Average over the SGD randomness.}

Taking expectation over SGD in \eqref{eq:stability_Ahat} and applying Jensen's
inequality to the concave square-root function gives
\begin{align}
&
\mathbb E_{\mathrm{SGD}}
\mathbb E_Y
\left[
    \operatorname{TV}
    \left(
        p_{t_0}(\cdot\mid Y),
        \widetilde p_{t_0}(\cdot\mid Y)
    \right)
\right]
\nonumber\\
&\qquad\le
C_{\mathrm{stab}}
\sqrt{
    H_K
    \mathbb E_{\mathrm{SGD}}
    \left[
        \widehat{\mathcal A}_K
    \right]
}
+
C_{\mathrm{disc}}
\sqrt{
    \frac{d_x+m_2}{K}
}
\nonumber\\
&\qquad=
C_{\mathrm{stab}}
\sqrt{
    H_K
    \left[
        \mathcal A_K(M)
        +
        \mathcal E_{\mathrm{opt}}(N,M)
    \right]
}
+
C_{\mathrm{disc}}
\sqrt{
    \frac{d_x+m_2}{K}
}.
\label{eq:learned_sampling_general}
\end{align}

Combining \eqref{eq:final_tv_triangle},
\eqref{eq:terminal_error_final}, and
\eqref{eq:learned_sampling_general} therefore gives
\begin{align}
&
\mathbb E_{\mathrm{SGD}}
\mathbb E_Y
\left[
    \operatorname{TV}
    \left(
        p_{t_0}(\cdot\mid Y),
        \widehat p_{t_0}(\cdot\mid Y)
    \right)
\right]
\nonumber\\
&\qquad\le
C_{\mathrm{term}}e^{-T}
+
C_{\mathrm{disc}}
\sqrt{
    \frac{d_x+m_2}{K}
}
\nonumber\\
&\qquad\quad+
C_{\mathrm{stab}}
\sqrt{
    H_K
    \left[
        \mathcal A_K(M)
        +
        \mathcal E_{\mathrm{opt}}(N,M)
    \right]
}.
\label{eq:final_general_training_sampling}
\end{align}
This proves the general decomposition asserted in
Theorem~\ref{thm:tv_bound}.
\end{proof}

\subsection{Explicit Approximation and Optimization Instantiation}
\label{subsec:final_explicit_instantiation}

We next substitute the approximation and finite-time optimization bounds into
\eqref{eq:final_general_training_sampling}.

By Lemma~\ref{lem:approx_cond_concise_main},
\begin{equation}
    \mathcal A_K(M)
    \le
    \widetilde C_{\mathrm{app}}
    \left(
        M^{-\eta_{\mathrm{app}}}
        +
        \Delta_{\max}
    \right),
    \label{eq:final_AKM_bound}
\end{equation}
where
\[
    \eta_{\mathrm{app}}
    =
    \frac{2\alpha_x}
    {2\alpha_x+d_x'+(\alpha_x/\alpha_y)d_y'}.
\]

By Proposition~\ref{prop:population_pl_opt_main}, if
\[
    \eta_r
    =
    \frac{2}{\mu_M(r+r_0)},
    \qquad
    r_0
    \ge
    \max
    \left\{
        4,\frac{2L_M}{\mu_M}
    \right\},
\]
then
\begin{align}
\mathcal E_{\mathrm{opt}}(N,M)
&\le
C_1
\frac{
    \Delta_{0,M}r_0^2
}{
    (R+r_0)^2
}
+
C_2
\frac{
    L_M\sigma_M^2
}{
    \mu_M^2B(R+r_0)
},
\label{eq:final_Eopt_explicit}
\end{align}
where
\[
    N=RB,
    \qquad
    \Delta_{0,M}
    :=
    \mathcal L_K(\theta_0)
    -
    \mathcal L_{K,M}^\star.
\]

Substituting \eqref{eq:final_AKM_bound} and
\eqref{eq:final_Eopt_explicit} into
\eqref{eq:final_general_training_sampling} yields
\begin{align}
&
\mathbb E_{\mathrm{SGD}}
\mathbb E_Y
\left[
    \operatorname{TV}
    \left(
        p_{t_0}(\cdot\mid Y),
        \widehat p_{t_0}(\cdot\mid Y)
    \right)
\right]
\nonumber\\
&\quad\le
C_{\mathrm{term}}e^{-T}
+
C_{\mathrm{disc}}
\sqrt{
    \frac{d_x+m_2}{K}
}
\nonumber\\
&\qquad+
C_{\mathrm{stab}}
\Bigg\{
H_K
\Bigg[
\widetilde C_{\mathrm{app}}
\left(
    M^{-\eta_{\mathrm{app}}}
    +
    \Delta_{\max}
\right)
\nonumber\\
&\hspace{44mm}
+
C_1
\frac{
    \Delta_{0,M}r_0^2
}{
    (R+r_0)^2
}
+
C_2
\frac{
    L_M\sigma_M^2
}{
    \mu_M^2B(R+r_0)
}
\Bigg]
\Bigg\}^{1/2}.
\label{eq:final_explicit_MN_bound}
\end{align}

Equation~\eqref{eq:final_explicit_MN_bound} is the explicit finite-time
training-to-sampling guarantee. It makes no assumption that the optimization
constants are independent of the score-class complexity \(M\). In particular,
increasing \(M\) decreases the approximation term but may change
\(L_M\), \(\mu_M\), \(\sigma_M^2\), or \(\Delta_{0,M}\).

\subsection{Verification of the Simplified \texorpdfstring{\(\epsilon\)}{epsilon}-Rate}
\label{subsec:final_simplified_rate}

We now verify Corollary~\ref{cor:stopped_rate}. Fix \(t_0>0\) independently of
\(\epsilon\), and suppose the optimization geometry is uniformly controlled
along the chosen sequence of score classes, horizons, and grids. Thus, for the
constants of each corresponding randomized-time objective,
\[
    L_M\le\overline L,
    \qquad
    \mu_M\ge\underline\mu>0,
    \qquad
    \sigma_M^2\le\overline\sigma^2,
    \qquad
    \Delta_{0,M}\le\overline\Delta_0.
\]
Take \(B\) fixed independently of \(\epsilon\) and a fixed
\[
    r_0
    \ge
    \max
    \left\{
        4,\frac{2\overline L}{\underline\mu}
    \right\}.
\]
Since \(N=RB\), we then have
\[
    R=\Theta(N),
\]
and \eqref{eq:final_Eopt_explicit} gives
\begin{equation}
    \mathcal E_{\mathrm{opt}}(N,M)
    =
    \mathcal O(N^{-1}),
    \label{eq:final_opt_N}
\end{equation}
with an initialization transient of order \(N^{-2}\).

Choose
\begin{equation}
    T
    =
    C_T\log(1/\epsilon)
    \label{eq:final_T_choice}
\end{equation}
with \(C_T\) sufficiently large. Then
\[
    e^{-T}
    =
    \mathcal O(\epsilon),
\]
and
\[
    H_K
    =
    T-t_0
    \le
    T
    =
    \mathcal O(\log(1/\epsilon)).
\]

For the approximation class, choose
\begin{equation}
    M
    =
    \widetilde{\Theta}
    \left(
        \epsilon^{-2/\eta_{\mathrm{app}}}
    \right).
    \label{eq:final_M_choice}
\end{equation}
Then
\[
    M^{-\eta_{\mathrm{app}}}
    =
    \widetilde{\mathcal O}(\epsilon^2).
\]
Since
\[
    \frac{2}{\eta_{\mathrm{app}}}
    =
    2
    +
    \frac{d_x'}{\alpha_x}
    +
    \frac{d_y'}{\alpha_y},
\]
the same requirement can be written as
\begin{equation}
    M
    =
    \widetilde{\Theta}
    \left(
        \epsilon^{
        -\left(
            2+d_x'/\alpha_x+d_y'/\alpha_y
        \right)}
    \right).
    \label{eq:final_M_dimension_rate}
\end{equation}

For online optimization, choose
\begin{equation}
    N
    =
    \widetilde{\Theta}(\epsilon^{-2}).
    \label{eq:final_N_choice}
\end{equation}
By \eqref{eq:final_opt_N},
\[
    \mathcal E_{\mathrm{opt}}(N,M)
    =
    \widetilde{\mathcal O}(\epsilon^2).
\]

Finally, choose an admissible reverse-time grid satisfying
\begin{equation}
    \Delta_{\max}
    \le
    \widetilde{\mathcal O}
    \left(
        M^{-\eta_{\mathrm{app}}}
    \right)
    =
    \widetilde{\mathcal O}(\epsilon^2)
    \label{eq:final_grid_requirement}
\end{equation}
and
\begin{equation}
    K
    =
    \widetilde{\Theta}
    \left(
        (d_x+m_2)\epsilon^{-2}
    \right).
    \label{eq:final_K_choice}
\end{equation}
For the admissible grids of
Lemma~\ref{lem:forward_law_tv_stability}, the maximum step size satisfies,
up to logarithmic factors,
\[
    \Delta_{\max}
    \lesssim
    \frac{H_K}{K},
\]
so the choice \eqref{eq:final_K_choice} is compatible with
\eqref{eq:final_grid_requirement} after absorbing the
\(H_K=O(\log(1/\epsilon))\) factor into the
\(\widetilde{\mathcal O}\)-notation.

With these choices,
\[
    C_{\mathrm{disc}}
    \sqrt{
        \frac{d_x+m_2}{K}
    }
    =
    \widetilde{\mathcal O}(\epsilon),
\]
and
\begin{align}
&
\sqrt{
H_K
\left[
    \mathcal A_K(M)
    +
    \mathcal E_{\mathrm{opt}}(N,M)
\right]
}
\nonumber\\
&\qquad\le
\widetilde{\mathcal O}
\left(
    \sqrt{
        \log(1/\epsilon)
        \left(
            \epsilon^2+\epsilon^2
        \right)
    }
\right)
=
\widetilde{\mathcal O}(\epsilon).
\label{eq:final_score_term_epsilon}
\end{align}
Substituting into \eqref{eq:final_general_training_sampling} gives
\begin{equation}
\mathbb E_{\mathrm{SGD}}
\mathbb E_Y
\left[
    \operatorname{TV}
    \left(
        p_{t_0}(\cdot\mid Y),
        \widehat p_{t_0}(\cdot\mid Y)
    \right)
\right]
\le
\widetilde{\mathcal O}(\epsilon).
\label{eq:final_stopped_epsilon}
\end{equation}
This is the claimed
\(\widetilde{\mathcal O}(\epsilon)\) expected total-variation guarantee.

The complexity requirements should therefore be read as two distinct
statements:
\begin{equation}
\boxed{
    M
    =
    \widetilde{\Theta}
    \left(
        \epsilon^{-2/\eta_{\mathrm{app}}}
    \right)
}
\qquad
\text{and}
\qquad
\boxed{
    N
    =
    \widetilde{\Theta}
    \left(
        \epsilon^{-2}
    \right)
}
\label{eq:final_separate_complexities}
\end{equation}
under uniform optimization geometry. The first is an approximation-capacity
requirement; the second is the number of fresh denoising samples consumed by
online SGD. They should not be combined into a single online sample-complexity
exponent.

If the optimization constants
\(L_M,\mu_M,\sigma_M^2,\Delta_{0,M}\) are not uniformly controlled as
\(M\) grows, then the appropriate result is the explicit bound
\eqref{eq:final_explicit_MN_bound}, rather than the simplified
\(N=\widetilde{\mathcal O}(\epsilon^{-2})\) statement.

This completes the proof of the stopped-law results. The extension from
\(p_{t_0}(\cdot\mid y)\) to the original conditional law
\(p_0(\cdot\mid y)\) requires the additional smoothing-bias analysis and is
treated separately in Appendix~\ref{app:tv_to_p0_extension}.

\section{Extension: Total Variation Guarantee to the Original Data Law}
\label{app:tv_to_p0_extension}

The primary result of the paper controls the implementable reverse sampler
relative to the positive-time conditional law
\(p_{t_0}(\cdot\mid y)\). In this section, we show how this stopped-law
guarantee can be extended to the original conditional data law
\(p_0(\cdot\mid y)\).

An additional regularity assumption is necessary. If the clean conditional law
is exactly supported on a strict lower-dimensional manifold in
\(\mathbb R^{d_x}\), then \(p_0(\cdot\mid y)\) is singular with respect to
Lebesgue measure, whereas \(p_{t_0}(\cdot\mid y)\) is Gaussian-smoothed and
absolutely continuous for every \(t_0>0\). In that setting, ambient total
variation between the two laws does not vanish as \(t_0\downarrow0\).
Accordingly, the extension below assumes that the response distribution is
full-dimensional.

\begin{assumption}[Full-dimensional regularity for the TV-to-\(p_0\) extension]
\label{ass:distributional_full}

Let \(Y\sim P_Y\) take values in \(\mathbb R^{d_y}\), and let
\(X_0\mid Y=y\sim p_0(\cdot\mid y)\) take values in
\(\mathbb R^{d_x}\). The following conditions hold.

\begin{enumerate}

    \item \textbf{Full-dimensional H\"older conditional density.}
    For \(P_Y\)-a.e.\ \(y\), the conditional law
    \(P_{X_0\mid Y=y}\) admits a Lebesgue density
    \(p_0(\cdot\mid y)\) on \(\mathbb R^{d_x}\). There exists
    \(B_{\mathrm{dens}}<\infty\) such that
    \[
        \operatorname*{ess\,sup}_{y\sim P_Y}
        \|p_0(\cdot\mid y)\|_\infty
        \le
        B_{\mathrm{dens}}.
    \]
    Moreover, for some \(\beta>0\), \(p_0(\cdot\mid y)\) is uniformly
    \(\beta\)-H\"older smooth in the response coordinate \(x\).
    More precisely, if
    \[
        m=\lceil\beta\rceil-1,
        \qquad
        \beta=m+\alpha,
        \qquad
        \alpha\in(0,1],
    \]
    then derivatives up to order \(m\) are uniformly bounded and the
    derivatives of order \(m\) are uniformly \(\alpha\)-H\"older continuous,
    with constants independent of \(y\).

    For the total-variation smoothing step, assume in addition the uniform
    integral regularity
    \[
    \operatorname*{ess\,sup}_{y\sim P_Y}
    \left\{
    \|p_0(\cdot\mid y)\|_{B^{\beta\wedge2}_{1,\infty}}
    +
    \int_{\mathbb R^{d_x}}
    (1+\|x\|_2)\|\nabla_x p_0(x\mid y)\|_2\,dx
    \right\}
    <\infty,
    \]
    where the derivative may be interpreted weakly. For \(s>0\), we use the
    equivalent modulus-of-smoothness norm
    \[
    \|f\|_{B^s_{1,\infty}}
    :=
    \|f\|_1+
    \sup_{0<\|h\|_2\le1}
    \|h\|_2^{-s}\|\Delta_h^r f\|_1,
    \qquad r\in\mathbb N,\quad r>s,
    \]
    with \(\Delta_h^r\) the \(r\)-fold finite difference. The Besov condition
    supplies the Gaussian mollification rate, while the weighted gradient term
    controls the deterministic OU rescaling.

    \item \textbf{Anisotropic approximation smoothness.}
    The conditioning law \(P_Y\) is supported on a
    \(d_y'\)-dimensional embedded submanifold
    \[
        \mathcal M_Y
        \subset
        \mathbb R^{d_y},
    \]
    and admits a density with respect to the intrinsic volume measure on
    \(\mathcal M_Y\).

    In local intrinsic coordinates for \(Y\), the conditional density
    \(p_0(x\mid y)\) satisfies the anisotropic regularity used by the
    conditional score-approximation argument: the response coordinate has
    smoothness \(\alpha_x>0\), while the intrinsic conditioning coordinate has
    smoothness \(\alpha_y>0\). The corresponding localized H\"older constants,
    chart constants, and density regularity quantities are uniformly controlled
    on the logarithmic localization regions used in the proof. The score class
    contains the later-time blockwise approximants used in the conditional
    approximation theorem, and the required density nondegeneracy and
    inverse-density quantities are uniformly controlled on those regions.

    \item \textbf{Sub-exponential tails.}
    There exist constants \(\lambda_Y,\lambda_X>0\) such that
    \[
        \mathbb E
        \exp\!\left(
            \lambda_Y\|Y\|_2
        \right)
        <\infty,
    \]
    and
    \[
        \operatorname*{ess\,sup}_{y\sim P_Y}
        \mathbb E
        \left[
            \exp\!\left(
                \lambda_X\|X_0\|_2
            \right)
            \,\middle|\,
            Y=y
        \right]
        <\infty.
    \]

\end{enumerate}

Define the later-time approximation exponents
\begin{equation}
    \rho_0
    :=
    \frac{2}{
        2\alpha_x+d_x+(\alpha_x/\alpha_y)d_y'
    },
    \qquad
    \zeta_0
    :=
    \frac{2\alpha_y}{
        2\alpha_y+d_y'
    },
    \qquad
    \gamma_0
    :=
    \frac{\alpha_y d_x}{
        2\alpha_y+d_y'
    }.
    \label{eq:p0_extension_exponents}
\end{equation}

For an approximation class of complexity \(M\), the later-time regime used
below requires
\begin{equation}
    M^{-\rho_0}
    \le
    t_0.
    \label{eq:p0_second_regime_condition}
\end{equation}
We additionally assume that the class \(\mathcal F_{\Theta,M}\) contains the
later-time blockwise conditional score approximants furnished by the
conditional approximation theorem, so that after localization and grid
transfer their error is bounded by
\(\widetilde{\mathcal O}(t_0^{-\gamma_0}M^{-\zeta_0}+\Delta_{\max})\).
\end{assumption}

\begin{remark}[Relation to the stopped-law assumption]
\label{rem:p0_assumption_difference}
Assumption~\ref{ass:distributional_full} is used only for convergence to the
original law \(p_0(\cdot\mid y)\). It is deliberately different from
Assumption~\ref{ass:distributional}. The latter permits exact
lower-dimensional response support because the primary target is the
positive-time smoothed law \(p_{t_0}\). The present assumption instead requires
a full-dimensional response density so that the smoothing bias
\[
    \operatorname{TV}
    \left(
        p_0(\cdot\mid y),
        p_{t_0}(\cdot\mid y)
    \right)
\]
can converge to zero as \(t_0\downarrow0\).
Its integral Besov and weighted-gradient conditions are stronger than
pointwise H\"older smoothness alone and are used only to obtain the stated
uniform total-variation smoothing rate.
\end{remark}

\subsection{Full-Dimensional Approximation Input}
\label{subsec:p0_full_dim_approx}

For this extension, define the best-in-class randomized-time score error
\begin{equation}
    \mathcal A_{K,0}(M)
    :=
    \inf_{\theta\in\Theta_M}
    \mathbb E
    \left[
        \left\|
            s_\theta(X_{t_J},Y,t_J)
            -
            s^\star(X_{t_J},t_J,Y)
        \right\|_2^2
    \right].
    \label{eq:p0_AKM_definition}
\end{equation}

Under Assumption~\ref{ass:distributional_full} and the second-regime condition
\eqref{eq:p0_second_regime_condition}, the later-time conditional
score-approximation input adapted from the blockwise construction of
\citet{tang2025conditional}, together with the class-containment clause in the
assumption, gives
\begin{equation}
    \mathcal A_{K,0}(M)
    \le
    \widetilde C_{\mathrm{app},0}
    \left(
        t_0^{-\gamma_0}M^{-\zeta_0}
        +
        \Delta_{\max}
    \right).
    \label{eq:p0_approx_general}
\end{equation}
The \(\Delta_{\max}\) term arises from transferring the time-integrated
approximation guarantee to the randomized grid-time objective. Consequently,
if
\begin{equation}
    \Delta_{\max}
    \le
    \widetilde{\mathcal O}
    \left(
        t_0^{-\gamma_0}M^{-\zeta_0}
    \right),
    \label{eq:p0_grid_fineness}
\end{equation}
then
\begin{equation}
    \mathcal A_{K,0}(M)
    \le
    \widetilde{\mathcal O}
    \left(
        t_0^{-\gamma_0}M^{-\zeta_0}
    \right).
    \label{eq:p0_approx_simplified}
\end{equation}

The response dimension appearing in this rate is the ambient dimension
\(d_x\), rather than the intrinsic dimension \(d_x'\), because the present
extension assumes a full-dimensional response density.

\subsection{General Total-Variation Bound to the Original Law}
\label{subsec:p0_general_bound}

\begin{corollary}[Total-variation guarantee to the original conditional law]
\label{cor:tv_to_p0}

Suppose Assumptions~\ref{ass:distributional_full} and
\ref{ass:local_population_pl} hold. Let
\(\widehat p_{t_0}(\cdot\mid y)\) denote the law generated by the admissibly
discretized reverse process initialized from
\(\mathcal N(0,I_{d_x})\) and using the score
\(\widehat s=s_{\widehat\theta}\) produced by online SGD.

Assume \(0<t_0<1\le T\).

Let \(M\) denote the score-class approximation complexity, and let
\[
    N=RB
\]
denote the total number of fresh online denoising samples used by SGD. Define
\[
    m_2
    :=
    \operatorname*{ess\,sup}_{y\sim P_Y}
    \mathbb E
    \left[
        \|X_0\|_2^2
        \,\middle|\,
        Y=y
    \right],
\]
which is finite by the sub-exponential tail assumption.

Suppose
\[
    M^{-\rho_0}
    \le
    t_0
\]
and that the admissible grid satisfies
\[
    \Delta_{\max}
    \le
    \widetilde{\mathcal O}
    \left(
        t_0^{-\gamma_0}M^{-\zeta_0}
    \right).
\]
Then
\begin{align}
&
\mathbb E_{\mathrm{SGD}}
\mathbb E_Y
\left[
    \operatorname{TV}
    \left(
        p_0(\cdot\mid Y),
        \widehat p_{t_0}(\cdot\mid Y)
    \right)
\right]
\nonumber\\
&\quad\le
C_{\mathrm{term}}e^{-T}
+
C_{\mathrm{disc}}
\sqrt{
    \frac{d_x+m_2}{K}
}
\nonumber\\
&\qquad+
C_{\mathrm{smooth}}
t_0^{(\beta\wedge2)/2}
+
C_{\mathrm{OU}}d_x t_0
\nonumber\\
&\qquad+
C_{\mathrm{stab}}
\sqrt{
H_K
\left[
    \widetilde C_{\mathrm{app},0}
    t_0^{-\gamma_0}M^{-\zeta_0}
    +
    \mathcal E_{\mathrm{opt}}(N,M)
\right]
}.
\label{eq:tv_to_p0_general_bound}
\end{align}
Here \(\mathcal E_{\mathrm{opt}}(N,M)\) denotes the finite-time population
optimization error defined in \eqref{eq:optimization_error_NM}.
\end{corollary}

The terms in \eqref{eq:tv_to_p0_general_bound} have the following
interpretation:
\[
\underbrace{e^{-T}}_{\text{terminal initialization}},
\qquad
\underbrace{\sqrt{(d_x+m_2)/K}}_{\text{time discretization}},
\]
\[
\underbrace{
t_0^{(\beta\wedge2)/2}
+
d_x t_0
}_{\text{smoothing bias}},
\]
and
\[
\underbrace{
\sqrt{
H_K
\left[
t_0^{-\gamma_0}M^{-\zeta_0}
+
\mathcal E_{\mathrm{opt}}(N,M)
\right]
}
}_{\text{approximation and finite-time optimization}}.
\]
Thus, unlike the stopped-law theorem, the original-law guarantee contains an
additional smoothing-bias contribution and uses the full-dimensional
later-time approximation rate.

\subsection{Proof of Corollary~\ref{cor:tv_to_p0}}
\label{app:obtaining_final_result_p0}

\begin{proof}

\medskip
\noindent\textbf{Step 1: Decompose the total-variation error.}

For each fixed \(y\), let \(p_{t_0}(\cdot\mid y)\) denote the OU-smoothed
conditional law at time \(t_0\). Let
\(\widetilde p_{t_0}(\cdot\mid y)\) denote the admissibly discretized reverse
law initialized from \(p_T(\cdot\mid y)\) and using the learned score
\(\widehat s\). Finally, let
\(\widehat p_{t_0}(\cdot\mid y)\) denote the same learned-score reverse chain
initialized from \(\mathcal N(0,I_{d_x})\).

By the triangle inequality,
\begin{align}
&
\operatorname{TV}
\left(
    p_0(\cdot\mid y),
    \widehat p_{t_0}(\cdot\mid y)
\right)
\nonumber\\
&\quad\le
\operatorname{TV}
\left(
    p_0(\cdot\mid y),
    p_{t_0}(\cdot\mid y)
\right)
+
\operatorname{TV}
\left(
    p_{t_0}(\cdot\mid y),
    \widetilde p_{t_0}(\cdot\mid y)
\right)
\nonumber\\
&\qquad+
\operatorname{TV}
\left(
    \widetilde p_{t_0}(\cdot\mid y),
    \widehat p_{t_0}(\cdot\mid y)
\right).
\label{eq:p0_tv_decomposition}
\end{align}
These are, respectively, the smoothing-bias, learned-score/discretization, and
terminal-initialization errors.

\medskip
\noindent\textbf{Step 2: Bound the smoothing bias.}

Lemma~\ref{lem:tv_smoothing_integral} gives
\begin{align}
&
\mathbb E_Y
\left[
    \operatorname{TV}
    \left(
        p_0(\cdot\mid Y),
        p_{t_0}(\cdot\mid Y)
    \right)
\right]
\nonumber\\
&\qquad\le
C_{\mathrm{smooth}}
t_0^{(\beta\wedge2)/2}
+
C_{\mathrm{OU}}d_x t_0.
\label{eq:p0_smoothing_bound}
\end{align}

\medskip
\noindent\textbf{Step 3: Bound the terminal-initialization error.}

For a fixed learned score, the laws
\(\widetilde p_{t_0}(\cdot\mid y)\) and
\(\widehat p_{t_0}(\cdot\mid y)\) are obtained by applying the same reverse
Markov kernel to the initial distributions \(p_T(\cdot\mid y)\) and
\(\mathcal N(0,I_{d_x})\). Therefore, by contraction of total variation under
Markov kernels,
\[
\operatorname{TV}
\left(
    \widetilde p_{t_0}(\cdot\mid y),
    \widehat p_{t_0}(\cdot\mid y)
\right)
\le
\operatorname{TV}
\left(
    p_T(\cdot\mid y),
    \mathcal N(0,I_{d_x})
\right).
\]
The conditional Gaussian-KL calculation in
Eq.~\eqref{eq:terminal_ou_kl}, followed by Pinsker's inequality, therefore
yields
\begin{equation}
\mathbb E_Y
\left[
    \operatorname{TV}
    \left(
        \widetilde p_{t_0}(\cdot\mid Y),
        \widehat p_{t_0}(\cdot\mid Y)
    \right)
\right]
\le
C_{\mathrm{term}}e^{-T}.
\label{eq:p0_terminal_bound}
\end{equation}

\medskip
\noindent\textbf{Step 4: Bound the score and discretization error.}

For a fixed realization of \(\widehat s\), apply
Lemma~\ref{lem:forward_law_tv_stability}. This gives
\begin{align}
&
\mathbb E_Y
\left[
    \operatorname{TV}
    \left(
        p_{t_0}(\cdot\mid Y),
        \widetilde p_{t_0}(\cdot\mid Y)
    \right)
\right]
\nonumber\\
&\qquad\le
C_{\mathrm{stab}}
\left(
\sum_{k=1}^{K}
\Delta_k
\mathbb E
\left[
    \left\|
        \widehat s(X_{t_k},Y,t_k)
        -
        s^\star(X_{t_k},t_k,Y)
    \right\|_2^2
\right]
\right)^{1/2}
\nonumber\\
&\qquad\quad+
C_{\mathrm{disc}}
\sqrt{
    \frac{d_x+m_2}{K}
}.
\label{eq:p0_reverse_stability}
\end{align}

Define
\begin{equation}
    \widehat{\mathcal A}_{K,0}
    :=
    \mathbb E
    \left[
        \left\|
            \widehat s(X_{t_J},Y,t_J)
            -
            s^\star(X_{t_J},t_J,Y)
        \right\|_2^2
    \right].
    \label{eq:p0_learned_score_error}
\end{equation}
Because
\[
    \mathbb P(J=k)
    =
    \frac{\Delta_k}{H_K},
\]
we have
\begin{align}
H_K\widehat{\mathcal A}_{K,0}
&=
\sum_{k=1}^{K}
\Delta_k
\mathbb E
\left[
    \left\|
        \widehat s(X_{t_k},Y,t_k)
        -
        s^\star(X_{t_k},t_k,Y)
    \right\|_2^2
\right].
\label{eq:p0_grid_score_identity}
\end{align}
Hence
\begin{align}
&
\mathbb E_Y
\left[
    \operatorname{TV}
    \left(
        p_{t_0}(\cdot\mid Y),
        \widetilde p_{t_0}(\cdot\mid Y)
    \right)
\right]
\nonumber\\
&\qquad\le
C_{\mathrm{stab}}
\sqrt{
    H_K\widehat{\mathcal A}_{K,0}
}
+
C_{\mathrm{disc}}
\sqrt{
    \frac{d_x+m_2}{K}
}.
\label{eq:p0_stability_score_error}
\end{align}

\medskip
\noindent\textbf{Step 5: Separate approximation and optimization.}

Let
\[
    Z_J
    :=
    \nabla_x\log q_{t_J}(X_{t_J}\mid X_0).
\]
The denoising identity gives
\[
    s^\star(X_{t_J},t_J,Y)
    =
    \mathbb E
    \left[
        Z_J
        \,\middle|\,
        X_{t_J},Y,t_J
    \right].
\]
Therefore, for every \(\theta\in\Theta_M\),
\begin{align}
\mathcal L_K(\theta)
&=
C_K
+
\mathbb E
\left[
    \left\|
        s_\theta(X_{t_J},Y,t_J)
        -
        s^\star(X_{t_J},t_J,Y)
    \right\|_2^2
\right],
\label{eq:p0_denoising_projection}
\end{align}
where \(C_K\) is independent of \(\theta\).

Taking the infimum over \(\Theta_M\) gives
\begin{equation}
    \mathcal L_{K,M}^\star
    =
    C_K+\mathcal A_{K,0}(M).
    \label{eq:p0_Lstar_projection}
\end{equation}
Applying \eqref{eq:p0_denoising_projection} at
\(\widehat\theta\),
\begin{align}
    \widehat{\mathcal A}_{K,0}
    &=
    \mathcal L_K(\widehat\theta)-C_K
    \nonumber\\
    &=
    \left[
        \mathcal L_K(\widehat\theta)
        -
        \mathcal L_{K,M}^\star
    \right]
    +
    \mathcal A_{K,0}(M).
\label{eq:p0_score_decomposition_exact}
\end{align}
Taking expectation over online SGD yields
\begin{equation}
\mathbb E_{\mathrm{SGD}}
\left[
    \widehat{\mathcal A}_{K,0}
\right]
=
\mathcal A_{K,0}(M)
+
\mathcal E_{\mathrm{opt}}(N,M).
\label{eq:p0_score_decomposition_expectation}
\end{equation}

By \eqref{eq:p0_approx_simplified},
\[
    \mathcal A_{K,0}(M)
    \le
    \widetilde{\mathcal O}
    \left(
        t_0^{-\gamma_0}M^{-\zeta_0}
    \right).
\]
Consequently,
\begin{equation}
\mathbb E_{\mathrm{SGD}}
\left[
    \widehat{\mathcal A}_{K,0}
\right]
\le
\widetilde C_{\mathrm{app},0}
t_0^{-\gamma_0}M^{-\zeta_0}
+
\mathcal E_{\mathrm{opt}}(N,M).
\label{eq:p0_score_bound_MN}
\end{equation}

Taking expectation over SGD in
\eqref{eq:p0_stability_score_error} and using Jensen's inequality,
\begin{align}
&
\mathbb E_{\mathrm{SGD}}
\mathbb E_Y
\left[
    \operatorname{TV}
    \left(
        p_{t_0}(\cdot\mid Y),
        \widetilde p_{t_0}(\cdot\mid Y)
    \right)
\right]
\nonumber\\
&\qquad\le
C_{\mathrm{disc}}
\sqrt{
    \frac{d_x+m_2}{K}
}
\nonumber\\
&\qquad\quad+
C_{\mathrm{stab}}
\sqrt{
H_K
\left[
    \widetilde C_{\mathrm{app},0}
    t_0^{-\gamma_0}M^{-\zeta_0}
    +
    \mathcal E_{\mathrm{opt}}(N,M)
\right]
}.
\label{eq:p0_sampling_error_MN}
\end{align}

\medskip
\noindent\textbf{Step 6: Combine all terms.}

Taking expectations in \eqref{eq:p0_tv_decomposition} and substituting
\eqref{eq:p0_smoothing_bound},
\eqref{eq:p0_terminal_bound}, and
\eqref{eq:p0_sampling_error_MN}, we obtain
\begin{align}
&
\mathbb E_{\mathrm{SGD}}
\mathbb E_Y
\left[
    \operatorname{TV}
    \left(
        p_0(\cdot\mid Y),
        \widehat p_{t_0}(\cdot\mid Y)
    \right)
\right]
\nonumber\\
&\quad\le
C_{\mathrm{term}}e^{-T}
+
C_{\mathrm{disc}}
\sqrt{
    \frac{d_x+m_2}{K}
}
\nonumber\\
&\qquad+
C_{\mathrm{smooth}}
t_0^{(\beta\wedge2)/2}
+
C_{\mathrm{OU}}d_x t_0
\nonumber\\
&\qquad+
C_{\mathrm{stab}}
\sqrt{
H_K
\left[
    \widetilde C_{\mathrm{app},0}
    t_0^{-\gamma_0}M^{-\zeta_0}
    +
    \mathcal E_{\mathrm{opt}}(N,M)
\right]
}.
\end{align}
This proves \eqref{eq:tv_to_p0_general_bound}.
\end{proof}

\subsection{Simplified \texorpdfstring{\(\epsilon\)}{epsilon}-Complexity}
\label{subsec:p0_simplified_complexity}

We now specialize the general bound under the uniform optimization-geometry
condition used in Corollary~\ref{cor:stopped_rate}. In this extension, the
uniformity is required along the full chosen sequence of score classes, grids,
horizons, and cutoffs \(t_0\downarrow0\). Thus, for the constants of each
corresponding randomized-time objective, assume
\[
    L_M\le\overline L,
    \qquad
    \mu_M\ge\underline\mu>0,
    \qquad
    \sigma_M^2\le\overline\sigma^2,
    \qquad
    \Delta_{0,M}\le\overline\Delta_0.
\]
With mini-batch size \(B\) fixed independently of \(\epsilon\) and a fixed admissible \(r_0\),
Proposition~\ref{prop:population_pl_opt_main} gives
\begin{equation}
    \mathcal E_{\mathrm{opt}}(N,M)
    =
    \widetilde{\mathcal O}(N^{-1}).
    \label{eq:p0_opt_simple}
\end{equation}

Define
\begin{equation}
    a_\beta
    :=
    \max
    \left\{
        1,\frac{2}{\beta}
    \right\}.
    \label{eq:a_beta_extension}
\end{equation}
Choose
\begin{equation}
    t_0
    =
    \widetilde{\Theta}
    \left(
        \epsilon^{a_\beta}
    \right).
    \label{eq:p0_t0_choice}
\end{equation}
Then both smoothing-bias contributions satisfy
\[
t_0^{(\beta\wedge2)/2}
=
\widetilde{\mathcal O}(\epsilon),
\]
and
\[
    d_x t_0
    =
    \widetilde{\mathcal O}(\epsilon),
\]
for fixed \(d_x\).

The approximation term requires
\[
    t_0^{-\gamma_0}M^{-\zeta_0}
    =
    \widetilde{\mathcal O}(\epsilon^2).
\]
If
\[
    M
    =
    \widetilde{\Theta}
    \left(
        \epsilon^{-\Gamma_{\mathrm{app},0}}
    \right),
\]
this condition is satisfied whenever
\[
    \Gamma_{\mathrm{app},0}\zeta_0
    -
    a_\beta\gamma_0
    \ge
    2.
\]
Equivalently,
\[
    \Gamma_{\mathrm{app},0}
    \ge
    2
    +
    \frac{d_y'}{\alpha_y}
    +
    \frac{a_\beta d_x}{2}.
\]

The second-regime requirement
\[
    M^{-\rho_0}
    \le
    t_0
\]
is satisfied whenever
\[
    \Gamma_{\mathrm{app},0}\rho_0
    \ge
    a_\beta,
\]
or equivalently,
\[
    \Gamma_{\mathrm{app},0}
    \ge
    \frac{a_\beta}{2}
    \left(
        2\alpha_x+d_x
        +
        \frac{\alpha_x}{\alpha_y}d_y'
    \right).
\]

Thus define
\begin{equation}
\Gamma_{\mathrm{app},0}
:=
\max
\left\{
    2+\frac{d_y'}{\alpha_y}
      +\frac{a_\beta d_x}{2},
    \;
    \frac{a_\beta}{2}
    \left(
        2\alpha_x+d_x
        +
        \frac{\alpha_x}{\alpha_y}d_y'
    \right)
\right\}.
\label{eq:p0_Gamma_app}
\end{equation}
Then the choice
\begin{equation}
    M
    =
    \widetilde{\Theta}
    \left(
        \epsilon^{-\Gamma_{\mathrm{app},0}}
    \right)
    \label{eq:p0_M_choice}
\end{equation}
controls both the full-dimensional approximation error and the later-time
regime condition.

Separately, choose
\begin{equation}
    N
    =
    \widetilde{\Theta}
    \left(
        \epsilon^{-2}
    \right).
    \label{eq:p0_N_choice}
\end{equation}
Then, under uniform optimization geometry,
\[
    \mathcal E_{\mathrm{opt}}(N,M)
    =
    \widetilde{\mathcal O}(\epsilon^2).
\]

Choose also
\[
    T
    =
    \Theta(\log(1/\epsilon)).
\]
Then
\[
    e^{-T}
    =
    \mathcal O(\epsilon),
    \qquad
    H_K
    \le
    T
    =
    \mathcal O(\log(1/\epsilon)).
\]

Finally, define
\begin{equation}
    \Lambda_0
    :=
    \max
    \left\{
        2,\,
        \Gamma_{\mathrm{app},0}\zeta_0
        -
        a_\beta\gamma_0
    \right\}.
    \label{eq:p0_Lambda}
\end{equation}
An admissible grid satisfying
\begin{equation}
    K
    =
    \widetilde{\Theta}
    \left(
        (d_x+m_2)\epsilon^{-2}
        \;\vee\;
        \epsilon^{-\Lambda_0}
    \right)
    \label{eq:p0_K_choice_appendix}
\end{equation}
controls both the reverse-process discretization term and the grid-fineness
requirement
\[
    \Delta_{\max}
    \le
    \widetilde{\mathcal O}
    \left(
        t_0^{-\gamma_0}M^{-\zeta_0}
    \right).
\]
Equivalently, retaining the total grid weight explicitly, it is sufficient to
take
\[
    K
    \gtrsim
    (d_x+m_2)\epsilon^{-2}
    \;\vee\;
    H_K
    \epsilon^{-
    \left(
        \Gamma_{\mathrm{app},0}\zeta_0
        -
        a_\beta\gamma_0
    \right)},
\]
up to logarithmic factors.

With the choices
\[
    t_0
    =
    \widetilde{\Theta}
    \left(
        \epsilon^{a_\beta}
    \right),
\]
\[
    M
    =
    \widetilde{\Theta}
    \left(
        \epsilon^{-\Gamma_{\mathrm{app},0}}
    \right),
\]
\[
    N
    =
    \widetilde{\Theta}
    \left(
        \epsilon^{-2}
    \right),
\]
\[
    T
    =
    \Theta(\log(1/\epsilon)),
\]
and the admissible grid above, every term in
\eqref{eq:tv_to_p0_general_bound} is
\(\widetilde{\mathcal O}(\epsilon)\). Therefore,
\begin{equation}
\mathbb E_{\mathrm{SGD}}
\mathbb E_Y
\left[
    \operatorname{TV}
    \left(
        p_0(\cdot\mid Y),
        \widehat p_{t_0}(\cdot\mid Y)
    \right)
\right]
\le
\widetilde{\mathcal O}(\epsilon).
\label{eq:p0_final_epsilon}
\end{equation}

As in the stopped-law result, the two main complexity requirements should be
kept distinct:
\begin{equation}
\boxed{
    M
    =
    \widetilde{\Theta}
    \left(
        \epsilon^{-\Gamma_{\mathrm{app},0}}
    \right)
}
\qquad\text{and}\qquad
\boxed{
    N
    =
    \widetilde{\Theta}
    \left(
        \epsilon^{-2}
    \right).
}
\label{eq:p0_separate_complexities}
\end{equation}
The first is the approximation-capacity requirement for the
full-dimensional score class. The second is the number of fresh online
denoising samples consumed by SGD under uniform optimization geometry. If the
optimization constants deteriorate with \(M\), the correct statement is the
general bound \eqref{eq:tv_to_p0_general_bound} together with the explicit
finite-time optimization bound from
Proposition~\ref{prop:population_pl_opt_main}.

\section{Supplementary Lemmas}
\label{sec:supplementary_lemmas}

\begin{lemma}[Forward-law TV stability for discretized reverse sampling]
\label{lem:forward_law_tv_stability}
Let \(0<t_0<1\le T\), and let
\[
t_0<t_1<\cdots<t_K=T,
\qquad
\Delta_k:=t_k-t_{k-1},
\qquad
k=1,\ldots,K.
\]
For \(P_Y\)-a.e.\ \(y\), let \(p_t(\cdot\mid y)\) denote the OU forward
marginal at time \(t\), and let
\[
s^\star(x,t,y):=\nabla_x\log p_t(x\mid y)
\]
be the conditional score. Let \(p_{t_0}(\cdot\mid y)\) be the law obtained by
running the continuous reverse SDE from \(p_T(\cdot\mid y)\) down to time
\(t_0\) using the true score \(s^\star\). For a fixed score estimator
\(\bar s\), let \(\bar p_{t_0}^{\mathrm{dis}}(\cdot\mid y)\) denote the law of
the corresponding exponential-integrator reverse sampler on the above grid,
initialized from \(p_T(\cdot\mid y)\) and using \(\bar s\) in place of
\(s^\star\), with the update specified in
Eq.~\eqref{eq:reverse_exponential_integrator}.

Assume the grid is chosen in the admissible form of
\citet[Corollary~1]{benton2024nearly}, with early-stopping parameter \(t_0\).
In the forward-time notation used here, the corresponding admissible grid
condition can be written as
\[
\Delta_k
\le
\kappa\min\{1,t_{k-1}\},
\qquad
\kappa
=
\Theta\!\left(\frac{T+\log(1/t_0)}{K}\right),
\]
with \(K\) large enough for the grid construction in that corollary. Suppose
also that the conditional second moments are uniformly bounded:
\[
m_2
:=
\operatorname*{ess\,sup}_{y\sim P_Y}
\mathbb E\!\left[\|X_0\|_2^2\,\middle|\,Y=y\right]
<\infty .
\]
Then, whenever the score-error term below is finite,
\begin{align}
\mathbb E_Y\!\left[
\mathrm{TV}\bigl(
p_{t_0}(\cdot\mid Y),
\bar p_{t_0}^{\mathrm{dis}}(\cdot\mid Y)
\bigr)
\right]
&\lesssim
\left(
\sum_{k=1}^{K}
\Delta_k\,
\mathbb E\!\left[
\left\|
\bar s(X_{t_k},Y,t_k)
-
s^\star(X_{t_k},t_k,Y)
\right\|_2^2
\right]
\right)^{1/2}
\nonumber\\
&\quad+
\widetilde{\mathcal O}\!\left(
\sqrt{\frac{d_x+m_2}{K}}
\right).
\label{eq:forward_law_tv_stability}
\end{align}
In the normalized presentation where \(m_2=\mathcal O(d_x)\), the last term is
\[
\widetilde{\mathcal O}\!\left(\sqrt{\frac{d_x}{K}}\right).
\]
If \(d_x\) and \(m_2\) are treated as fixed problem-dependent constants in the
main-text notation, this term is written as
\(\widetilde{\mathcal O}(K^{-1/2})\).
\end{lemma}

\begin{proof}
The result is a conditional version of the reverse-process KL discretization
bound of \citet{benton2024nearly}, converted to total variation. We give the
reduction for completeness.

Fix \(y\) such that the conditional law \(p_0(\cdot\mid y)\) has finite second
moment. Apply the discretization theorem of \citet{benton2024nearly} to the
conditional OU diffusion with data law \(p_0(\cdot\mid y)\), with the stopping
time \(t_0\) playing the role of their early-stopping parameter. Their
path-space argument bounds the KL divergence between the exact reverse process
and the admissibly discretized reverse process by a score-estimation term plus a
discretization term. In the present notation this gives
\begin{align}
\mathrm{KL}\!\left(
p_{t_0}(\cdot\mid y)
\,\middle\|\,
\bar p_{t_0}^{\mathrm{dis}}(\cdot\mid y)
\right)
&\lesssim
\sum_{k=1}^{K}
\Delta_k\,
\mathbb E\!\left[
\left\|
\bar s(X_{t_k},y,t_k)
-
s^\star(X_{t_k},t_k,y)
\right\|_2^2
\,\middle|\,Y=y
\right]
\nonumber\\
&\quad+
\mathrm{Disc}_K(y),
\label{eq:conditional_benton_kl}
\end{align}
where the expectation in the score-error term is under the forward noised
marginal \(X_{t_k}\sim p_{t_k}(\cdot\mid y)\).

The use of the forward marginal in the score-error term is the key point. It is
exactly the form appearing in the randomized-time denoising objective. Thus this
bound does not require a true-score discretized reverse chain to have exact grid
marginals \(p_{t_k}(\cdot\mid y)\).

The initialization term is absent from \eqref{eq:conditional_benton_kl} because
both processes in this lemma are initialized from the same law
\(p_T(\cdot\mid y)\). Equivalently, this follows from the path-KL decomposition
in Eq.~(17) of \citet{benton2024nearly}: the path KL separates into a reverse
path-measure term and an initial-law mismatch term, and the latter is zero when
the two reverse processes start from the same \(p_T(\cdot\mid y)\).

It remains to control the discretization contribution. The general
second-moment bound in \citet[Theorem~2]{benton2024nearly} gives
\[
\mathrm{Disc}_K(y)
\lesssim
\kappa^2d_xK
+\kappa d_xT
+\kappa\mathbb E[\|X_0\|_2^2\mid Y=y].
\]
Combining this expression with the admissible-grid construction in
\citet[Corollary~1]{benton2024nearly} yields, up to factors polynomial in
\(T\) and logarithmic in \(1/t_0\),
\[
\mathrm{Disc}_K(y)
\le
\widetilde{\mathcal O}\!\left(
\frac{d_x+\mathbb E[\|X_0\|_2^2\mid Y=y]}{K}
\right).
\]
Taking the essential supremum over \(y\) gives
\[
\mathrm{Disc}_K(y)
\le
\widetilde{\mathcal O}\!\left(
\frac{d_x+m_2}{K}
\right)
\]
for \(P_Y\)-a.e.\ \(y\).

Pinsker's inequality and data processing from path space to the terminal
marginal yield, for each such \(y\),
\begin{align}
\mathrm{TV}\bigl(
p_{t_0}(\cdot\mid y),
\bar p_{t_0}^{\mathrm{dis}}(\cdot\mid y)
\bigr)
&\le
\sqrt{
\frac12\,
\mathrm{KL}\!\left(
p_{t_0}(\cdot\mid y)
\,\middle\|\,
\bar p_{t_0}^{\mathrm{dis}}(\cdot\mid y)
\right)
}
\nonumber\\
&\lesssim
\left(
\sum_{k=1}^{K}
\Delta_k\,
\mathbb E\!\left[
\left\|
\bar s(X_{t_k},y,t_k)
-
s^\star(X_{t_k},t_k,y)
\right\|_2^2
\,\middle|\,Y=y
\right]
\right)^{1/2}
\nonumber\\
&\quad+
\widetilde{\mathcal O}\!\left(
\sqrt{\frac{d_x+m_2}{K}}
\right).
\end{align}
Finally, take expectation over \(Y\). Jensen's inequality gives
\begin{align}
&\mathbb E_Y
\left[
\left(
\sum_{k=1}^{K}
\Delta_k\,
\mathbb E\!\left[
\left\|
\bar s(X_{t_k},Y,t_k)
-
s^\star(X_{t_k},t_k,Y)
\right\|_2^2
\,\middle|\,Y
\right]
\right)^{1/2}
\right]\nonumber\\
&\le
\left(
\sum_{k=1}^{K}
\Delta_k\,
\mathbb E\!\left[
\left\|
\bar s(X_{t_k},Y,t_k)
-
s^\star(X_{t_k},t_k,Y)
\right\|_2^2
\right]
\right)^{1/2}.
\end{align}
This proves \eqref{eq:forward_law_tv_stability}.
\end{proof}

\begin{lemma}[Tail localization for randomized-time denoising]
\label{lem:randomized_time_tail_localization}
Suppose Assumption~\ref{ass:distributional} holds. Let \(J\sim\pi_K\) be any
random grid index supported on \(\{1,\ldots,K\}\), with
\(t_J\in[t_0,T]\), and define
\[
X_{t_J}
=
e^{-t_J}X_0+\sqrt{1-e^{-2t_J}}\,\varepsilon,
\qquad
\varepsilon\sim\mathcal N(0,I_{d_x}).
\]
For \(R>0\), define
\[
\mathcal G(R)
:=
\left\{
\|Y\|_2\le R,\;
\|X_0\|_2\le R,\;
\|\varepsilon\|_2\le R
\right\}.
\]
Then there exist constants \(C,c>0\), depending only on the tail parameters and
the dimensions, such that
\[
\mathbb P(\mathcal G(R)^c)
\le
C e^{-cR}+C e^{-cR^2}.
\]
Moreover, on \(\mathcal G(R)\),
\[
\|X_{t_J}\|_2\le 2R,
\]
and the denoising target
\[
Z_J
:=
\nabla_x\log q_{t_J}(X_{t_J}\mid X_0)
=
-\frac{X_{t_J}-e^{-t_J}X_0}{1-e^{-2t_J}}
\]
satisfies
\[
\|Z_J\|_2
\le
\frac{R}{\sqrt{1-e^{-2t_0}}}.
\]
Finally, suppose \(s_\theta\) belongs to a bounded-parameter smooth-activation
network class whose output has polynomial growth in the input. Then, for every
integer \(m\ge1\),
\[
\mathbb E\!\left[
\left\|
s_\theta(X_{t_J},Y,t_J)
-
s^\star(X_{t_J},t_J,Y)
\right\|_2^m
\right]
\le
C_m\,\mathfrak P_m(\theta,M,t_0),
\]
where \(C_m<\infty\), and \(\mathfrak P_m(\theta,M,t_0)\) is a polynomial
envelope depending on the network-size and parameter bounds and on
\((1-e^{-2t_0})^{-1/2}\). In particular, for the approximation classes used in
the theorem, this moment grows at most polynomially in \(M\) and polylogarithmic
factors.
\end{lemma}

\begin{proof}
By Assumption~\ref{ass:distributional},
\[
\mathbb E e^{\lambda_Y\|Y\|}<\infty,
\qquad
\operatorname*{ess\,sup}_{y\sim P_Y}
\mathbb E\!\left[
e^{\lambda_X\|X_0\|}
\,\middle|\,Y=y
\right]<\infty .
\]
Markov's inequality therefore gives constants \(C_1,c_1>0\) such that
\[
\mathbb P(\|Y\|_2>R)+\mathbb P(\|X_0\|_2>R)
\le
C_1e^{-c_1R}.
\]
Since \(\varepsilon\sim\mathcal N(0,I_{d_x})\), standard Gaussian concentration
gives constants \(C_2,c_2>0\) such that
\[
\mathbb P(\|\varepsilon\|_2>R)\le C_2e^{-c_2R^2}.
\]
A union bound gives the stated bound on \(\mathbb P(\mathcal G(R)^c)\).

On \(\mathcal G(R)\),
\[
\|X_{t_J}\|_2
\le
e^{-t_J}\|X_0\|_2+
\sqrt{1-e^{-2t_J}}\|\varepsilon\|_2
\le
2R.
\]
Also,
\[
X_{t_J}-e^{-t_J}X_0
=
\sqrt{1-e^{-2t_J}}\,\varepsilon .
\]
Therefore,
\[
\|Z_J\|_2
=
\frac{\|\varepsilon\|_2}{\sqrt{1-e^{-2t_J}}}
\le
\frac{R}{\sqrt{1-e^{-2t_0}}},
\]
because \(t_J\ge t_0\).

For the moment bound, the denoising identity gives
\[
s^\star(X_{t_J},t_J,Y)
=
\mathbb E[Z_J\mid X_{t_J},Y,t_J].
\]
By Jensen's inequality,
\[
\mathbb E\!\left[
\|s^\star(X_{t_J},t_J,Y)\|_2^m
\right]
\le
\mathbb E\!\left[
\|Z_J\|_2^m
\right]
<\infty,
\]
where finiteness follows from Gaussian moments of \(\varepsilon\) and the lower
cutoff \(t_J\ge t_0>0\). Polynomial growth of \(s_\theta\), together with the
sub-exponential moments of \(X_0\) and \(Y\), Gaussian moments of \(\varepsilon\),
and the bounded time interval, gives the claimed polynomial envelope.
\end{proof}

\subsection{Manifold-adapted approximation input}
\label{subsec:manifold_approx_input}

\begin{lemma}[Localized manifold-adapted score approximation]
\label{lem:localized_manifold_score_approx}
Suppose Assumption~\ref{ass:distributional} holds, let \(M\ge2\), and let
\(J\sim\pi_K\) be the randomized grid-time index with
\(t_J\in[t_0,T]\). Define
\[
\Delta_{\max}:=\max_{1\le k\le K}(t_k-t_{k-1}),
\qquad
H_K:=\sum_{k=1}^{K}(t_k-t_{k-1}).
\]
Let
\[
\eta_{\mathrm{app}}
:=
\frac{2\alpha_x}{
2\alpha_x+d_x'+(\alpha_x/\alpha_y)d_y'
}.
\]
Let
\[
\mathcal G_M
:=
\left\{
\|Y\|_2\le C\log M,\;
\|X_0\|_2\le C\log M,\;
\|\varepsilon\|_2\le C\log M
\right\},
\]
where
\[
X_t=e^{-t}X_0+\sqrt{1-e^{-2t}}\,\varepsilon .
\]
Let \(u_0=t_0\) and define a dyadic cover of the time interval by
\[
u_{i+1}:=\min\{2u_i,T\},
\qquad
i=0,\ldots,I-1,
\]
so that
\[
[t_0,T]\subseteq \bigcup_{i=0}^{I-1}[u_i,u_{i+1}],
\qquad
I\le 1+\left\lceil \log_2\!\left(\frac{T}{t_0}\right)\right\rceil.
\]
Assume that the approximation-scale class \(\mathcal F_{\Theta,M}\) in
\eqref{eq:network_class_prelim} is chosen large enough to contain smooth
time-gated realizations of the dyadic-block conditional neural-network construction:
\[
(x,y,t)\mapsto
\sum_{i=0}^{I-1}\chi_i(t)\phi_i(x,y,t),
\]
where \(\{\chi_i\}_{i=0}^{I-1}\) is a nonnegative smooth partition of unity adapted to the
dyadic blocks, the time derivatives of the gates are controlled by
polylogarithmic factors in the theorem regime, and each \(\phi_i\) belongs to
the corresponding blockwise conditional score-approximation class used in the
anisotropic manifold-adaptive approximation theorem of
\citet{tang2025conditional}. Then there exists
\(\widetilde s\in\mathcal F_{\Theta,M}\) such that
\[
\mathbb E\!\left[
\left\|
\widetilde s(X_{t_J},Y,t_J)
-
s^\star(X_{t_J},t_J,Y)
\right\|_2^2
\mathbf 1_{\mathcal G_M}
\right]
\le
\widetilde{\mathcal O}\!\left(
M^{-\eta_{\mathrm{app}}}
+
\Delta_{\max}
\right).
\]
In particular, if
\[
\Delta_{\max}
\le
\widetilde{\mathcal O}\!\left(
M^{-\eta_{\mathrm{app}}}
\right),
\]
then
\[
\mathbb E\!\left[
\left\|
\widetilde s(X_{t_J},Y,t_J)
-
s^\star(X_{t_J},t_J,Y)
\right\|_2^2
\mathbf 1_{\mathcal G_M}
\right]
\le
\widetilde{\mathcal O}\!\left(
M^{-\eta_{\mathrm{app}}}
\right).
\]
The hidden logarithmic factors may depend on the localization radius, the dyadic
block count \(I\), the lower cutoff \(t_0\), the upper horizon \(T\), and
the uniform chart and local smoothness constants in
Assumption~\ref{ass:distributional}.
\end{lemma}

\begin{proof}
The proof reduces the localized approximation problem to the anisotropic
time-dependent conditional manifold approximation theorem of
\citet{tang2025conditional}. The additional steps are localization, smooth
time-gating, and transfer from continuous-time integrated error to the discrete
randomized grid-time error used in \(\mathcal L_K\).

On the event \(\mathcal G_M\), the variables \(X_0\), \(Y\), and
\(\varepsilon\) are restricted to a compact set of radius
\[
R_M:=C\log M.
\]
Since \(t\ge t_0>0\), the OU variance \(1-e^{-2t}\) is uniformly bounded below.
Consequently,
\[
s^\star(x,t,y)=\nabla_x\log p_t(x\mid y)
\]
is a Gaussian-smoothed conditional score on the localized region. The smoothness
part of Assumption~\ref{ass:distributional}, together with the lower cutoff
\(t_0\), gives local anisotropic H\"older and time-regularity constants on this
compact region. These constants grow at most polylogarithmically in \(M\), with
additional dependence on \(t_0^{-1}\), \(T\), and the chart constants.

We now identify the approximation input. In the notation of
\citet{tang2025conditional}, the covariate is denoted by \(x\), the noisy
response by \(w\), and the approximating network by \(\phi_i(w,x,t)\) on a
dyadic time block. We apply the same construction after the relabeling
\[
x_{\mathrm{Tang}} \leftrightarrow y,
\qquad
w_{\mathrm{Tang}} \leftrightarrow x .
\]
Thus their conditional score
\(\nabla\log p_t(\cdot\mid x)(w)\) corresponds here to
\[
\nabla_x\log p_t(x\mid y)=s^\star(x,t,y).
\]

The manifold assumptions needed for this application are supplied locally by
Assumption~\ref{ass:distributional}. The conditioning variable \(Y\) has
intrinsic dimension \(d_y'\), and, for each \(y\), the clean conditional support
of \(X_0\mid Y=y\) has intrinsic dimension \(d_x'\). The fibered chart
regularity in Assumption~\ref{ass:distributional} ensures that, on the localized
region, the clean pair \((X_0,Y)\) is represented by
\(d_x'+d_y'\) intrinsic coordinates with uniformly controlled chart,
encoder--decoder, and local smoothness constants. The approximation is
anisotropic: the response coordinate has smoothness \(\alpha_x\), while the
conditioning coordinate has smoothness \(\alpha_y\).

The anisotropic balance gives the effective approximation exponent
\[
\eta_{\mathrm{app}}
=
\frac{2\alpha_x}{
2\alpha_x+d_x'+(\alpha_x/\alpha_y)d_y'
}.
\]
Equivalently, if \(h_x\) denotes the response-coordinate resolution and
\(h_y\) the conditioning-coordinate resolution, the balanced choice satisfies
\[
h_y \asymp h_x^{\alpha_x/\alpha_y},
\qquad
h_x^{2\alpha_x}\asymp h_y^{2\alpha_y},
\]
and the effective intrinsic covering dimension is
\[
d_x'+(\alpha_x/\alpha_y)d_y' .
\]
Choosing the approximation scale indexed by \(M\) according to
\[
h_x
\asymp
M^{-1/\{2\alpha_x+d_x'+(\alpha_x/\alpha_y)d_y'\}}
\]
yields squared localized approximation error
\[
h_x^{2\alpha_x}
=
M^{-\eta_{\mathrm{app}}}.
\]
This is the anisotropic version of the Lipschitz-specialized rate
\(M^{-2/(2+d_x'+d_y')}\), which is recovered only when
\(\alpha_x=\alpha_y=1\).

Applying the blockwise anisotropic conditional score approximation theorem on
each dyadic block \([u_i,u_{i+1}]\), with the localized chart and smoothness
constants absorbed into logarithmic factors, gives blockwise networks
\(\phi_i\) satisfying
\[
\frac{1}{H_K}
\sum_{i=0}^{I-1}
\int_{u_i}^{u_{i+1}}
\mathbb E\!\left[
\left\|
\phi_i(X_t,Y,t)-s^\star(X_t,t,Y)
\right\|_2^2
\mathbf 1_{\mathcal G_M}
\right]dt
\le
\widetilde{\mathcal O}\!\left(
M^{-\eta_{\mathrm{app}}}
\right).
\]
The logarithmic factors absorb the dyadic block count \(I\), the localized
radius \(R_M=C\log M\), the chart cover, the lower-time cutoff \(t_0\), and the
upper horizon \(T\).

The cited construction is blockwise in time. Our network class uses a single
time-conditioned network. Let \(\{\chi_i\}_{i=0}^{I-1}\) be smooth gates forming
a partition of unity over a slightly enlarged dyadic cover, with controlled
time derivatives. Define
\[
\widetilde s(x,y,t)
:=
\sum_{i=0}^{I-1}
\chi_i(t)\phi_i(x,y,t).
\]
By the containment condition on \(\mathcal F_{\Theta,M}\),
\(\widetilde s\in\mathcal F_{\Theta,M}\). By convexity of the squared norm and
the partition-of-unity property, the same integrated approximation bound holds
for \(\widetilde s\), up to logarithmic factors:
\[
\frac{1}{H_K}
\int_{t_0}^{T}
\mathbb E\!\left[
\left\|
\widetilde s(X_t,Y,t)-s^\star(X_t,t,Y)
\right\|_2^2
\mathbf 1_{\mathcal G_M}
\right]dt
\le
\widetilde{\mathcal O}\!\left(
M^{-\eta_{\mathrm{app}}}
\right).
\]

It remains to transfer the continuous-time integrated approximation bound to
the discrete randomized-time quantity used in \(\mathcal L_K\). Define
\[
a(t)
:=
\mathbb E\!\left[
\left\|
\widetilde s(X_t,Y,t)-s^\star(X_t,t,Y)
\right\|_2^2
\mathbf 1_{\mathcal G_M}
\right].
\]
On \([t_0,T]\), the localized score, the OU transition map, and the
smooth time-gated network have time-regularity constants controlled by
polylogarithmic factors in the theorem regime. Equivalently, on each grid
interval \([t_{k-1},t_k]\), the localized approximation-error curve \(a(t)\) is
absolutely continuous with modulus
\[
|a(t)-a(t')|
\le
\widetilde{\mathcal O}(1)\,|t-t'|,
\qquad
t,t'\in[t_{k-1},t_k],
\]
where the hidden factor may depend on \(t_0^{-1}\), \(T\), the localized
smoothness constants, and the time-gate derivative bounds. Therefore,
\[
a(t_k)
\le
\frac{1}{t_k-t_{k-1}}
\int_{t_{k-1}}^{t_k}a(t)\,dt
+
\widetilde{\mathcal O}(\Delta_{\max}).
\]
This is the only place where time regularity of the smoothed score and the
smooth time gates is used; equivalently, it can be viewed as a standard
Riemann-sum transfer bound for the localized approximation-error function
\(a(t)\).

Multiplying by \(\Delta_k=t_k-t_{k-1}\), summing over \(k\), and dividing by
\(H_K\), we obtain
\[
\sum_{k=1}^{K}
\frac{\Delta_k}{H_K}a(t_k)
\le
\frac{1}{H_K}
\int_{t_0}^{T}a(t)\,dt
+
\widetilde{\mathcal O}(\Delta_{\max}).
\]
Since \(\mathbb P(J=k)=\Delta_k/H_K\), the left-hand side is exactly
\[
\mathbb E\!\left[
\left\|
\widetilde s(X_{t_J},Y,t_J)
-
s^\star(X_{t_J},t_J,Y)
\right\|_2^2
\mathbf 1_{\mathcal G_M}
\right].
\]
Combining the grid-transfer bound with the anisotropic dyadic-block
approximation bound gives
\[
\mathbb E\!\left[
\left\|
\widetilde s(X_{t_J},Y,t_J)
-
s^\star(X_{t_J},t_J,Y)
\right\|_2^2
\mathbf 1_{\mathcal G_M}
\right]
\le
\widetilde{\mathcal O}\!\left(
M^{-\eta_{\mathrm{app}}}
+
\Delta_{\max}
\right).
\]
If
\[
\Delta_{\max}
\le
\widetilde{\mathcal O}\!\left(M^{-\eta_{\mathrm{app}}}\right),
\]
the displayed bound reduces to
\[
\widetilde{\mathcal O}\!\left(M^{-\eta_{\mathrm{app}}}\right).
\]
This proves the lemma.
\end{proof}
\section{Experiments}
\label{sec:experiments}

The theory analyzes a single time-conditioned conditional score network
\[
s_\theta(x_t,y,t)
\]
trained with fresh randomized-time denoising samples. The experiments in this
appendix are therefore organized around the corresponding online training
protocol. The goal is not to verify the asymptotic rates or prove the structural
assumptions empirically. Rather, the experiments test whether the analyzed
training model can be implemented directly, whether the approximation-complexity
and online-budget effects can be separated in practice, and what trajectory
behavior is observed in relation to the local geometry assumed in the proof.

\subsection{Experimental Goals and Scope}
\label{subsec:experimental_goals}

The empirical study addresses five questions motivated by the theorem.
First, can a single randomized-time score network be trained successfully using
fresh online denoising samples? Second, can the effects of model complexity
\(M\) and online denoising budget \(N=RB\) be varied separately? Third, does the
randomized-time objective exhibit stable finite-run loss and parameter
trajectories? Fourth, does
score-estimation difficulty increase with intrinsic dimension in a generator
that satisfies the fibered-manifold structure? Finally, does the online
training procedure remain informative under conditional distributions with
genuine sub-exponential tails?

All experiments use controlled synthetic conditional laws for which the
conditional score can be computed analytically or by exact one-dimensional
normal-Laplace score evaluation. This lets us measure score error directly,
rather than relying only on downstream sample quality. Since the theory assumes
smooth activations, the score network is a fully connected MLP with
\(\mathrm{Softplus}_{10}\) activation, a smooth approximation to ReLU. This MLP
is an implementation surrogate for the theoretical smooth-gated blockwise
class; it does not instantiate the exact approximation construction.

\subsection{Online Randomized-Time DDPM Training}
\label{subsec:online_randomized_time_training}

We use an implementation-time coordinate \(u\) and a variance-preserving
forward process
\[
x_u=\alpha(u)x_0+\sqrt{1-\alpha(u)^2}\,\xi,
\]
with a linear \(\beta\)-schedule. Define the effective OU time
\(\tau(u):=-\log\alpha(u)\); then
\(x_u=e^{-\tau(u)}x_0+\sqrt{1-e^{-2\tau(u)}}\,\xi\).
For an implementation grid \(u_0<\cdots<u_K\), set
\(t_k:=\tau(u_k)\). The interval weights are formed on this effective OU-time
grid using the right-endpoint convention in the analysis. At each SGD
iteration, we draw a fresh mini-batch of conditional data \((X_0,Y)\), a grid
index \(J\), and Gaussian diffusion noise. We then form
\[
X_{t_J}=\alpha(u_J)X_0+\sqrt{1-\alpha(u_J)^2}\,\varepsilon
\]
and train with the denoising loss
\[
\ell(\theta)
=
\left\|s_\theta(X_{t_J},Y,t_J)
+\frac{X_{t_J}-\alpha(u_J)X_0}{1-\alpha(u_J)^2}
\right\|_2^2 .
\]
The reported online budget is \(N=RB\), where \(R\) is the number of SGD
updates and \(B\) is the mini-batch size. The optimizer is plain SGD with a
decreasing step size of the form \(\eta_r=c/(r+r_0)\).

Our baseline conditional benchmark is
\[
x_0 = W_y y + W_z z + \sigma_{\mathrm{obs}}\varepsilon,
\]
with Gaussian \(y,z,\varepsilon\). Conditional on \(y\), the law of \(x_t\) is
Gaussian, so the true conditional score is available in closed form. The primary
readout is randomized-time score MSE evaluated on fresh Monte Carlo batches.

\begin{figure}[t]
  \centering
  \includegraphics[width=0.82\linewidth]{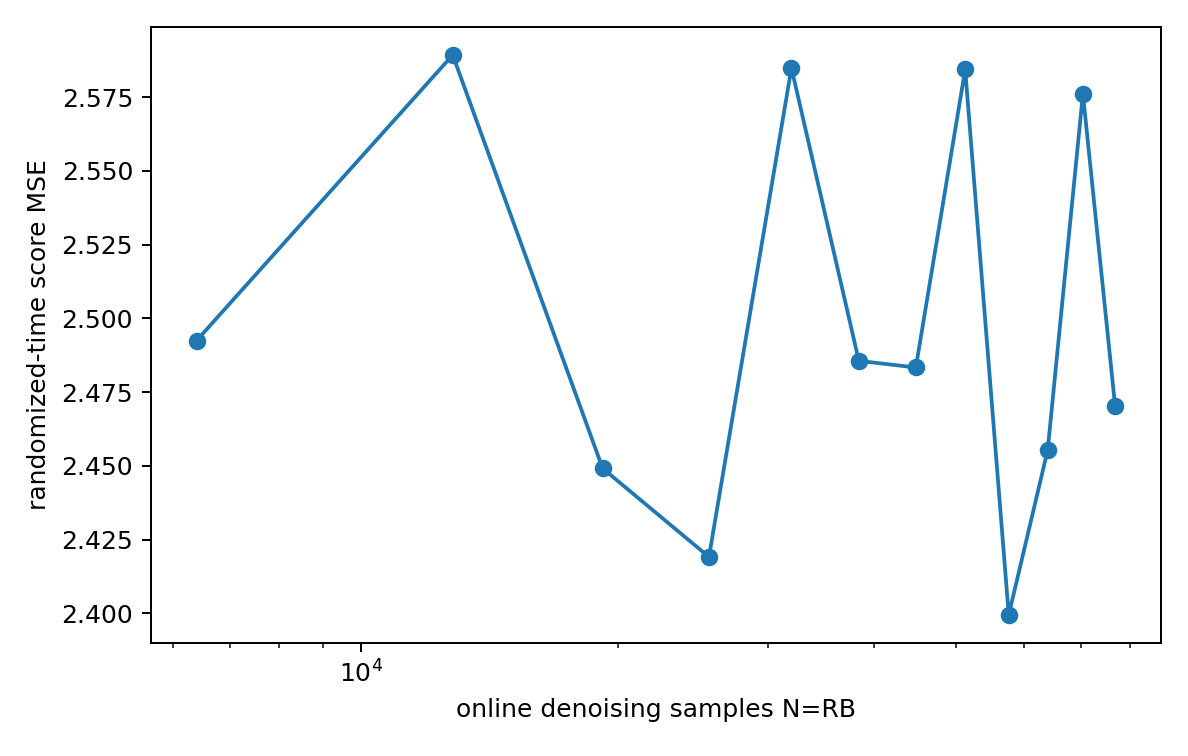}
  \caption{Randomized-time score MSE versus online denoising budget \(N=RB\) for a single time-conditioned score network trained with fresh online SGD samples.}
  \label{fig:online-score-budget}
\end{figure}

Figure~\ref{fig:online-score-budget} shows the pilot online-budget trajectory
from the online randomized-time training loop. The error remains finite and fluctuates
within a relatively narrow range, but it does not decrease systematically over
the displayed budget. Thus this single-seed pilot establishes only that the
online randomized-time protocol can be executed without separate frozen-time
training; it is not evidence of a sample-complexity trend or successful
convergence. Larger multi-seed runs and additional tuning would be needed for
such a claim.

\subsection{Separating Approximation Complexity and Online Budget}
\label{subsec:capacity_budget}

The theorem separates an approximation term depending on the score
class complexity \(M\) from an online optimization term depending on the fresh
sample budget \(N\). To mirror this decomposition, we vary network width while
holding depth and activation fixed. Width and parameter count are used only as
operational proxies for \(M\); we do not claim that parameter count realizes the
theoretical approximation index exactly.

\begin{figure}[t]
  \centering
  \includegraphics[width=0.82\linewidth]{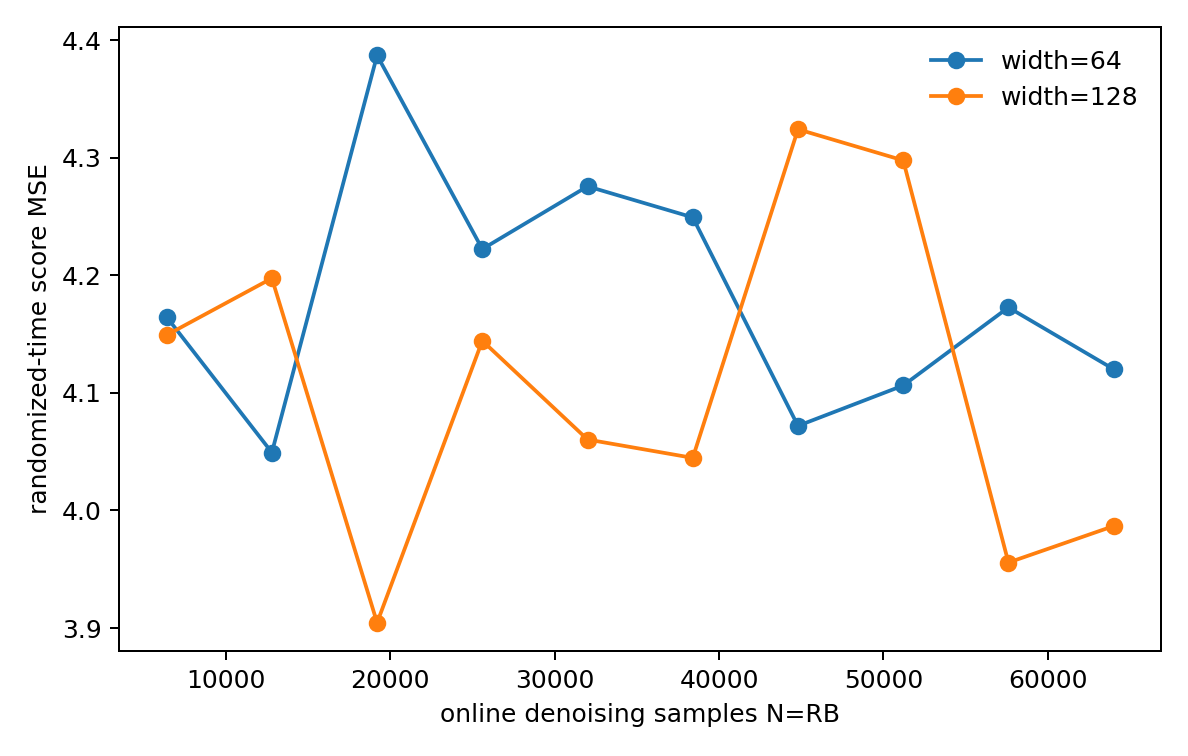}
  \caption{Randomized-time score MSE versus online budget \(N\), with network width used as an operational proxy for approximation complexity \(M\).}
  \label{fig:capacity-budget-curves}
\end{figure}

\begin{figure}[t]
  \centering
  \includegraphics[width=0.72\linewidth]{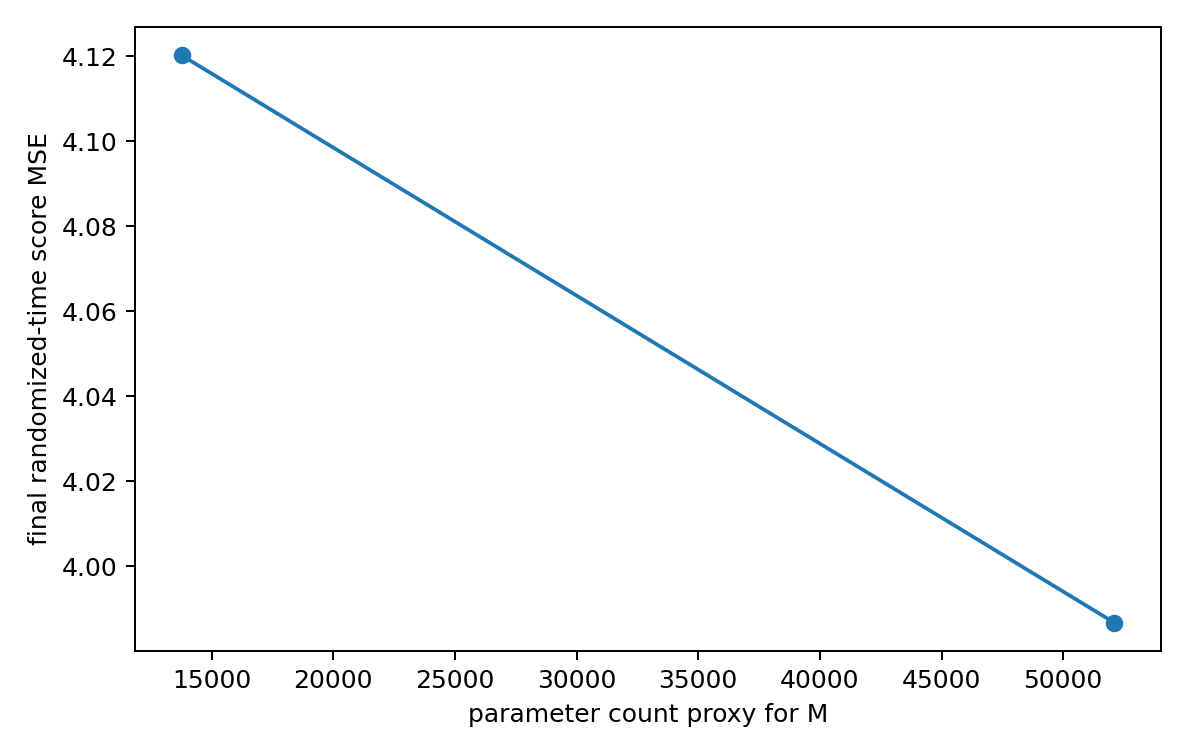}
  \caption{Final randomized-time score MSE versus parameter count, used as a practical proxy for the complexity index \(M\).}
  \label{fig:capacity-final}
\end{figure}

\begin{figure}[t]
  \centering
  \includegraphics[width=0.72\linewidth]{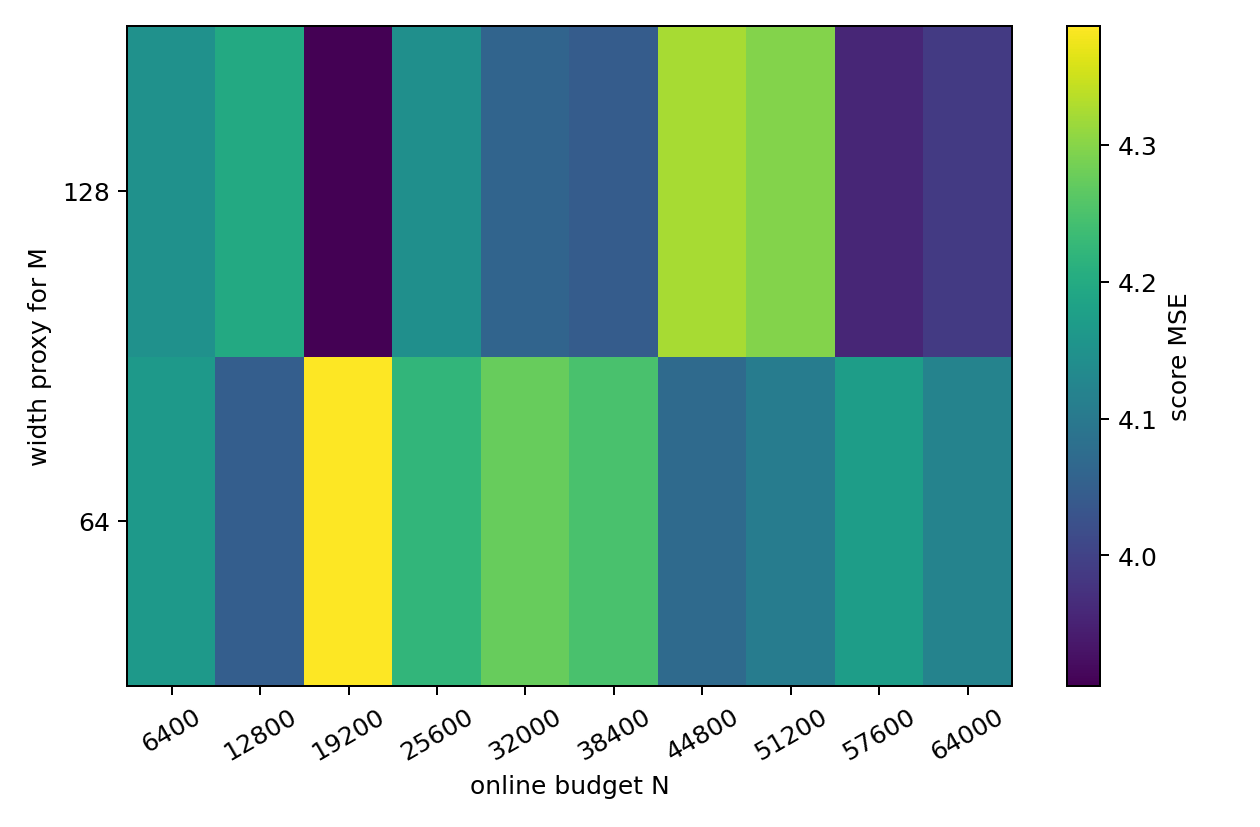}
  \caption{Joint \(M\)-versus-\(N\) diagnostic. The heat map summarizes randomized-time score MSE over width and online denoising budget.}
  \label{fig:capacity-budget-heatmap}
\end{figure}

Figures~\ref{fig:capacity-budget-curves}--\ref{fig:capacity-budget-heatmap}
show the resulting \(M\)-versus-\(N\) diagnostic. In the current pilot run, the
curves are not monotone at every checkpoint, reflecting the variability of
plain small-batch SGD. Nevertheless, the experiment cleanly separates the two
axes emphasized by the theorem: one can hold architecture fixed and vary the
online budget, or hold budget fixed and compare model capacities. We therefore
interpret this experiment as an implementation-level diagnostic for the
decomposition
\[
\mathcal A_K(M)+\mathcal E_{\mathrm{opt}}(N,M),
\]
not as an empirical fit of the theoretical exponents.

\subsection{Randomized-Time Optimization and Trajectory Diagnostics}
\label{subsec:randomized_time_pl}

The optimization analysis assumes a localized PL-type condition for the
randomized-time denoising objective. We report related finite-trajectory
diagnostics using evaluation batches independent of the online training samples;
these diagnostics do not test the PL inequality. At checkpoints \(r\), we
estimate an evaluation loss \(\widehat{\mathcal L}_{K,\mathrm{eval}}(\theta_r)\)
and its gradient norm. Let \(L_{\mathrm{ref}}\) be the smallest evaluation loss
observed along the trajectory. The plotted quantity
\(\widehat{\mathcal L}_{K,\mathrm{eval}}(\theta_r)-L_{\mathrm{ref}}\) is a
reference-subtracted loss, not an estimate of the population excess risk
\(\mathcal L_K(\theta_r)-\mathcal L_{K,M}^\star\). We also track the parameter
displacement \(\|\theta_r-\theta_0\|_2\). These are trajectory diagnostics only.

\begin{figure}[t]
  \centering
  \includegraphics[width=0.82\linewidth]{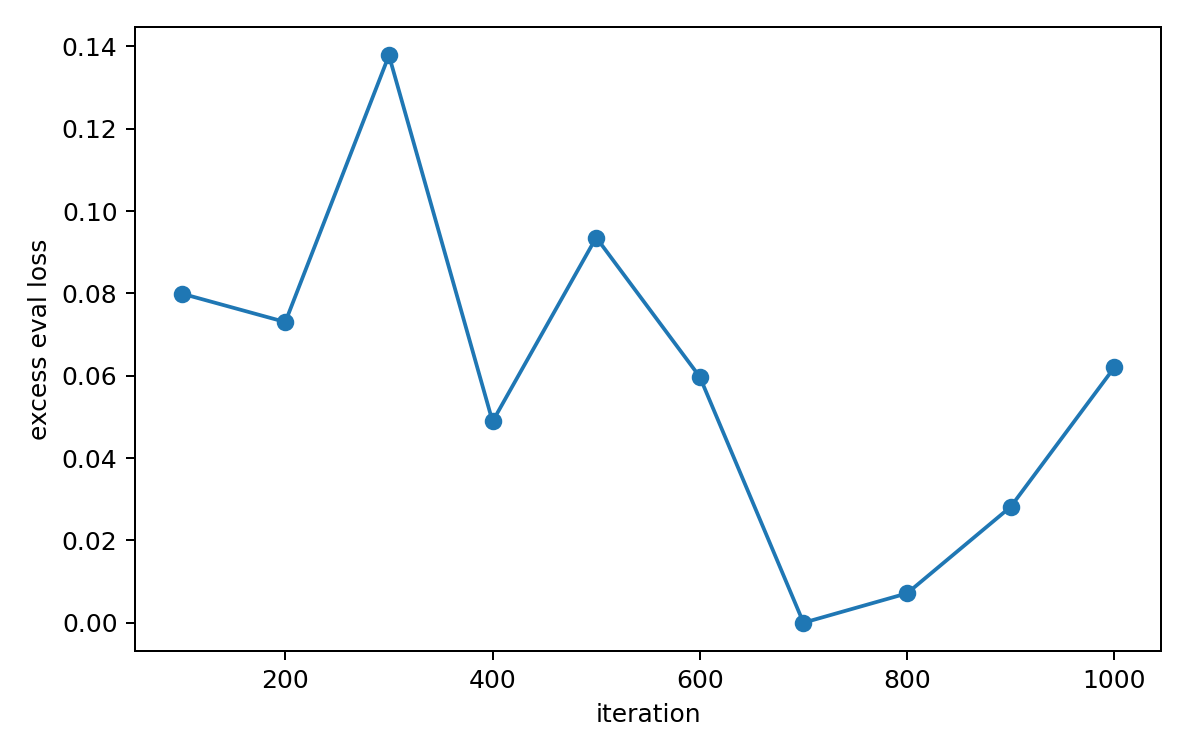}
  \caption{Randomized-time reference-subtracted evaluation loss along the online SGD trajectory.}
  \label{fig:randomized-time-excess}
\end{figure}

\begin{figure}[t]
  \centering
  \includegraphics[width=0.82\linewidth]{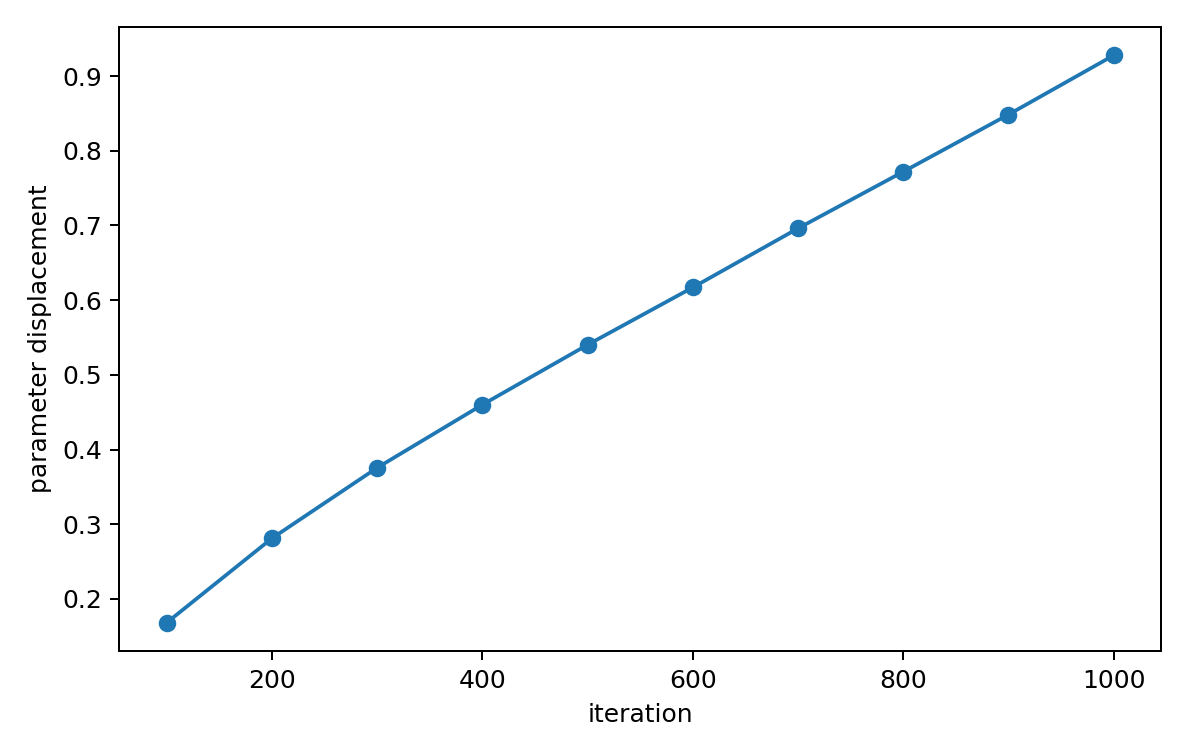}
  \caption{Parameter displacement \(\|\theta_r-\theta_0\|_2\) along the online SGD trajectory.}
  \label{fig:parameter-displacement}
\end{figure}

The plug-in PL ratio obtained by replacing the unknown optimum with
\(L_{\mathrm{ref}}\) is not reported: its denominator becomes numerically
unstable near a checkpoint attaining the empirical reference loss and produces
an artificial spike. The displacement curve records the radius explored in
this finite run, but it does not identify a region on which the population PL
inequality holds and does not establish the trajectory-containment assumption.

\subsection{Fiber-Compatible Intrinsic Dimension}
\label{subsec:fiber_intrinsic_dimension}

The theory uses a joint fibered-manifold structure. To avoid making the
clean response full dimensional through ambient observation noise, we use the
synthetic construction
\[
V\in\mathbb R^{d_y'}, \qquad Z\in\mathbb R^{d_x'},
\]
\[
Y=\Phi_Y(V), \qquad X_0=m(Y)+BZ,
\]
where \(B\in\mathbb R^{d_x\times d_x'}\) has rank \(d_x'\). The map
\(\Phi_Y\) concatenates linear, sinusoidal, and quadratic features, giving a
mildly nonlinear conditioning manifold. Conditional on \(Y=y\), the positive
time law \(X_t\mid Y=y\) is Gaussian with covariance
\[
\alpha(t)^2BB^\top+(1-\alpha(t)^2)I_{d_x},
\]
so the conditional score remains available in closed form.

\begin{figure}[t]
  \centering
  \includegraphics[width=0.78\linewidth]{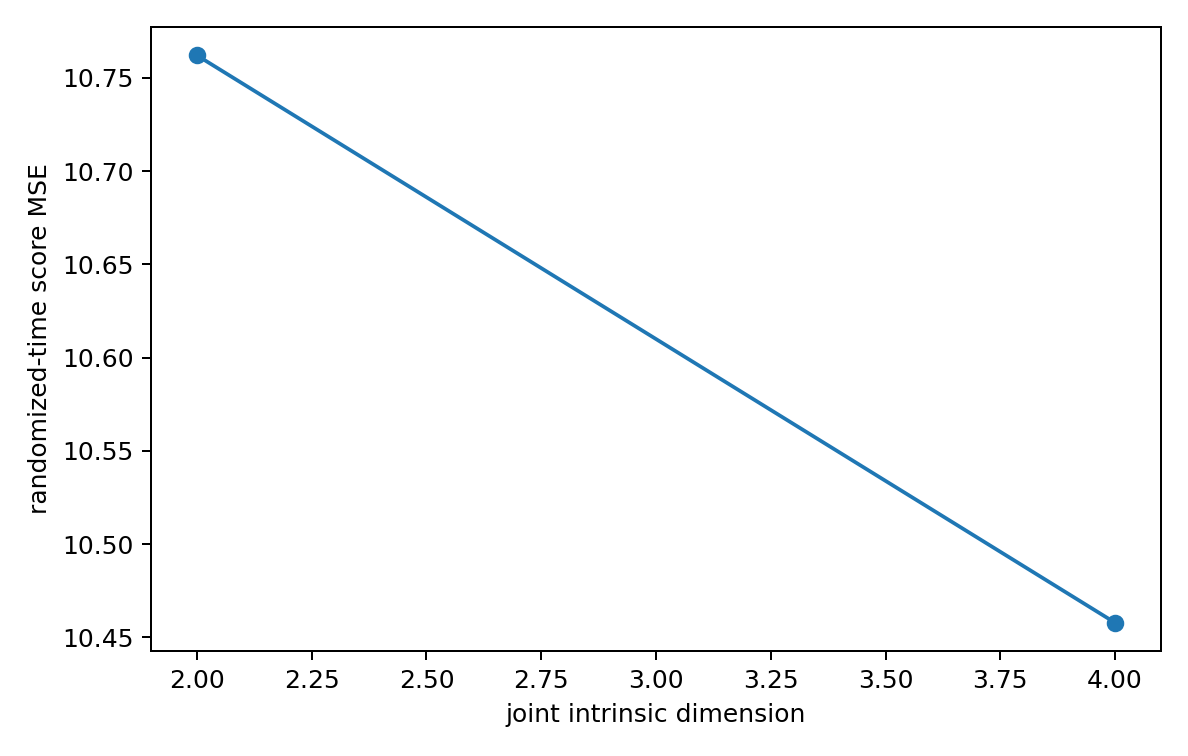}
  \caption{Randomized-time score MSE for the fiber-compatible intrinsic-dimension generator. The ambient dimension is fixed while the joint intrinsic dimension is varied.}
  \label{fig:fiber-intrinsic-dim}
\end{figure}

Figure~\ref{fig:fiber-intrinsic-dim} reports the pilot intrinsic-dimension
diagnostic. The current smoke run covers only a reduced grid, so we do not read
off a scaling law from it. Its main role is to instantiate the exact
fiber-compatible construction required by the theory and confirm that
the analytic score evaluation remains numerically usable at positive diffusion
times.

\subsection{Sub-Exponential Tail Stress Test}
\label{subsec:subexponential_tail}

Finally, we stress-test the online learner under conditionals with genuine
sub-exponential tails. We use
\[
x_0=W_y y+\eta_G+\eta_L,
\]
where \(\eta_G\) is coordinatewise Gaussian and \(\eta_L\) is coordinatewise
Laplace. The total coordinatewise variance is matched across regimes, so changes
in performance are driven by tail shape rather than by a trivial scale change.
After forward diffusion, each coordinate follows a normal-Laplace law, and the
score is evaluated using the exact normal-Laplace density.

\begin{figure}[t]
  \centering
  \includegraphics[width=0.82\linewidth]{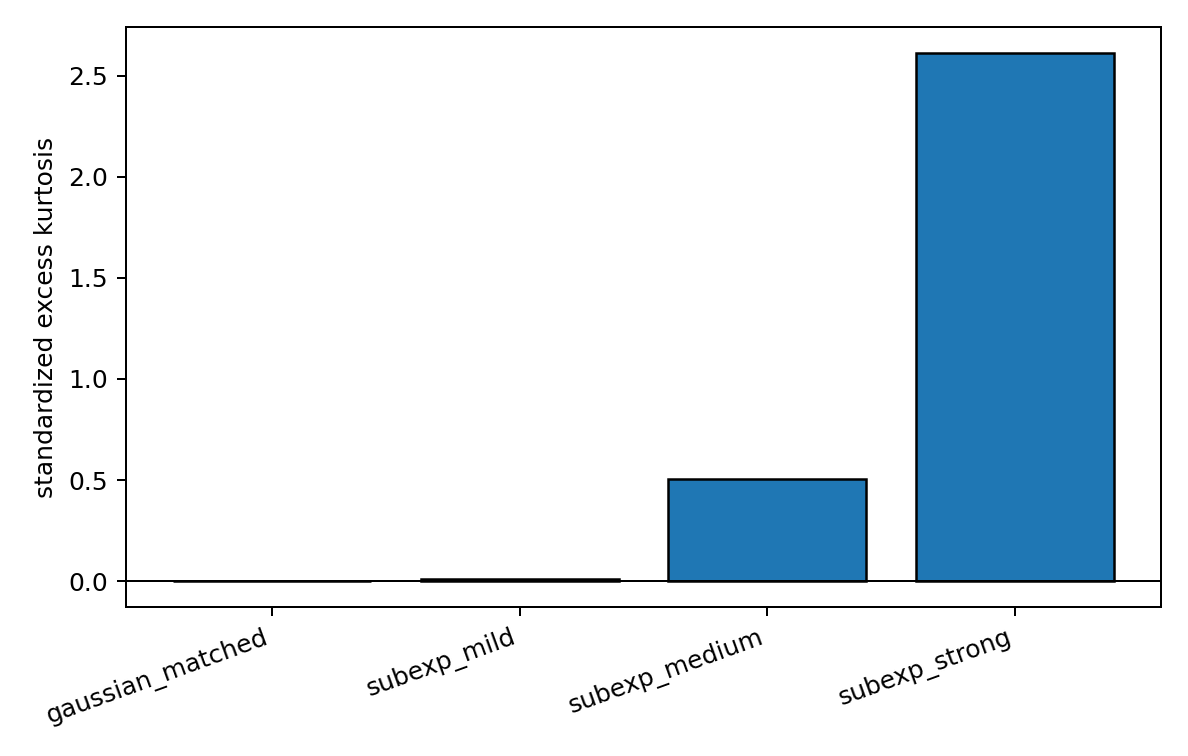}
  \caption{Standardized excess kurtosis for the Gaussian-matched and sub-exponential regimes. Larger Laplace components produce visibly heavier tails.}
  \label{fig:subexp-online-kurtosis}
\end{figure}

\begin{figure}[t]
  \centering
  \includegraphics[width=0.82\linewidth]{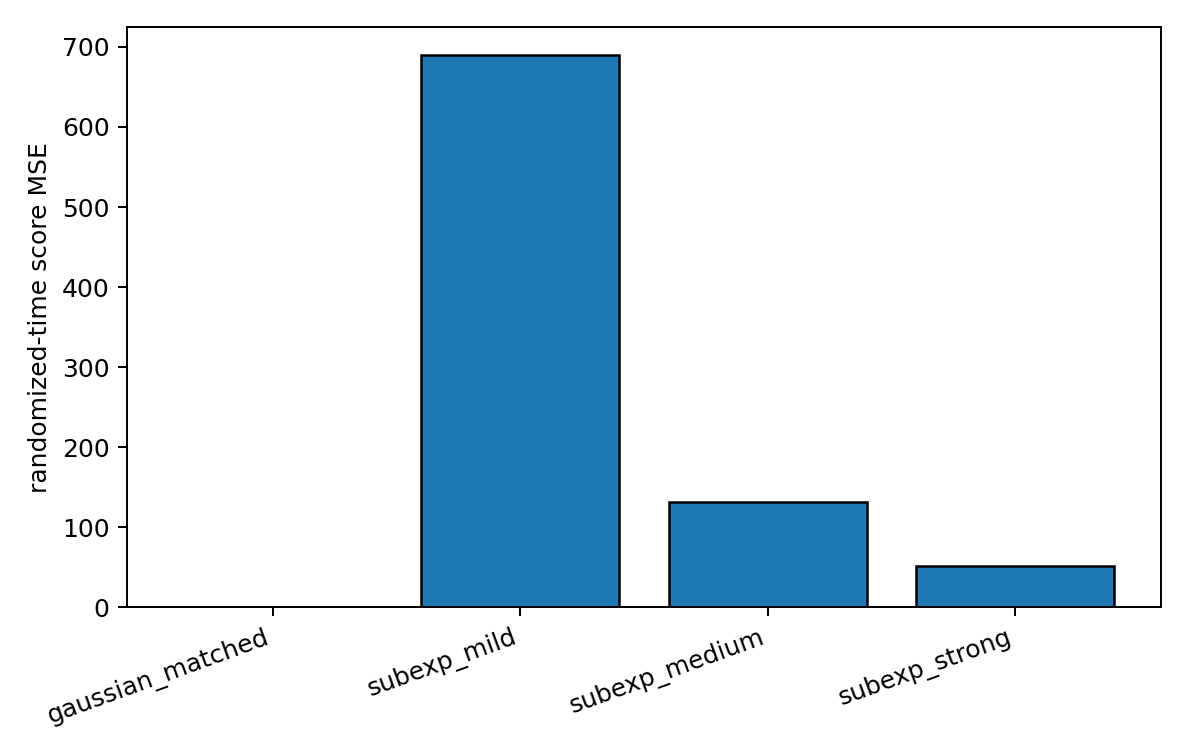}
  \caption{Randomized-time score MSE under online SGD for Gaussian-matched and sub-exponential conditional laws.}
  \label{fig:subexp-online-score}
\end{figure}

Figure~\ref{fig:subexp-online-kurtosis} confirms that the synthetic regimes
become increasingly heavy-tailed as the Laplace component grows. In the current
online-SGD pilot, Figure~\ref{fig:subexp-online-score} also reveals that exact
normal-Laplace score evaluation can make the score-MSE readout numerically
sensitive for some intermediate regimes. We therefore interpret this experiment
as a stress test of the online pipeline under unbounded sub-exponential tails:
the tail construction itself matches the assumption, while the score-error
readout should be stabilized with larger evaluation batches and full multi-seed
runs before being used as a final quantitative claim.

\subsection{Summary}
\label{subsec:experimental_summary}

The experiments implement the key training model appearing in the
theorem: a single smooth time-conditioned score network trained by online SGD
with fresh randomized-time denoising samples. The implementation also separates
model capacity from online budget and adds randomized-time trajectory
diagnostics, implements a fiber-compatible intrinsic-dimension generator, and
retains a sub-exponential-tail stress test. The current pilot results show that
the pipeline is operational, while also showing that the online-SGD diagnostics
are noisy and do not yet display a clear learning curve. The appropriate
empirical claim is therefore limited: these experiments instantiate the
training protocol and stress-test several ingredients of the setup, but they do
not validate the population PL condition, establish convergence or scaling
exponents, or replace the formal theorem.

\end{document}